\documentclass{article}
\usepackage[letterpaper, margin=1in]{geometry}

\usepackage{times}
\usepackage{url} %
\usepackage{booktabs} %
\usepackage{graphicx} %
\usepackage{appendix}
\usepackage{amsmath}
\usepackage{amsthm}
\usepackage{amssymb}
\usepackage{color}
\usepackage[ruled,vlined,linesnumbered]{algorithm2e}
\usepackage{subcaption}
\usepackage[authoryear, round]{natbib}
\usepackage{rotating}
\usepackage{wrapfig}
\usepackage{placeins}
\usepackage{soul}   %
\usepackage{authblk}

\usepackage{caption}
\usepackage{subcaption}
\usepackage{float}
\usepackage{accents}
\usepackage{multirow}

\usepackage{hyperref} 
\hypersetup{
    colorlinks=true,
    linkcolor=blue,
    filecolor=blue,      
    urlcolor=blue,
    citecolor=blue,
    }

\DeclareMathOperator{\supp}{supp}
\renewcommand{\v}[1]{\ensuremath{\mathbf{#1}}}

\newcommand{\softmax}{\ensuremath{\operatorname{Softmax}}}
\newcommand{\x}{\ensuremath{\mathbf{x}}}
\newcommand{\z}{\ensuremath{\mathbf{z}}}
\newcommand{\E}{\ensuremath{\mathbb{E}}}
\newcommand{\Dmf}{\ensuremath{D_{\mathrm{MF}}}}
\newcommand{\Ahat}{\ensuremath{\hat{A}}}

\usepackage{xspace}
\newcommand{\stcond}{$(s,t)$-conditional \xspace}

\newcounter{tool}
\newcommand{\tool}[1]{\refstepcounter{tool}\paragraph{Lens \thetool: #1}}

\definecolor{hcInduction}{RGB}{200,50,40}
\definecolor{hcNameMover}{RGB}{40,150,80}
\definecolor{hcDupTok}{RGB}{230,140,40}
\definecolor{hcPrevTok}{RGB}{60,175,190}
\definecolor{hcOther}{RGB}{120,120,120}
\definecolor{hcPositional}{RGB}{20,70,180}
\definecolor{hcCont}{RGB}{140,70,160}
\newcommand{\hInd}[1]{\textcolor{hcInduction}{#1}}
\newcommand{\hMover}[1]{\textcolor{hcNameMover}{#1}}
\newcommand{\hDup}[1]{\textcolor{hcDupTok}{#1}}
\newcommand{\hPrev}[1]{\textcolor{hcPrevTok}{#1}}
\newcommand{\hOther}[1]{\textcolor{hcOther}{#1}}
\newcommand{\hPos}[1]{\textcolor{hcPositional}{#1}}
\newcommand{\hCont}[1]{\textcolor{hcCont}{#1}}

\newcounter{mfass}
\newenvironment{mfass}[1]
  {\par\medskip\refstepcounter{mfass}%
   \noindent\textbf{Assumption~\themfass.~#1.}\enspace}
  {\par\medskip}

\newcounter{mfprop}

\title{Attention Mean Fields Predict Average Representation Dynamics and Reveal Context-Specific Computation}

\author[ ]{Micah Adler,}
\author[1]{John W. Byers,}
\author[1,2]{and Mark Crovella}

\affil[1]{Department of Computer Science, Boston University}
\affil[2]{Faculty of Computing \& Data Sciences, Boston University}
\date{September 14, 2026}

\begin{document}

\maketitle

\begin{abstract} A language model's representation geometry is not predetermined; it evolves as the model runs, with each layer reshaping %
it
through attention that depends upon the surrounding context. A faithful account of that geometry must capture that dynamic process, and so cannot be based 
solely on model-independent statistics such as co-occurrence.
Here we introduce a mean-field analysis of attention.
The average attention from one token to another
defines a kernel that carries
representations 
layer to layer and can be iterated through the
network to model how the geometry is transformed. 
We condition this average two ways.  Conditioned on a whole corpus, the kernel predicts the 
average-case evolution of representation geometry. 
Conditioned instead on a single context, it predicts the 
expected geometry for that context.
A head's departure from that prediction, its \emph{mean-field deviation},
isolates the context-specific computation that the mean field misses.

Under the corpus-conditional reading, the kernel yields an open-loop model: from the input embeddings and the frozen weights alone, we can iterate the kernel and the model's own MLPs over token representations, never consulting a measured deviation at any layer.
With no free parameters, the rollout reproduces the trajectory of each token's
mean representation 
across models from GPT-2 to Qwen-3-14B.
We show that several alternatives, including one built from co-occurrence, are all poorer predictors.

In early training the model and its corpus mean field are indistinguishable.
Replace every attention head with its mean field, and  
across three scales of Pythia 
— 160m, 1.4B, and 12B —
the substitution leaves the loss on real text unchanged.
Around the onset of induction, the two diverge,
and
the gap widens as representations become contextualized and attention becomes correlated with the values it transports.
Recent theory characterizes early-training weights as closed-form compositions of corpus statistics; we give a mean-field account of the computation, %
under which the young model's behavior reduces to its corpus mean field until in-context computation emerges.

Under the context-conditional reading,
deviation from the mean field is a task-agnostic measure of context-specific computation. 
The residual decomposes additively into unusual attention routing and contextualization of the transported values. 
Across controlled induction and few-shot settings, greater deviation 
tracks
greater reliance on in-context information.
Ranking heads by corpus-level deviation recovers documented in-context machinery while also revealing positional and content-tracking heads.
On ICL tasks, task-conditioned residuals act as function vectors; we show that selecting function-vector heads by mean-field deviation achieves transfer accuracy comparable to causal-mediation selection,
 and only requires a single forward pass.
 \end{abstract}

\section{Introduction} \label{sec:intro} What happens inside a language model, \emph{on average}, as it processes a large text corpus?
This is an important question from a model interpretability standpoint: when studying the behavior of an attention head, a natural baseline can be useful.  For example, when a head attends to a particular token and writes a particular vector, it is useful to know how much of that behavior is typical for those token types and how much is specific to the present context.

In this paper we address this need by developing empirical methods for studying the average-case behavior of attention heads.  We construct \emph{mean-field approximations} of a head's behavior by averaging the attention of the head over a corpus. Importantly, this is a corpus-averaged statistic of the trained model’s behavior, not a statistic of the corpus alone. Its entries summarize how the model routes attention when processing text. The approximation yields a kernel at the \emph{type} level -- a characterization of a head's behavior that is parameterized by pairs of types (the entries in the model's token vocabulary). The kernel compresses a head’s context-dependent routing into a type-level operator that can be estimated in a single corpus pass, without optimization or fitted parameters.

We show three uses for the resulting approximation.  First, it yields a surprisingly accurate, compact description of the evolution of average type representations.  This  description allows a drastic simplification of the study of average type representations.   Considerable work has focused on 
understanding representation geometry  \citep{Karkada2026Symmetry,Nava2026Hierarchical,Engels2025NotAll,Gurnee2024SpaceTime,Park2024Linear}, including studies arguing that representation geometry is important for understanding model behavior and steering \citep{lu2026adversarialconceptsearchpredicting,wurgaft2026manifoldsteeringrevealsshared}.  A simple description of how geometry evolves over the layers of a model can inform that work.  

Second, the mean-field yields a highly accurate description of \emph{model behavior in early training stages}.  We show that mean-field substitution reveals an early training phase in which
mean-field attention can replace the model’s attention on frequent query types with almost no loss cost.
This is followed by a transition where context-specific attention becomes behaviorally important under our intervention, and studying the evolution of the mean field reveals a major reorganization at the same point.

Third, we use \emph{deviation} from mean-field predictions to explore \emph{contextual computation} in the model.  Model behavior that is not well predicted by the mean field includes syntactic processing, positional computation, and, importantly, in-context learning (ICL).  Across controlled induction and few-shot settings, prompts requiring more in-context computation tend to exhibit larger mean-field deviation. Further, when we rank heads by their average mean field deviation, we find 
previously documented ICL heads
clustered at the top of the ranking.  When we use mean-field deviation to study ICL tasks, we find that task-conditioned mean-field deviations are themselves function vectors: selecting heads by deviation magnitude and injecting their residuals yields transfer competitive with established causal-mediation methods, at a fraction of the computational cost.

Thus, when the mean field succeeds in predicting representations, it gives a compressed description of average representation dynamics. When it fails, the structured residual identifies computation that depends on the particular context.   The picture that emerges from our work is that a representation can be thought of as a corpus-average component and a context-supplied component: global structure plus local variation. Mean-field attention describes the global structure, while the residual measures local, contextual innovation.

\section{Related Work} \label{sec:relwork} 
\paragraph{Corpus statistics and representation geometry.}  Much recent work has focused on the importance of understanding representation geometry as a factor in model behavior and steering \citep{lu2026adversarialconceptsearchpredicting,wurgaft2026manifoldsteeringrevealsshared}.
To that end, a number of studies have derived representational geometry directly from corpus
co-occurrence statistics. \citet{Karkada2026Symmetry} argue, theoretically and
empirically, that symmetries in pairwise word co-occurrence determine the
circular and continuous geometries observed in word embeddings and language
model representations, covering months, years and geographic coordinates;
\citet{Nava2026Hierarchical} extend the account to hierarchical geometry,
connecting the spectrum of the co-occurrence matrix to coarse-to-fine
conceptual organization. These provide insights into phenomena described by
\citet{Engels2025NotAll}, who find circular representations of days and months
and give causal evidence that models use them, and by
\citet{Gurnee2024SpaceTime}, who find linear spatial and temporal
representations at multiple scales. \citet{Park2024Linear} and
\citet{Park2025Categorical} give formal accounts of that geometry, including
the role of the inner product and the distinction between a representation and
its linear accessibility, and \citet{Jiang2024Origins} derive the emergence of
linear semantic structure from a latent-variable model of next-token
prediction.   

Those accounts relate corpus
statistics to a \emph{static} geometry, and papers deriving representations typically do so in simple models or abstractions of models. The mean field studied here  describes how representation geometry is \emph{transformed} layer by layer in real language models. At the same time, our paper documents the strong correlation between the corpus-level attention mean-field and co-occurrence statistics.  We argue that this helps explain why, in real models, co-occurrence appears to have a strong impact on representation geometry.

\paragraph{Distributional semantics.}
The premise that learned geometry can reflect low-order corpus statistics
predates transformers. \citet{Pennington2014GloVe} learn embeddings directly
from global co-occurrence counts, \citet{Levy2014Implicit} show that skip-gram
with negative sampling implicitly factorises a shifted PMI matrix, and
\citet{Arora2016Latent} derive embedding geometry from a latent discourse
process.  \citet{Chan2022Distributional} show that distributional properties of
the data --- burstiness, rare classes, Zipfian frequency --- govern the balance
between in-context and in-weights learning, and \citet{Park2025Competition}
analyze competition among unigram, bigram, retrieval and inference strategies.
This body of work motivates the hypothesis that a low-order corpus-average operator
should capture a substantial share of model behavior. Our paper makes this hypothesis precise and
then tests it directly, finding it borne out for centroid geometry and for early
training, and informatively violated for in-context learning.

\paragraph{Training Dynamics.}
\citet{Chang2022WordAcquisition} find that early
predictions resemble unigram frequencies, then bigram probabilities, becoming
contextually nuanced only later, and \citet{Choshen2022Grammar} report
consistent orders of acquisition across architectures, initializations and
datasets.  Further, \citet{im2026associate} show analytically that early stages of learning in transformer models are driven primarily by corpus-average statistics.  Our results are consistent with these papers, but go further by demonstrating that in the early stages of model training, a model's attention mechanism is \emph{fully replaceable} by its mean field without affecting model performance.

\paragraph{Attention heads, corpus statistics, and contextual algorithms.}
\citet{Elhage2021Framework} analyze attention-only transformers in closed
form, decomposing a head into a QK circuit that determines where attention is
sent and an OV circuit that determines what is written when it arrives. Their
analysis documents that this machinery can respond in ways that reflect
corpus-average bigram and skip-trigram statistics --- that attention is, among
other things, a means by which a model can act on the aggregate statistics of
its training distribution. The present work is consistent with that observation
and goes beyond it. We show that corpus-average statistics are
sufficient to characterize substantial aspects of model behavior: the
layerwise evolution of type centroids, and the trajectory a model follows early
in training. Further, what the mean-field fails to capture is
also informative, and we use its residual to \emph{separate} the aspects of
behavior that a second-order corpus average cannot in principle supply --- the
prompt-specific computation underlying in-context learning. 
Like our paper, \citet{Vig2019Attention} analyze average attention behavior at corpus scale.  However, their focus is on the relationship between average attention and
syntax, dependency relations, distance and depth. Finally,
\citet{ethayarajh2019contextual} measures how far representations vary across
contexts, showing that contextuality increases in upper layers --- a property
that our mean-field-versus-residual decomposition allows one to quantify.

\paragraph{Input-independent and fixed attention.}
One of the conclusions of our work is that certain aspects of model behavior are relatively 
unchanged if the attention mechanism is replaced by a fixed token-level kernel.
Previous work has explored this notion in other contexts.
\citet{Tay2021Synthesizer} replace token--token attention with synthetic and
even random alignment matrices and find them competitive, concluding that
learning attention from query--key interactions is ``useful but not that
important''; \citet{Raganato2020Fixed} replace all but one attention head per
encoder layer with fixed, non-learnable positional patterns without loss of
translation quality; and \citet{You2020HardCoded} replace learned
self-attention with fixed, input-agnostic Gaussians at minimal cost, while
finding cross-attention less dispensable. Taken together these results
establish that a large part of attention's input-dependence is dispensable for
task performance, which can be seen to motivate our work.  In other words, if a model
can be trained to work with attention that never looks at its input, it is
reasonable to ask how much of an existing model's behavior a fixed,
corpus-averaged operator already accounts for.  We additionally go beyond those papers
by pursuing the question not at the coarse level of overall loss, but at the finer grain of 
individual layerwise representations.

\paragraph{Mean-field theories of attention.}
The term ``mean field'' is used in several unrelated senses in the literature. 
\citet{Kim2024Nonconvex} take
a mean-field limit over a distribution of model \emph{parameters} to study
optimization dynamics. \citet{PocLopez2024Dynamical} use dynamical mean-field
theory to characterize expected self-attention dynamics,
 and \citet{Dohmatob2025Softmax} gives an exact mean-field analysis of softmax
attention on a toy problem. 
Our approach has some conceptual similarities with \citet{Burger2025Analysis}, \citet{Castin2025Unified},
\citet{Geshkovski2025Mathematical}
and \citet{Rigollet2025MeanField}.  Like those papers, we treat tokens as interacting particles driven by 
a mean-field abstraction of the attention mechanism. However, those papers are concerned with deriving
continuum or infinite-token and infinite-depth limits, while our work is explicitly based on finite-sample statistics.  
To our knowledge, ours is the first mean-field treatment measured directly
on trained language models and validated against their behavior.

In other words, previous uses of attention mean-fields study limits in the statistical-physics sense,
and they illuminate what attention does asymptotically. The sense used here is
a different one: our mean field
is an expectation over a \emph{corpus distribution}, taken at fixed finite
width, depth and token count with the model's weights frozen. Nothing is taken
to a limit; the averaging is over contexts drawn from text.

\paragraph{Rogue dimensions, attention sinks, and anisotropy.} In studying
the geometry of language model representations, there are certain phenomena that must be 
taken into account.  These generally relate to 
degenerate directions in transformer representations.
\citet{timkey2021rogue} show that a few dimensions dominate similarity
measures while contributing little to behavior; together with the anisotropy documented by
\citet{ethayarajh2019contextual} and the common-direction analyses of
\citet{gao2019degeneration} and \citet{mu2018allbutthetop}, this is why every
metric we report is computed \emph{after centering} to remove these effects.

Further, \citet{sun2024massive} document
``massive activations'' that are approximately context-invariant and act as an
implicit bias rather than carrying token content.  
Related outlier phenomena are
reported by \citet{kovaleva2021bert}, \citet{puccetti2022outliers} and
\citet{dettmers2022int8}. Finally, \citet{xiao2024streaming} and
\citet{gu2025sink} describe attention sinks --- excess probability mass
deposited on a few uninformative positions, canonically the first --- while
\citet{bondarenko2023quantizable} and \citet{FrancoCrovella:Neurips25} connect that ``do-nothing'' routing to the
outlier activations themselves.

Taken together, it becomes clear that these rogue dimensions are not conceptually part of the
context-dependent geometry we seek to predict.  We do not attempt to correct for these in the results reported in the paper body,  but in Appendix~\ref{app:rogue}, we analyze them and show that omitting them 
from analysis improves the predictions of our method.

\paragraph{In-context learning.}
Accounts of in-context learning include induction heads
\citep{olsson2022induction}, implicit gradient descent
\citep{VonOswald2023Gradient}, preconditioned gradient descent
\citep{Ahn2023Preconditioned}, general statistical algorithms with algorithm
selection \citep{Bai2023Statisticians}, and compressed task representations
\citep{Hendel2023TaskVectors}. We do not offer a competing theory of what
in-context learning implements.
The mean field is used instead as a
diagnostic: it quantifies when a prediction requires prompt-specific
computation rather than a corpus-average operator, and is therefore compatible
with any of these accounts of what fills that gap.

\section{Mean Field Analysis of Attention} \label{sec:mf-analysis} 
\subsection{Notation}
\label{sec:notation}

\begin{figure}[t]
  \centering
  \includegraphics[width=0.78\linewidth]{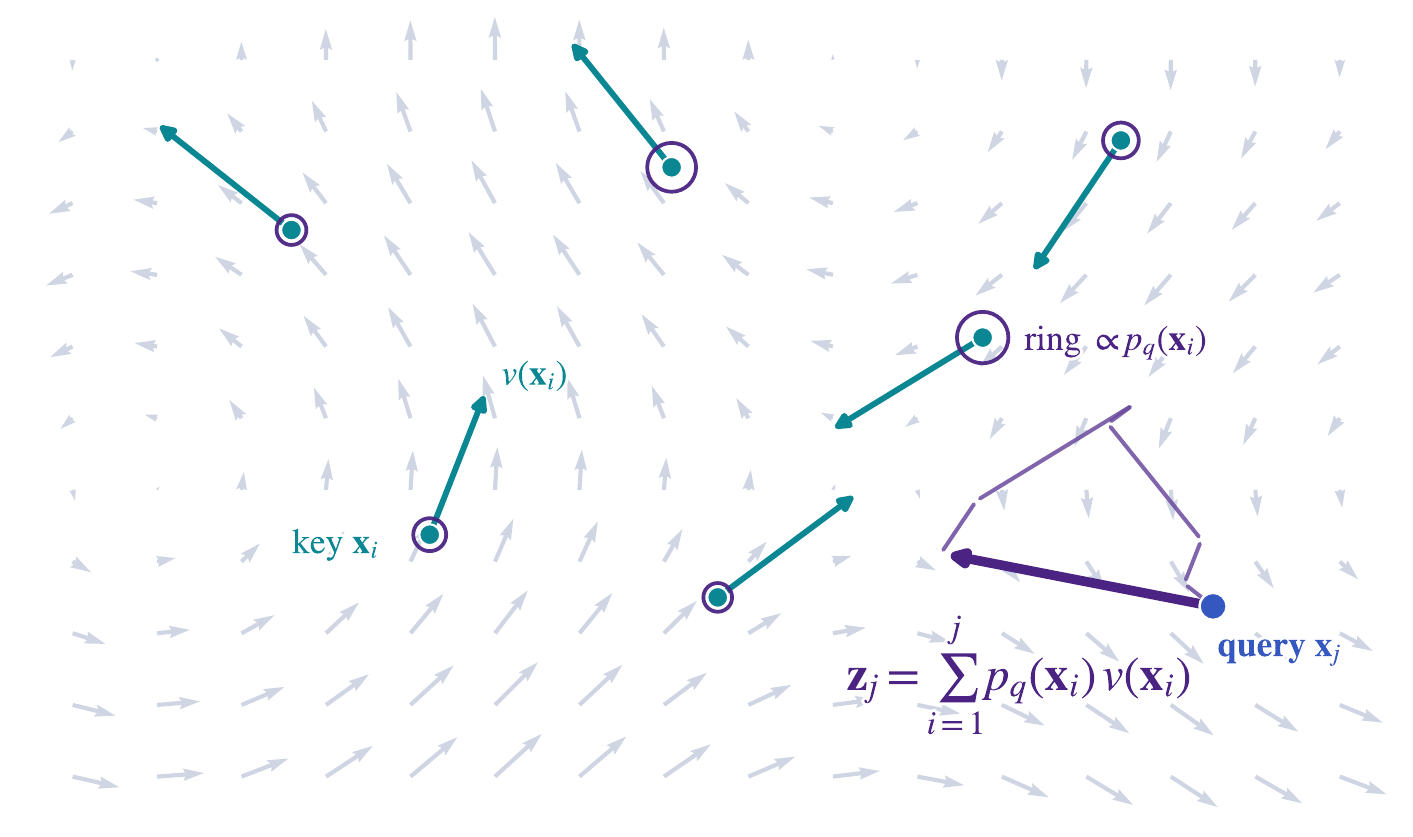}
  \caption{A head's contribution on a single context
  (Equation~\ref{eq:head-contribution}). The value map $v(\cdot)$ assigns a
  vector to every point of representation space (light-gray field); each key
  $\x_i$ (teal) samples the field at its own position. The attention weights
  $p_q(\x_i)$ (purple rings, size proportional to weight) set how much of each
  sampled value the head moves: the contribution $\v z_j$ (heavy purple arrow)
  is the attention-weighted sum of the value vectors, shown assembled
  term-by-term (thin purple path) at the query $\x_j$.
  }
  \label{fig:head-contribution}
\end{figure}

In the attention mechanism, a head at any layer operates on an input%
$\{\x_i\}_{i=1}^m$, with $\x_i \in \mathbb{R}^D$ 
the head's \emph{representation} of the $i$th input token: a running vector that the head reads and updates.
The head computes a separate update for each position in the input, so we say that each position defines one \emph{context}, denoted $X$.  The context at position $j\in\{1,\dots,m\}$ is the prefix $X = \x_{1:j}$ of the input; the attention mechanism uses the representation $\x_j$ as the \emph{query} and the representations $\x_{1:j}$ as the \emph{keys} for that context.  
The mechanism computes a distribution over the
keys, $p_q(\x_i)\ge 0$ with $\sum_{i=1}^{j} p_q(\x_i)=1$, and reads a
\emph{value} $v(\x_i)\in\mathbb{R}^D$ from each key $\x_i$.
The vector the head adds to $\x_j$, the head's
\emph{contribution} to $\x_j$, is\footnote{
The mean-field approach does not depend on the specifics of how $p()$ and $v()$ are realized. As an example, in a typical implementation, tokens are first passed through a layer norm, giving $\tilde{\x} = \text{LN}(\x)$, and then $p()$ and $v()$ are computed as $p(\tilde{X}) = \softmax(\{\tilde{\x}_j^\top W_Q^\top W_K\tilde{\x}_i\}_{i=1}^j)$ and $v(\tilde{\x})= W_OW_V\tilde{\x}$.  Different models will vary in these details, and we consider models with different implementations of $p()$ and $v()$ in our experiments.}
\begin{equation}
\v z_j \;=\; \sum_{i=1}^{j} p_q(\x_i)\,v(\x_i).
\label{eq:head-contribution}
\end{equation}
Figure~\ref{fig:head-contribution} illustrates this contribution geometrically: the value map defines a vector field over representation space, and the head delivers to the query an attention-weighted sample of that field.

Each of the model's input tokens has a \emph{token type}, its entry in the model's \emph{vocabulary}
$\mathcal{T}$. The vocabulary $\mathcal{T}$ is the set of all possible input tokens ---
such as `October', or `was'.
Even though representations are not in the model vocabulary, they still naturally associate with a token type.
Since $X$ arises from a specific forward pass, each
$\x_i$ traces back to the model's $i$th input token, and
$\x_i$ inherits that token's type.
For instance, given the input ``October was cold,'' the position-1
representation has token type `October' at every layer, but its
representation vector differs from layer to layer.  Every other
\emph{occurrence}
of `October' in the corpus, each token of this type at its own position and context, shares that type but has its own
representation vector.
We write $T(\x)$ for the token type of $\x$, and
shorten \emph{token type} to \emph{type} where no ambiguity arises.

Our analysis will focus primarily on the contribution $\v{z}$ that a given head makes to a 
token's representation.  
We will also
consider averages over various conditioning sets; we always denote an
average by a bar with the conditioning set in the subscript, e.g., $\overline{\v z}_s$ is the average contribution given
query type $s$. The same convention names other averaged quantities, such
as $\overline{\v v}_{tX}$, the average value over type-$t$ keys in a context
$X$. A corpus-level average carries only type subscripts, while a
context-level average also carries the context $X$. Our mean field estimates of $\overline{\v{z}}$ will always be denoted $\hat{\v  z}$.
Finally, when the particular head $(\ell, h)$ matters, we will note the layer and the head in the superscript, e.g., as $\v{z}^{\ell h}$.

The contribution $\v{z}$ (Equation \ref{eq:head-contribution}) is what a single head computes
on a single context.  The mean-field approach asks what this contribution
looks like \emph{on average over a corpus}: not for one context, but averaged over a set
of contexts that share some structure.  Different choices of that set give
different mean fields, and we mainly develop 
three here.  The first averages over contexts holding only the query type fixed, and yields a
description of how the model attends on average for queries of a given type.
 We refer to this as the \emph{corpus mean field} (\S \ref{sec:meanfield}). 
The second  averages over just the contexts
that share a pair of token types: the query and the key, and we refer to this as the \stcond mean field.
 The third 
 conditions on an individual context, 
yielding an average description 
fine enough to predict what the model does on a specific input. We call this the \emph{context-conditional mean field}
(\S\ref{sec:context-mf}).  Each applies  the same principle over
different conditioning sets, and we show that they are each useful for a distinct set of analyses.
\subsection{The Corpus Mean Field}

\label{sec:meanfield}
 
A corpus is a collection of inputs (prompts),
each of which passes through the model independently.
Because many
positions across the corpus
may share the same type,
each type is typically realized by
many representations at every layer.
At a given layer, the type's
\emph{centroid}, 
$
  {\v{c}}_{s} = 
  \mathbb{E}\!\left[\x \mid T(\x) = s\right],
$ 
is the mean
representation 
over all 
positions of type $s$ across the inputs of the corpus. Our first mean field, defined in this section, 
is concerned with how each head updates the centroid for type $s$.
This is achieved by considering all contexts whose query type is $s$, ie, $T(q) = T(\v{x}_j) = s$,
and applying $\eqref{eq:head-contribution}$.
Averaging over the corpus, the amount the head adds to the type-$s$
centroid is
\begin{equation}
  \overline{\v{z}}_s = \mathbb{E}_s\!\left[\sum_{i=1}^j p_q(\x_i)\,v(\x_i)\right],
  \label{eq:exact-contribution}
\end{equation}
where $\mathbb{E}_s$ denotes the average over all corpus contexts whose
query is type $s$, ie, $\mathbb{E}_s \stackrel{\text{def}}{=} \mathbb{E}_{\{X: T(\v{x}_j) =s\}}.$

\sloppypar
Attention is per-position, but several positions may share a type, so we can
rewrite \eqref{eq:exact-contribution} as a sum
over \emph{token types} rather than positions, pooling positions of the same type.
For a
query of type $s$ and a context $X$, let
 $A_{st} = \sum_{i:\,T(\x_i) = t} p_q(\x_i)$, the total attention placed on 
positions
of type $t$ (defined to be 0 when a token type is absent), 
and let
\[
  \overline{\v{v}}_t = \frac{\sum_{i:\,T(\x_i)=t} p_q(\x_i)\,v(\x_i)}
                            {\sum_{i:\,T(\x_i)=t} p_q(\x_i)}
\]
be the attention-weighted mean of the value vectors $v(\x_i)$ over those positions.
Then the contribution the head makes to the type-$s$ centroid is, by
linearity of expectation,
\begin{equation}
\overline{\v{z}}_{s} = \mathbb{E}_s \left[\sum_{t\in \mathcal{T}} A_{st} \overline{\v{v}}_t\right]
= \sum_{t\in \mathcal{T}} \mathbb{E}_s\!\left[ A_{st} \overline{\v{v}}_t\right] = 
\sum_{t\in \mathcal{T}} \left[ \mathbb{E}_s[A_{st}]\,\mathbb{E}_s[\overline{\v{v}}_t] +
\operatorname{Cov}_s(A_{st},\overline{\v{v}}_t)\right].
\label{eq:exact-contribution-by-type}
\end{equation}

\paragraph{Mean-field approximation.}
To construct
a mean-field estimate $\hat{\v z}_s$ for $\overline{\v z}_s$, we introduce two \emph{mean-field assumptions.}  We emphasize that we do not assert that these assumptions hold \emph{in general}; rather, they are useful in analyzing the behavior of the model's centroids specifically.  Additionally, our subsequent analyses (\S\ref{sec:waypoint} and \S\ref{sec:icl}) will concern when mean-field assumptions \emph{do not} hold.

\begin{mfass}{Attention-Value independence}\label{ass:value-independence}
$\operatorname{Cov}_s(A_{st}, \overline{\v{v}}_t) \approx 0$.
This assumption posits that attention is approximately uncorrelated with the \emph{value} a head moves. Note, however, that attention may still correlate with the type's \emph{representation}.
\end{mfass}

\begin{mfass}{Negligible curvature}\label{ass:curvature}
$\mathbb{E}_s[\overline{\v{v}}_t] \approx v({\v{c}_{t}}).$
This assumption posits that averaging the value map over a type's occurrences is approximately the same as applying it once at the type's centroid, as it would be if the value map were linear.
\end{mfass}

Assumption~\ref{ass:curvature} involves two subtleties.  First, the sets over which the expectation is taken differ on
  the two sides of the equation: $\mathbb{E}_s[\overline{\v{v}}_t]$ is the mean 
  type-$t$ value across type-$s$ contexts, whereas 
  $\v{c}_{t}$
  is type
  $t$'s global centroid, based on all occurrences of type $t$.  Second, typical
  value maps are nonlinear (e.g., due to the layer norm), so the
  approximation does not hold exactly. The discrepancy due to nonlinearity is the value map's \emph{Jensen gap.}%
  \footnote{The Jensen gap is a general phenomenon: for any nonlinear map, it is the difference
between applying the map at an average and averaging the map's outputs.  The same
gap arises later for the MLP (\S\ref{sec:centroid-rollout}).} 
We evaluate the impact of the value map's Jensen gap, and show that it is small, in Appendix~\ref{app:clustering}.

Because of the central role of $\mathbb{E}_s [A_{st}]$, we refer to it as the
(corpus) \emph{attention mean field} and denote it by $P$, with entries
\begin{equation}
\label{eq:P}
P_{st} = \mathbb{E}_s [A_{st}],
\end{equation}
the mean attention a type-$s$ query places on type $t$ across the
corpus.

\begin{figure}[t]
  \centering
  \includegraphics[width=0.78\linewidth]{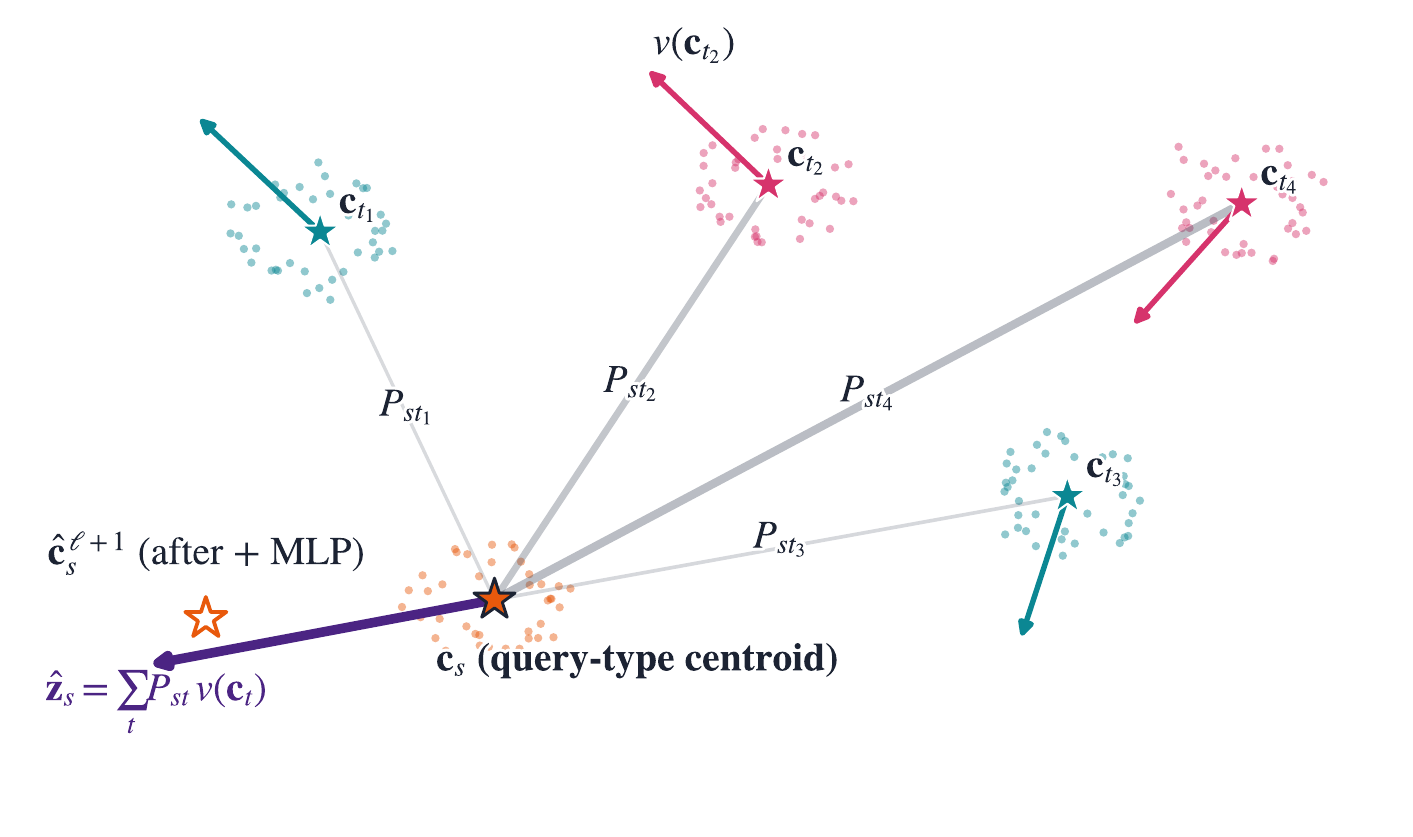}
  \caption{
  The corpus mean-field update
  (Equation~\ref{eq:mean-field}). Each token type $t$ has a centroid $\v c_t$
  (filled stars) surrounded by its cloud of corpus occurrences (small dots).
  The mean field replaces per-context attention with the fixed kernel $P_{st}$
  (gray edges, width proportional to the mean attention that a type-$s$ query
  places on type $t$): the estimate $\hat{\v z}_s$ (heavy purple arrow) applies
  the value map once at each centroid, $v(\v c_t)$ (colored arrows), and
  combines the results with weights $P_{st}$. Summing the head contributions
  and applying the layer's MLP yields the next-layer centroid estimate
  $\hat{\v c}^{\,\ell+1}_s$ (open star); this is the update that the rollout of
  \S\ref{sec:centroid-rollout} iterates across layers.}
  \label{fig:meanfield-update}
\end{figure}

Applying this substitution and the two assumptions to
\eqref{eq:exact-contribution-by-type} yields the head's \emph{mean-field
contribution} 
$\hat{\v z}_s$, an approximation to the true averaged contribution
$\overline{\v z}_s$:

\begin{equation}
    \hat{\v z}_s = \sum_{t\in \mathcal{T}} P_{st} 
    \,v(
    {\v{c}_{t}})
   \;\approx\; \overline{\v z}_s .
    \label{eq:mean-field}
\end{equation}
Figure~\ref{fig:meanfield-update} illustrates the update: the per-context attention of Figure~\ref{fig:head-contribution} is replaced by the fixed kernel $P$, applied to the value map at the type centroids.

Equation \eqref{eq:mean-field}
represents a drastic simplification of the model's processing.  Rather than context-specific attention applied to each representation over the entire corpus, the mean-field estimate applies
a single constant linear map $P_{st}$ and the value maps of the type centroids $v({\v{c}}_{t})$ 
to derive $\hat{\v z}_s$. 
This estimate only requires precomputing $P$ once from a corpus. 
 
Each row of $P$ is an average of probability distributions, so $P$ is
row-stochastic: 
the corpus-conditional mean field is itself a valid attention pattern, one we can use in a replacement model.
This is
the form the rollout in \S\ref{sec:centroid-rollout} applies at each
head and layer.

\subsection{Conditional Mean Fields}
\label{sec:context-mf}

Next we introduce two more narrowly-conditioned mean fields: the \stcond mean field and the context-conditional mean field.  The \stcond mean field is of independent interest, and we use it to introduce certain key concepts.  Then we move to the context-conditional mean field which is the basis for the results in \S\ref{sec:waypoint} and \S\ref{sec:icl}.

For the purposes of exposition, in the remainder of this subsection, we make the simplifying assumption that a given type $t$ can only occur \emph{at most once} in a context.  Making this assumption considerably simplifies the presentation compared to the case where a type can occur multiple times in a context.  We give full derivations that remove this assumption in Appendix~\ref{app:full-conditional-meanfield}.

\paragraph{The \stcond mean field.}  We first show that the contribution to \emph{each} 
centroid (as a query) from each token type (as a key)
can be approximated under a mean-field assumption.

Start from the exact contribution to the centroid \eqref{eq:exact-contribution-by-type}:
\[
\overline{\v z}_s = \sum_{t\in\mathcal{T}} \mathbb{E}_s[A_{st}\overline{\v v}_t]
\]

We now condition on a type $t$ being in the context.  We use $\mathbb{E}_{st}$ to denote the conditional expectation over contexts having $s$ as the query type and in which type $t$ is present.  We use $\Pr_s(t)$ to denote the probability that $t$ is in a context chosen uniformly at random having query type $s$.%
\footnote{Note the corner case for $t=s$:  $\Pr_s(s) = 1$.} %
We then have

\begin{equation}
\mathbb{E}_s[A_{st}\overline{\v v}_t] = \mathbb{E}_{st}[A_{st}\overline{\v v}_t] \,{\Pr}_s(t).
\label{eq:st-split}
\end{equation}
This follows because when type $t$ is not in the context, $A_{st} = 0$.

Denote $\mathbb{E}_{st}[A_{st}\overline{\v v}_t]$ as $\overline{\v z}_{st}$.  This is the average contribution to the $s$ type centroid by type $t$ when $t$ is in the context.  This is a natural measure of how each type influences type $s$ when it appears in $s$'s context.
Then 
\begin{equation}
\overline{\v z}_s = \sum_{t\in\mathcal{T}} \overline{\v z}_{st} \,{\Pr}_s(t)
\label{eq:st-decomp}
\end{equation}
This is an exact decomposition of the corpus contribution in terms of $t$-conditional contributions. We can rewrite $\overline{\v z}_{st}$ as:

\[
\overline{\v z}_{st} = \mathbb{E}_{st}[A_{st}\overline{\v v}_t] = \mathbb{E}_{st}[A_{st}]\,\mathbb{E}_{st}[\overline{\v v}_t] + \operatorname{Cov}_{st}(A_{st}, \overline{\v v}_t)
\]
Now we introduce two mean-field assumptions: first, the \stcond analog of Assumption~\ref{ass:value-independence}, and second, that the average value mapped by a key of type $t$ is independent of $s$:

\begin{mfass}{Attention-value independence, \stcond}
\label{ass:value-independence-st}
This posits that $\operatorname{Cov}_{st}(A_{st}, \overline{\v v}_t) \approx 0$, now on the finer conditioning set of contexts having a query of type $s$ and type $t$ among the keys.  
\end{mfass}

Next, define $\boldsymbol\mu_t$ to be
 the corpus average value map of type $t$ acting as a key, ie, $\boldsymbol\mu_t = \mathbb{E}\!\left[v(\x_i) \mid T(\x_i) = t\right]$.
This is computed directly from the corpus by the same recipe as the centroid ${\v c}_t$: condition on type $t$ and average over its corpus positions, here averaging the frozen value map's output $v(\x)$ rather than the representation $\x$.\footnote{
$\boldsymbol\mu_t$ is the measured average of the value outputs, as
against the mapped centroid $v(\v c_t)$ of Assumption~\ref{ass:curvature}; both
quantities are used in the sections that follow.  Defining $\mu_t$
also makes that assumption's Jensen gap precise, as the difference
$\boldsymbol\mu_t - v(\v c_t)$.}

\begin{mfass}{Value-query independence}
\label{ass:value-query-indep}
This assumption posits that 
$ \mathbb{E}_{st}[\overline{\v v}_t]
\approx
\boldsymbol\mu_t$, 
i.e., conditioning on the query type $s$
has little effect on the average value transmitted by type $t$ keys.
\end{mfass}

As in the corpus mean field case, we
define the \stcond\ mean field
\begin{equation}
W_{st} = \mathbb{E}_{st}[A_{st}],
\label{eq:W}
\end{equation}
and note that \eqref{eq:st-split} with $A_{st}$ alone in place of $A_{st}\overline{\v v}_t$ gives $P_{st} = W_{st}\,\Pr_s(t)$.  So $W$ and $P$ are easily computed from one
another. 
We now apply Assumptions~\ref{ass:value-independence-st} and
\ref{ass:value-query-indep}
to give our mean-field approximation to the contribution to the centroid of $s$,
as a per-$t$ contribution, and show that it decomposes $\hat{\v{z}}$:
\[
\hat{\v z}_{st} = W_{st}\boldsymbol\mu_t
\qquad\Longrightarrow\qquad
\hat{\v z}_s = \sum_{t\in\mathcal{T}} \hat{\v z}_{st}\,{\Pr}_s(t) \approx \overline{\v z}_s
\]
Substituting this back also confirms that the 
\stcond\ mean field is consistent with the corpus mean field it refines: $\hat{\v z}_s = \sum_{t\in\mathcal{T}} P_{st}\boldsymbol\mu_t$, which is \eqref{eq:mean-field} with $\boldsymbol\mu_t$ in place of $v(
{\v{c}_{t}})$.
The two estimators differ by  $\sum_{t}P_{st}\bigl(\boldsymbol\mu_t - v(\v c_t)\bigr)$, the value map's Jensen gap of each key type weighted by the kernel, so they coincide when that gap is zero. In our experiments the gap is small; we document the small Jensen gap of the value map and its small effect on predictions in Appendix~\ref{app:clustering}. So we find that the two constructions agree in practice.

\paragraph{The context-conditional mean field.}
Next, we use a similar strategy to make an average case prediction for a particular context $X$.  We assume that $X$ has query type $s$.
We denote the contribution made to query type $s$ when it has context $X$ as ${\v z}_{sX}$, the value map from type $t$ in that context as $\boldsymbol{\mu}_{tX}$, the attention paid by $s$ to $t$ conditioned on the context being $X$ as $W_{stX}$, and the analog of $\Pr_s(t)$ for context $X$ as $\Pr_X(t)$.  Two of these quantities degenerate because $X$ is a single fixed context:
the average $W_{stX}$ is  the actual attention $A_{stX}$ the head pays from $s$ to $t$ in $X$, and $\Pr_X(t)$ is 1 for the types in $X$ and 0 otherwise.

As in the \stcond decomposition \eqref{eq:st-decomp}, these quantities give an exact per-$t$ decomposition of the contribution, now for the single context $X$:
\begin{equation}
{\v z}_{sX} = \sum_{t\in\mathcal{T}} W_{stX}\,\boldsymbol{\mu}_{tX}\,{\Pr}_X(t)
\, =
\sum_{t\in X} A_{stX}\,\boldsymbol{\mu}_{tX}
\label{eq:ccmf-exact}
\end{equation}

To construct a mean field analog, we introduce two mean-field assumptions.

\begin{mfass}{Value-context independence}
\label{ass:value-context-indep}
This posits that $\boldsymbol{\mu}_{tX} \, \approx \, \boldsymbol{\mu}_t$. That is, the value map %
of type $t$ in context $X$ is well-approximated by
its corpus average.  This is an 
important 
simplification; the difference between $\boldsymbol{\mu}_{tX}$ and $\boldsymbol{\mu}_t$ can be thought of as the \emph{contextualization} of the value map of $t$ induced by its context $X$.
\end{mfass}

\begin{mfass}{Proportionality of context attention}
\label{ass:proportionality}
This assumption posits that for each $s$, the actual attention values $A_{stX}$ are proportional to $W_{st}$ across the types present in $X$.  The constant of proportionality is then fixed by the requirement that a head's attention over its context sums to one.
\end{mfass}

Assumption~\ref{ass:proportionality} lets our 
mean-field kernel $\hat{A}_{stX}$ estimate the real attention $ A_{stX}$ by renormalizing $W_{st}$ over the types in $X$
(the general case of repeated types is treated in Appendix~\ref{app:full-conditional-meanfield}):
\begin{equation}
\hat{A}_{stX} = \frac{W_{st}}{\sum_{t'\in X}W_{st'}} \, \approx \, W_{stX} \, = \, A_{stX}.
\label{eq:ctx-kernel}
\end{equation}

 Continuing from \eqref{eq:ccmf-exact}, we substitute the corpus average $\boldsymbol{\mu}_t$ for $\boldsymbol{\mu}_{tX}$ (Assumption~\ref{ass:value-context-indep}) and $\hat{A}_{stX}$ for the attention $A_{stX}$ (Equation~\eqref{eq:ctx-kernel}) to derive our \emph{context-conditional} mean field estimator:
\begin{equation}
\hat{\v{z}}_{sX} = \sum_{t\in X} \hat{A}_{stX}\, \boldsymbol{\mu}_t 
\, \approx \, {\v z}_{sX}.
\label{eq:ccmf}
\end{equation}

\paragraph{Analysis of \Ahat.}
Unlike the corpus mean field and the \stcond\ mean field, the context-conditional mean field computed here is a nonlinear function of model-induced statistics (because of the normalization used in \eqref{eq:ctx-kernel}).  As such, a desirable property would be \emph{unbiasedness}; specifically, we would like that $\E_{st}[A_{stX}-\Ahat_{stX}] = 0.$  We analyze this question in Appendix~\ref{app:ccmf-analysis}, and show: (1) \Ahat\ is not, in general, unbiased by this definition; (2) the bias we observe in \Ahat\ in practice is quite small; and (3) removing the observed bias in \Ahat\ does not change any of our key results.

\section{Mean Field Prediction of Representation Evolution}
\label{sec:centroid-rollout}
As a first application
of mean-field attention analysis, we use the mean field approximation to predict type centroid evolution across the layers of a language model.  The true model modifies the representation of each input token at each layer, 
given its context. By tracking type centroids we can observe the evolution of type representation geometry layer by layer.

Concretely, the mean-field approximation \eqref{eq:mean-field} gives the average contribution to a type's centroid for a single head
from the static mean-field attention
and the frozen weights alone.
Iterating this through the layers, and through the MLP at each layer,
predicts how a type's centroid evolves across the whole network.

\subsection{Method: Open-Loop Rollout}

To construct the model's mean-field replacement, we run a purely open-loop computation,
analogous to a dynamical system, shown in Algorithm~\ref{alg:rollout}.
The process starts with input embeddings (the layer-0 centroids),%
\footnote{In GPT-2, a mean positional offset is added. Initializing GPT-2 with input embeddings rather than measured centroids has negligible effect. Using measured centroids in GPT-2 gives L1 centered cosine prediction of 0.989, while initializing from raw $W_E$ embeddings gives 0.984.}
and then for each layer and each type, it first applies the mean-field
approximation \eqref{eq:mean-field} for each head (line 4), then sums head contributions and adds them to the current centroid (line 5), and then passes the resulting updated centroids through the layer's MLP (line 6).  The resulting centroid approximations are then passed to the next layer, where the process repeats.

\begin{algorithm}
\caption{Open-loop mean-field rollout}
\label{alg:rollout}
\SetKwInOut{Input}{input}
\SetKwInOut{Output}{output}
\Input{layer-0 centroid representations $\{\hat{\v c}^0_s\}$ (input embeddings); mean fields $P^{\ell h}$; frozen weights}
\Output{predicted centroid representations $\{\hat{\v c}^\ell_s\}$ for each type $s$ and layer $\ell$}
\For{each layer $\ell = 0$ \KwTo $L-1$}{
\For{each type $s \in\mathcal{T}$}{
  \For{each head $h$}{
    $\hat{\v z}^{\ell h}_s \gets \sum_{t\in\mathcal{T}} P^{\ell h}_{st}\, v(\hat{\v c}^\ell_t)$ \tcp*{per-head contribution, \eqref{eq:mean-field}}
  }
  $\v m^\ell_s \gets \hat{\v c}^\ell_s + \sum_h \hat{\v z}^{\ell h}_s$ \tcp*{add attention contributions}
  $\hat{\v c}^{\ell+1}_s \gets \v m^\ell_s + \mathrm{MLP}^\ell(\mathrm{LN}(\v m^\ell_s))$ \tcp*{apply MLP to the average}
}
}
\end{algorithm}

The rollout is a self-contained 
computation
of the network's average-case
behavior for each type, built only from mean-field kernel and the frozen weights.
Each layer maps type centroids to type centroids, so one
layer's output is the next layer's input, and the rollout iterates with no
forward pass on any text. Analogous to the mean-field substitution for
attention \eqref{eq:mean-field}, we apply the MLP only once, at the
post-attention average. This is again a drastic simplification of what the
MLP does: one execution at the centroid average, rather than averaging many
executions of the MLP over each specific type-$s$ context. Ultimately, the
centroid trajectory that the rollout produces is a prediction, whose accuracy we
demonstrate in the remainder of this section. \S\ref{sec:nae} attributes the rollout's error among these approximations.

\subsection{Basic Results and Ablations}
\label{sec:validation}

We now evaluate the performance of mean-field prediction of type centroids as obtained via Algorithm~\ref{alg:rollout}.  Our evaluation spans five models: GPT-2 Small \citep{radford2019gpt2}, Gemma-2 2B \citep{gemmateam2024gemma2}, Llama-3.1-8B \citep{grattafiori2024llama3}, Llama-3.2-3B \citep{metaai2024llama32}, and Qwen-3-14B \citep{yang2025qwen3}.
Kernel and centroid statistics
are estimated from
OpenWebText~\citep{gokaslan2019openwebtext}, an open reproduction of GPT-2's
training corpus: text samples of 1.5–2M tokens
are tokenized with each model's tokenizer and packed into
512-token inputs. %
The kernel and per-type statistics 
on the $N = 1000$ most frequent types
are estimated 
as defined in \S\ref{sec:meanfield} (the attention mean field $P$, \eqref{eq:P}, and the type centroids).
The top 1000 tokens cover 
$64$--$68\%$ of corpus positions and $67$--$72\%$ of
non-BOS attention across all five models, a range nearly independent of vocabulary
size: $1000$ types out of Gemma-2's $256{,}000$ cover as much text as $1000$ out of
GPT-2's $50{,}257$ (Appendix~\ref{app:coverage}).
Held-out evaluation uses a document-interleaved
split: alternating documents are assigned to two disjoint halves of the
corpus, the kernel and all corpus statistics are estimated on one half, and
the true centroids used for scoring are measured on the other (GPT-2 and
Gemma-2; the Qwen-3 and Llama results use a single split, as noted where
they appear).  
Appendix~\ref{app:modeling-details} gives the two extra kernel columns, one
for BOS and one pooling every non-tracked type, that keep each row of $P^{\ell h}$
stochastic when only the top $N$ types are tracked.

In Appendix~\ref{app:corpus} we perform additional evaluations on three different corpora: FineWeb \citep{penedo2024fineweb}, WikiText \citep{merity2017pointer}, and Pre-1919 books \citep{rae2020pg19}.  We find that the results in this section are not specific to the OpenWebText corpus, and in fact that results shown here even generalize across corpora when using one corpus for training and a different corpus for testing.

We assess prediction quality with four metrics: centered cosine similarity, RSA, relative Euclidean error, and norm ratio (defined in Appendix~\ref{app:full-mean-field-results}). Each is computed per type and then averaged.
Relative error is computed as a ratio of means (mean Euclidean error
over mean distance to the entire type set's centroid) so that types lying near the centroid do not arbitrarily inflate it. 
All are computed after mean-centering the representation cloud. Centering is necessary because the representations are strongly anisotropic as discussed in Appendix~\ref{app:anisotropy}.  Further, we evaluate the effect of rogue dimensions (aka `massive activations', e.g., \cite{timkey2021rogue,kovaleva2021bert}) on our results in Appendix~\ref{app:rogue}.

\begin{figure}[h]
    \centering
    \includegraphics[width=0.32\linewidth]{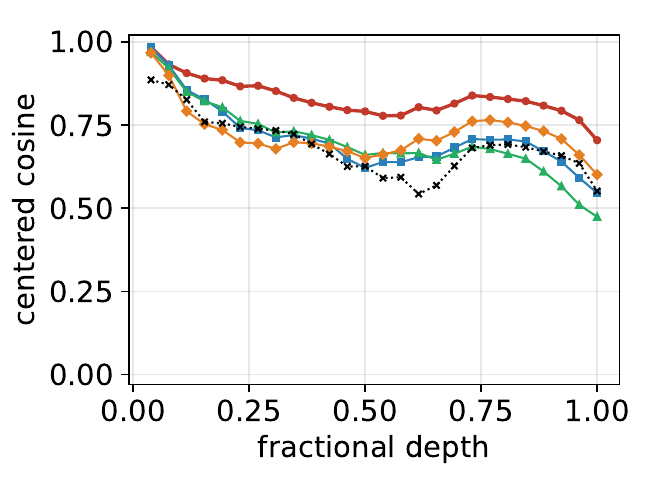}
    \includegraphics[width=0.32\linewidth]{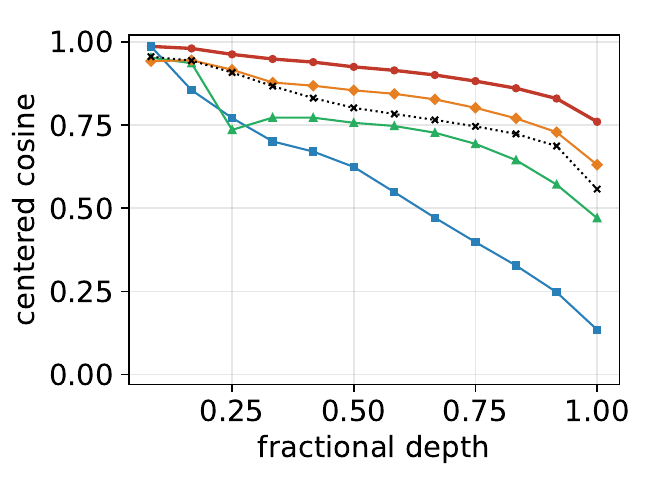}
    \includegraphics[width=0.32\linewidth]{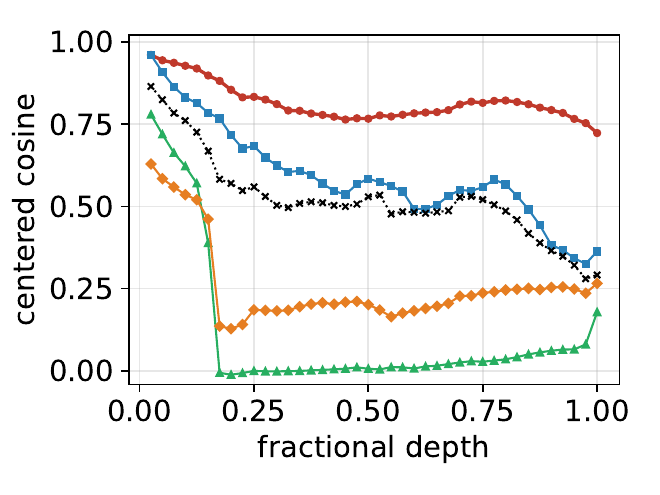}\\
    \includegraphics[width=0.85\linewidth]{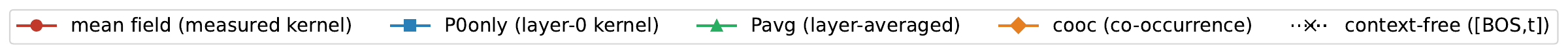}
    \caption{Mean-field predictions are accurate and superior to alternative kernels.  Average per-type centered cosine distance. Left: Gemma-2, Middle: GPT-2, Right: Qwen-3-14B.}%
    \label{fig:mf-vs-alternative-kernels}
\end{figure}

\begin{figure}[h]
    \centering
    \includegraphics[width=0.32\linewidth]{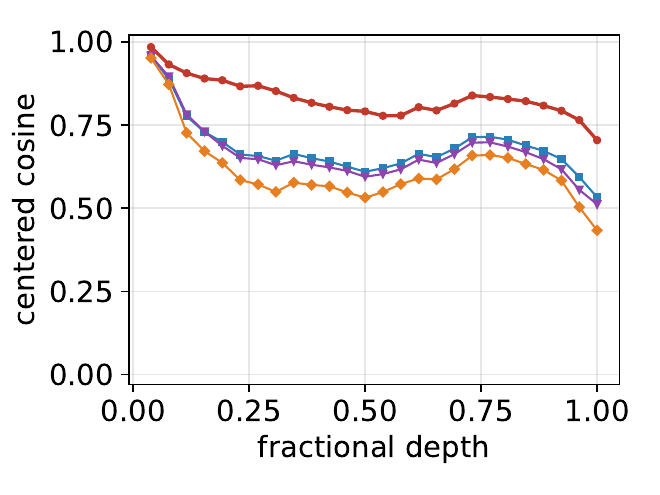}
    \includegraphics[width=0.32\linewidth]{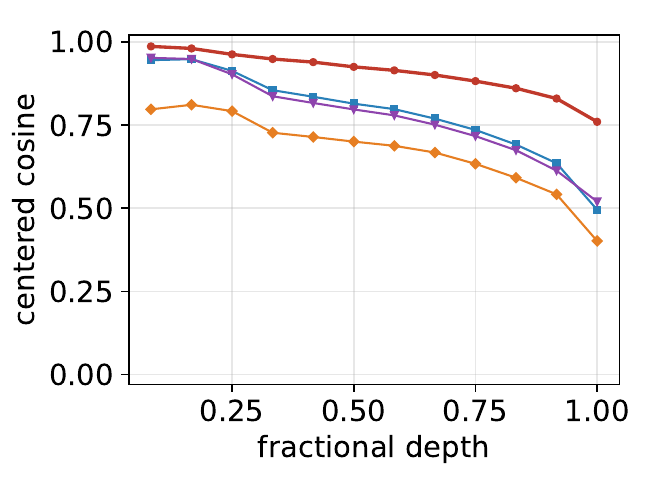}
    \includegraphics[width=0.32\linewidth]{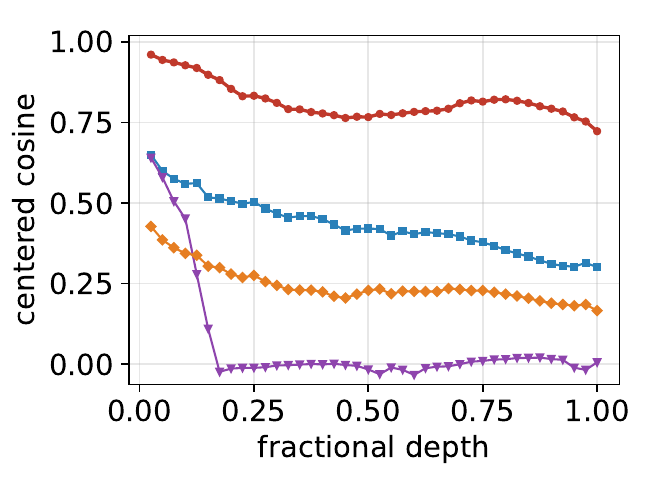}\\
    \includegraphics[width=0.85\linewidth]{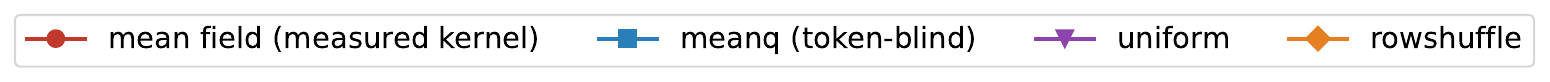}
    \caption{Mean-field predictions are superior to query-blind and key-blind alternatives.  Average per-type centered cosine distance. Left: Gemma-2, Middle: GPT-2, Right: Qwen-3-14B.}
    \label{fig:mf-vs-query-options}
\end{figure}

The results for cosine similarity are shown in Figure~\ref{fig:mf-vs-alternative-kernels} for GPT-2, Gemma-2 and Qwen-3.  Results for these models on the other three metrics are consistent and are given in Appendix~\ref{app:full-mean-field-results}.
The Llama models are the exception, and Appendix~\ref{app:llama} treats them
separately; \S\ref{sec:nae} explains why.
Overall, the figures show that
\emph{mean field predictions of 
type centroids prove highly 
accurate}.
In GPT-2, centered cosine similarity exceeds 0.9 through the middle layers 
and remains above 0.75 even in the last 
layer.  In Qwen-3 and Gemma-2, cosine similarities remain well above 0.75 through most of the network.
Given the high dimension of the representation space, these reflect very close alignment of predictions with actual centroids.%
\footnote{\textbf{Why many of the curves are not monotone.} The rollout is open-loop,
so one expects error to compound and accuracy to fall steadily with depth. However, as the curves in Figures~\ref{fig:mf-vs-alternative-kernels} and \ref{fig:mf-vs-query-options} show, cosine is not always monotone. This is explained by noting that cosine error improves when the prediction's extent along the true direction grows faster than its sideways error. For example, this can happen when the residual stream's magnitude grows faster than the rollout accumulates new error.  As discussed in Appendix \ref{app:anisotropy}, residual stream magnitude typically grows steadily with depth.}

To assess the importance of the mean-field kernel in producing the accurate results here, we compare against four alternatives. Each alternative replaces one part of the
construction with something simpler; their curves appear alongside the mean field in
Figure~\ref{fig:mf-vs-alternative-kernels}.
Our first substitute is  
\texttt{context-free}: a baseline %
that propagates each type through the network in isolation (i.e., running the model on just [BOS, $\v{c}_t$]). The mean field approach improves on this comparison at every layer, and increasingly so with depth.
Second, because the mean field is similar to a corpus-average object, a natural question is whether we need to compute attention at all: can co-occurrence 
statistics over the corpus suffice?  
In the \texttt{cooc} alternative,
the type-to-type distribution $S$ is taken from a symmetric $\pm$ 5-word co-occurrence matrix over the corpus, $C$, normalized to be row-stochastic.  This is used to distribute attention across the centroids, correcting for the fraction of attention that is allocated to the top 1000 tokens.\footnote{We normalize $\tilde{C}_{st} = C_{st}/\sum_{t'}C_{st'}$.  We then form the row sums $T_s = \sum_{t=1}^{1000}P_{st}$ and use as the final attention-substitute $T_s \tilde{C}_{st}.$}
Mean field outperforms \texttt{cooc} as well.
Finally, we consider whether the kernel's depth structure can be simplified.
\texttt{P0only} reuses the \emph{layer-0} kernel at
every layer (head by head) testing whether depth evolution of the kernel matters.
The last alternative, \texttt{Pavg}, replaces it with
the kernel averaged over layers. Both are distinctly worse than the measured mean field in both models, meaning that the kernel's variation across depth carries important content for prediction.

We next ask whether the specific structure of attention mean field $P$ is necessary, or whether a structureless version could do as well.
For example, how important is it to distinguish between query types, as well as between key types?  We study three cases: First, \texttt{meanq} (query-blind attention): for each $(\ell, h)$,
every query row in $P$ is replaced by the \emph{average row}. Every type then reads the same mixture of key
types, so the attention term contributes an identical vector to every
type --- pure drift. Second, \texttt{uniform} (key-blind attention): each query's total type-to-type budget
$\sum_t P_{s,t}$ is preserved but spread evenly over all
key types. Finally, \texttt{rowshuffle} (negative control): for each head,
the query rows of the kernel are permuted randomly
so type $q$ receives the actual attention profile of some
\emph{other} type. This allows us to distinguish ``the graph structure of $P$ is uninformative'' (in which case shuffling would
match \texttt{meanq}) from ``$P$'s graph carries type-specific
information'' (where shuffling should do \emph{worse} than \texttt{meanq}, because
wrong type-specific content is actively injected).
Figure~\ref{fig:mf-vs-query-options} shows that the full structure of $P$ is important:  
none of the structureless approximations match the measured kernel, so
accurate prediction of centroid evolution requires type-level distinctions among both queries and keys,
and the type-resolved structure matters increasingly with depth.

Table~\ref{tab:ablation-summary} reports
the cross-layer mean (standard error over layers) of the centered cosine and relative error (lower is better)
for the mean field,
the context-free baseline, \texttt{cooc} and
the other ablations we consider.
The measured mean field is most accurate in both metrics and all three models. 
However, among the alternatives, co-occurrence (\texttt{cooc}) is closest to matching the mean-field on GPT-2 and Gemma-2.
This is an interesting observation that \emph{simple co-occurrence can reproduce much (but not all) of the effects of mean-field attention on centroid evolution.}

\begin{table}[ht]
\centering
\caption{Cross-layer fidelity of the mean-field rollout, the context-free baseline, \texttt{cooc}, and mean-field ablations.}

\label{tab:ablation-summary}
\resizebox{\textwidth}{!}{%
\begin{tabular}{lcccccc}
\toprule
 & \multicolumn{2}{c}{GPT-2} & \multicolumn{2}{c}{Gemma-2} & \multicolumn{2}{c}{Qwen-3-14B} \\
\cmidrule(lr){2-3}\cmidrule(lr){4-5}\cmidrule(lr){6-7}
 & cosine & rel.\ error & cosine & rel.\ error & cosine & rel.\ error \\
\midrule
Mean field & \textbf{0.907} (0.019) & \textbf{0.434} (0.051) & \textbf{0.830} (0.011) & \textbf{0.593} (0.026) & \textbf{0.816} (0.009) & \textbf{0.640} (0.023) \\
\midrule
Context-free & 0.797 (0.033) & 0.643 (0.063) & 0.686 (0.018) & 0.809 (0.024) & 0.521 (0.021) & 1.183 (0.036) \\
\texttt{cooc} & 0.834 (0.026) & 0.567 (0.042) & 0.724 (0.015) & 0.738 (0.027) & 0.258 (0.020) & \emph{1.429}$^*$ (0.067) \\
\midrule
\texttt{meanq} & 0.786 (0.038) & 0.615 (0.050) & 0.682 (0.017) & 0.786 (0.028) & 0.430 (0.014) & 0.949 (0.004) \\
\texttt{uniform} & 0.776 (0.038) & 0.630 (0.054) & 0.666 (0.018) & 0.809 (0.030) & \textit{0.061} (0.027) & \textit{96.1}$^{*}$ (12.9) \\
\texttt{rowshuffle} & 0.672 (0.034) & 0.754 (0.036) & 0.616 (0.021) & 0.894 (0.035) & 0.245 (0.009) & 1.042 (0.013) \\
\midrule
\texttt{P0only} & 0.561 (0.074) & 1.124 (0.153) & 0.712 (0.019) & 0.747 (0.034) & 0.591 (0.024) & 1.091 (0.055) \\
\texttt{Pavg} & 0.732 (0.039) & 0.830 (0.069) & 0.702 (0.022) & 0.740 (0.031) & \textit{0.114} (0.036) & \textit{24.4}$^{*}$ (3.0) \\
\bottomrule
\end{tabular}}

\vspace{4pt}
\begin{minipage}{\textwidth}
\footnotesize
\textbf{$^{*}$Note on the Qwen-3-14B ablations.} Qwen's \texttt{cooc}, \texttt{uniform} and \texttt{Pavg} entries show large relative errors which are produced by a numerical instability of that model's rollout concentrated at layer~0 and not a reflection of kernel quality; Appendix~\ref{app:rogue} gives the full accounting.
\end{minipage}
\end{table}

\subsection{Understanding Approximation Errors}
\label{sec:nae}

Algorithm~\ref{alg:rollout} incorporates two 
approximations at every layer: the mean-field prediction of each head's
contribution $\hat{\v z}^{\ell h}_s$, and the application of the MLP once at the
resulting centroid.  Only the first is a property of mean-field attention; the
second is a property of the rollout as a way of propagating centroids.  

\begin{figure}[t]
    \centering
    \includegraphics[width=0.78\linewidth]{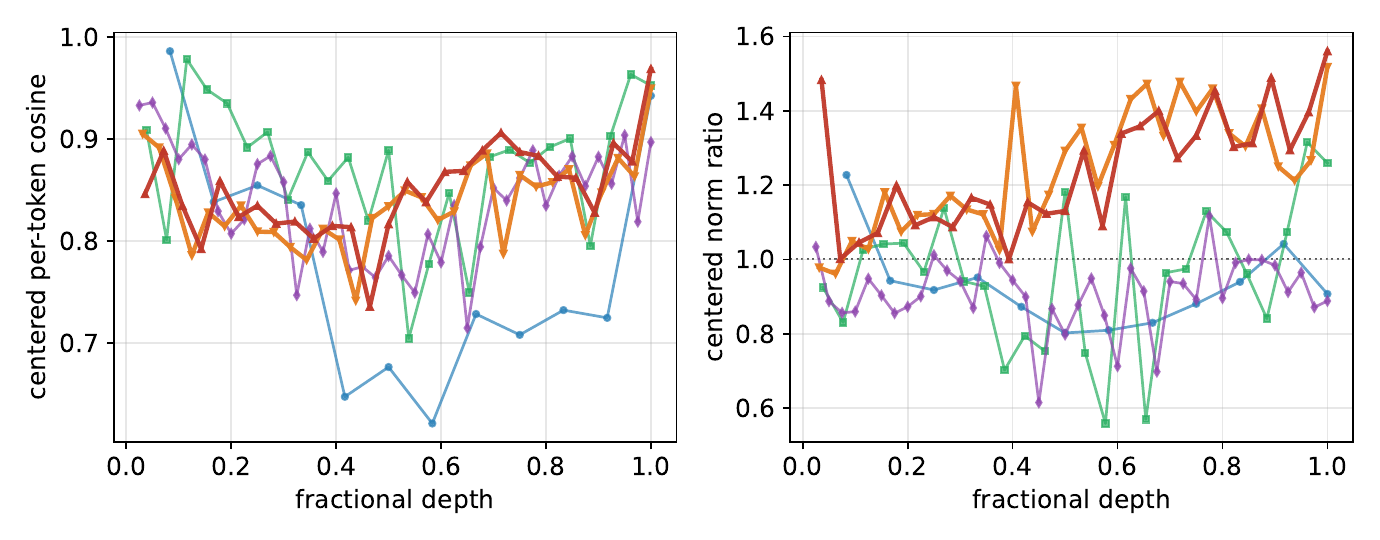}\\
    \includegraphics[width=0.72\linewidth]{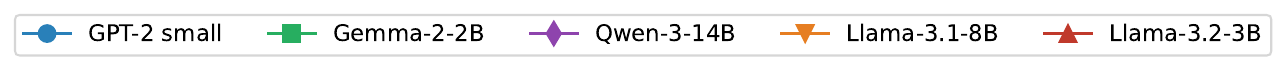}
    \caption{Teacher-forced prediction of the head contribution
    $\overline{\v z}$, by fractional depth.  Left: centered cosine; right: centered norm
    ratio, with the dotted line marking zero error.}
    \label{fig:tf-attention}
\end{figure}

Accordingly, we first ask: how good is the mean-field approximation of attention?  We measure this by comparing $\overline{\v z}$ with its approximation $\hat{\v z}$.
Figure~\ref{fig:tf-attention} does this
teacher-forced: starting from the \emph{true} centroids at each layer, it compares
the predicted attention-block output $\sum_h \hat{\v z}^{\ell h}_s$
against the same quantity measured over the corpus,
$\sum_h \overline{\v z}^{\ell h}_s$, averaging across all types $s$.%
\footnote{Bootstrap CIs over types are less than $\pm$ 0.01 and so not shown.}
The prediction is good at every
layer of every model: across all five models, cross-layer mean centered cosine lies between $0.78$ and
$0.87$, and the norm ratio stays near one (cross-layer means $0.91$ to $1.24$),
so the mean field does well at recovering both the direction and the scale of what each head
writes.

We next decompose the per-layer error using three increasingly approximate estimators of the layer-$\ell{+}1$ centroid,
 measured for GPT-2, Gemma-2, and Qwen-3-14B;
Llama is discussed below.
All three are applied to the true layer-$\ell$ centroids and end with the
MLP applied once at the resulting centroid; each makes one more approximation in
the attention block than the last.
Estimator (a) isolates the MLP gap by using exact attention for the rest of the prediction ($\v z_s$, per \eqref{eq:exact-contribution-by-type}), so that applying the MLP at the centroid is its only approximation.%
\footnote{For Gemma-2, this also includes the post-attention norm at the centroid.}
Estimator (b) adds the further approximation of taking the per-type value at the centroid, i.e., replacing $\mathbb{E}_s[\overline{\v{v}}_t]$ with $v(\v{c}_t)$ (still keeping the attention-covariance term); and estimator (c) applies the full mean field of \eqref{eq:mean-field}, which additionally drops the covariance term.

\begin{figure}[t]
    \centering
    \includegraphics[width=0.95\linewidth]{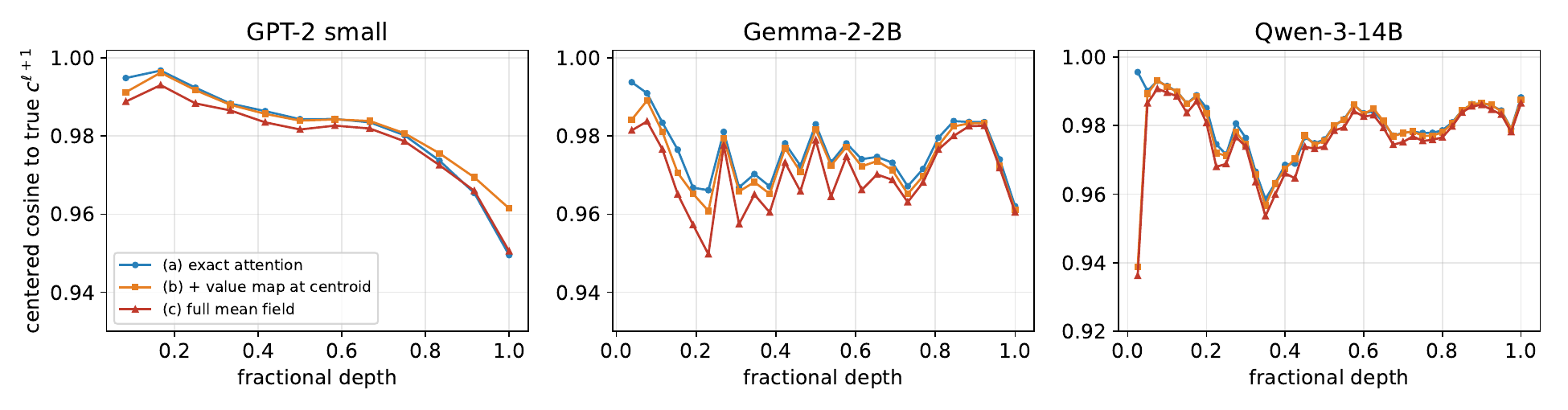}
    \caption{Per-layer accounting of mean-field approximation error. Starting from the true centroids at layer $\ell$, three nested predictions of the layer-$\ell{+}1$ centroid are compared to the true value (centered cosine): (a) exact attention with the MLP applied at the centroid; (b) additionally taking the value map at the centroid; (c) the full mean field of \eqref{eq:mean-field}. Left: GPT-2; middle: Gemma-2; right: Qwen-3-14B.}
    \label{fig:error-decomp}
\end{figure}

The decomposition confirms that MLP-introduced error is the main source of rollout inaccuracy (Figure~\ref{fig:error-decomp}):  the dominant per-layer source of error is the MLP evaluated at the centroid. Estimator~(a) (exact attention at the centroid rather than per occurrence) accounts for almost all of the per-layer error, falling short of perfect prediction by %
approximately $0.02-0.03$ across all models
in centered cosine similarity on average.
The two further approximations then add little:
taking the value map at the centroid ((a)$\to$(b)) and dropping the covariance term ((b)$\to$(c)) 
each cost at most ${\sim}0.005$ per layer. The final error of the full open-loop rollout (Figure~\ref{fig:mf-vs-alternative-kernels}) therefore does not arise from any single step being inaccurate, but from the accumulation of these small per-layer errors across depth, with the MLP-at-centroid approximation the leading per-step contributor.
Measured on its own, the MLP gap is less significant than this may suggest, as we explain in Appendix~\ref{app:mlp-gap}.

The same decomposition explains the two Llama models, whose rollouts predict centroid evolution distinctly less well than the others (full details in Appendix~\ref{app:llama}).  The mean-field \emph{attention} prediction for Llama is as accurate as for the other models (Figure~\ref{fig:tf-attention}); the error is in the MLP-at-centroid step, whose Jensen gap is larger because many of Llama's centroids, especially in early layers, sit where the MLP is more sharply curved.
Almost all of the Llama-3.2-3B rollout error traces to poor MLP approximation at layer~1.
Because this is a property of the MLP rather than of the mean-field attention, Llama's centroid rollout requires a Jensen-gap correction to be accurate.

Finally, we note that the small error introduced via the value map suggests that, in general, representations are comparatively closely clustered around their type means. In Appendix \ref{app:clustering} we confirm this property quantitatively. 
And we note that the covariance term is small in magnitude but not in importance: it contributes little to centroid prediction, yet is precisely what carries context-specific computation (\S\ref{sec:icl}).

\section{The Mean Field as a Training Waypoint} \label{sec:waypoint} 
We now show how mean-field analysis can provide insight into the training dynamics of a model. In this section we show that early in training,
the model's computation \emph{reduces to} its mean field, then deviates as in-context computation arrives.  Analyzing the resulting dynamics shows which changes in model weights drive early-stage learning.

To do so, we run Pythia \citep{biderman2023pythia}, replacing every attention
head's contribution with the checkpoint's context-conditional mean field and measuring the
resulting change in next-token loss.
We perform these measurements across three scales of Pythia (160m, 1.4B, and 12B parameters) at 19 checkpoints each, spanning each model's full training run.   
Because the mean field ignores token order, zero difference in loss would mean the model's context-dependence adds nothing beyond its corpus average.

\emph{Near-zero difference in loss is in fact what we find early in training.}  After the mean-field-equivalent period, the difference rises sharply; the departure from
the mean field and the arrival of in-context computation coincide.
\citet{im2026associate} characterize the early weights as closed-form
compositions of corpus statistics; our mean-field account is the
computational side of the same picture, showing that the young model's
behavior reduces to its corpus mean field until in-context computation
emerges.

\subsection{Replacing Attention with its Mean Field}
\label{sec:replacement}

The experiment described above requires replacing the attention computed by the model for each specific context $X$ with a mean field approximation.  For that purpose we use the context-conditional mean field of \S\ref{sec:context-mf}, Equation \eqref{eq:ccmf}.  Specifically, we substitute $\hat{\v z}^{\ell h}_{sX}$ for each $\v{z}^{\ell h}_{sX}$ computed by the model, leaving the rest of the model's computation unchanged.
We note an important property of $\hat{\v z}_{sX}$: it is unaffected by the ordering of the tokens in the context $X$ ($\hat{\v z}_{sX}$ is `order-free').
If performing the substitution does not change the model's loss, then
what the model was computing was reproducible without considering token order.

We compute $\hat{A}^{\ell h}_{st}$ and $\boldsymbol\mu^\ell_t$ (the ingredients of Equation \ref{eq:ccmf}) from the same corpus as used in \S\ref{sec:validation}. 
On a disjoint held-out sample
(100{,}000 tokens) we then run the model's forward pass with every head's
contribution at every layer replaced by its per-context contribution
$\hat{\v z}^{\ell h}_{sX}$.
The substitution is made for the top 1000 types, for which $W$ can be estimated reliably from a 2M token pass of each model. For Pythia, these 1000 types constitute 65\% of positions in the corpus (Appendix~\ref{app:coverage}); remaining query types keep the model's own computation.  In Appendix \ref{app:waypoint-vocab} we show parallel results for the top $N = 2000$ and $N = 4000$ types (covering 72\% and 80\% of corpus positions, respectively);  the effects in this section are unchanged over this range of $N$.

We report two diagnostics: $\Delta L$ and $\Dmf$.  The paired loss
gap $\Delta L = L_{\mathrm{MF}} - L_{\mathrm{model}}$ measures how much
prediction is contextual: the loss the model
gives up when its attention computation is replaced by the mean field.
 Write
$p(\cdot\mid X)$ for the model's next-token distribution on context $X$, and
$p_{\mathrm{MF}}(\cdot\mid X)$ for the distribution obtained when the attention
output of every head at the query position is replaced by its context-conditional
mean-field contribution \eqref{eq:ccmf}, with the rest of the forward pass left
intact.  Writing $y$ for the token that actually follows $X$,
\begin{equation}
  L_{\mathrm{model}} = -\log p(y\mid X),
  \qquad
  L_{\mathrm{MF}} = -\log p_{\mathrm{MF}}(y\mid X),
  \qquad
  \Delta L = L_{\mathrm{MF}} - L_{\mathrm{model}}.
  \label{eq:excess-loss}
\end{equation}
$\Delta L$ is the loss the model avoids by doing something that the mean field does
not.

Both losses are next-token cross-entropies averaged over the $100{,}000$
held-out tokens above, so the single $\Delta L$ we report per checkpoint is a
mean over that evaluation sample; $\Dmf$ likewise averages
\eqref{eq:dmf} over the sampled positions of the same text.
The \emph{mean deviation} at a sampled position $(s,X)$:
\begin{equation}
D_{sX} = \operatorname{mean}_{\ell,h}\,\bigl(1-\cos(\v z^{\ell h}_{sX},
\hat{\v z}^{\ell h}_{sX})\bigr),
\label{eq:dmf}
\end{equation}
measures how far the
per-head contributions themselves move from their corpus-averaged
counterparts. The per-checkpoint diagnostic we report is its
average over sampled positions,
$\Dmf = \operatorname{mean}_{s,X}\, D_{sX}$
\footnote{$\Dmf$ is an un-centered cosine
distance between individual head contributions, distinct from the centered
cosine over layers that Section~\ref{sec:validation} uses to score centroid
prediction.}

\subsection{Models Reduce to Their Mean Field, Then Depart from It}
\label{sec:phases}

\begin{figure}[t]
\centering
\includegraphics[width=\textwidth]{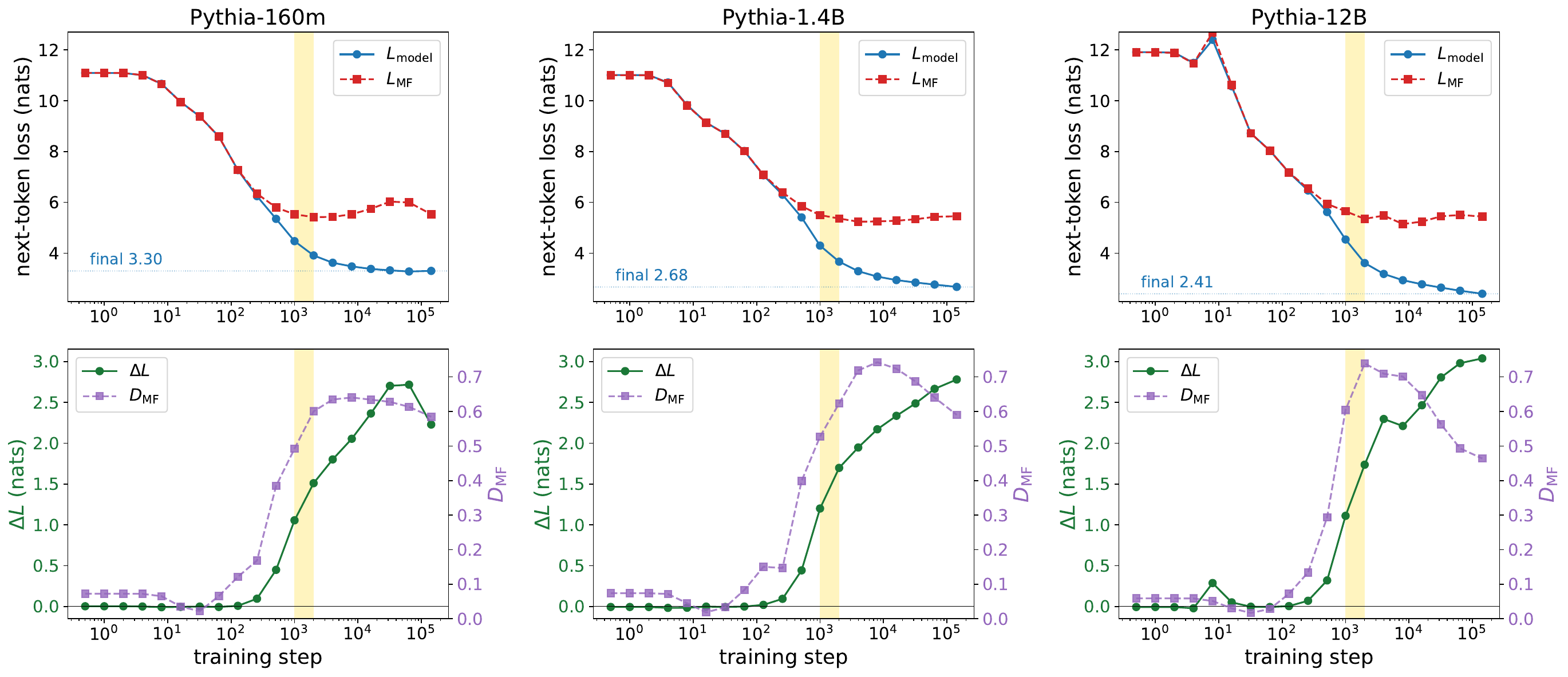}
\caption{Pythia models side by side (160m, 1.4B, 12B, left to right).
\textbf{Top:}
the model's loss (blue) and the loss under the iterated mean field (red)
across training steps (log scale); the
yellow band marks the induction-formation window.
The curves are visually indistinguishable through step
$10^2$ and separate permanently shortly
thereafter.
\textbf{Bottom:} the gap $\Delta
L$ (green, left axis) and the deviation $\Dmf$ (purple, right axis).}
\label{fig:mfgap}
\end{figure}

Figure~\ref{fig:mfgap} plots the experimental results for Pythia-160m (left), Pythia-1.4B (middle), and Pythia-12B (right).  The top row of plots depicts the next-token loss actually incurred by the model (blue) and by the mean-field replacement (red), as training elapses from left to right. The bottom two plots depict our corresponding measures over this experiment: the gap $\Delta L$ (green)
between the two upper curves and the deviation $\Dmf$ (purple, right axis). 
The yellow shading shows the induction band (steps 1000-2000), the window in which induction heads are reported to form in Pythia \citep{tigges2024consistent}.
\footnote{According to \citet{tigges2024consistent}, roughly $2$--$5$ billion tokens, or steps $954$--$2384$ at Pythia's $2.1$M tokens per step.}

\paragraph{In early training, neither measure moves.}  Up through training step $10^2$, both bottom panels show $\Delta L$ negligible and $\Dmf$ also near its floor.  
These measures show that departure from the mean field has not yet occurred: replacing the entire attention computation with the iterated mean field costs nothing in terms of loss. The $\Delta L$ gap is within a few millinats of zero, and the attention contributions of the heads sit, on average, close to their corpus means.
Yet the \emph{losses themselves are falling fast over this
stretch,} by more than three nats, so this regime by no means experiences an absence of learning.
However, what the model has learned so far depends only on which tokens are present, not their order, which is exactly what the mean field captures.
\S\ref{sec:kernel-dynamics} gives the mechanistic reason.

\paragraph{Next, across the induction window, both measures rise in parallel.}
Starting around step 128 and continuing through the induction band, $\Delta L$ and $\Dmf$ both rise sharply.
Of note is that the rise occurs at the
same token count in all three models despite a
$75\times$ difference in parameters (160m to 12B), an indication that the transition is set by data seen rather than model size.
Departure from the mean field here reflects context-dependent computation the mean field does not represent, a topic \S\ref{sec:icl} takes up in detail.

\paragraph{After the induction window, the two measures diverge.}
After the induction band, the two measures behave differently: the per-head deviation $\Dmf$ plateaus in the 160m case
and eases down from a mid-training peak in the 1.4B and 12B cases (most sharply at 12B), while the loss cost of imposing the mean field $\Delta L$ continues to rise.
The top panels show why: the mean-field loss stops tracking
the model loss.  Imposing
the mean field now actively harms the model, not because the mean field intrinsically became worse,
but because the model has learned to rely on something the mean field does not supply.  Comparing the final performance across the three models, their true losses are smaller for larger models, while their mean-field losses remain quite similar.  This shows that the performance benefits obtained by larger models are related to context-dependent processing.

\paragraph{Mean-field loss augments the weight-space account.}
\citet{im2026associate} show early attention weights to be closed-form compositions of corpus statistics (verified on Pythia-1.4B at the earliest checkpoints). 
The mean-field results here are the computational counterpart, but they extend further because similar weights need not mean similar behavior.
And the correspondence of the model with its mean field does not fade gradually: functionally it holds almost exactly, then fails over a narrow interval, so the drift seen in weight space corresponds to a specific event, the arrival of in-context learning.

\paragraph{The effect replicates in OLMo-2.}  The only other model family in which training checkpoints exist in early (pre-1B-token) stages is OLMo \citep{olmo2024olmo2}.  Even in that model, there are limited data points available in the key region.  However, there is enough data to confirm that the same effect appears to take place in OLMo-2, at roughly the same point in the training process.  Details are in Appendix~\ref{app:waypoint-olmo} (Figure~\ref{fig:crossmodel}).

\subsection{Kernel Dynamics Across Training}
\label{sec:kernel-dynamics}

\begin{figure}[t]
\centering
\includegraphics[width=\textwidth]{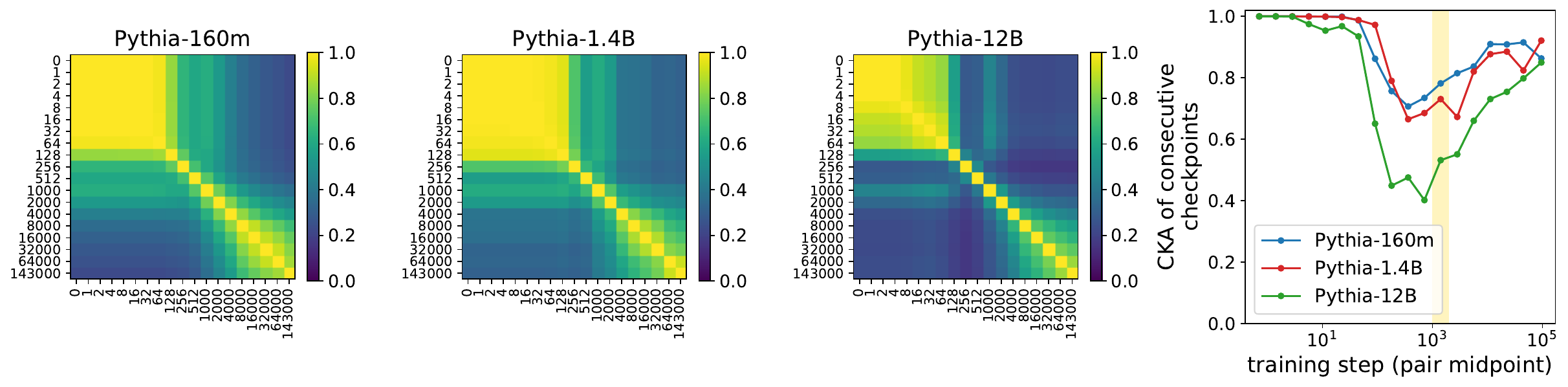}
\caption{
How much the kernel changes across training, all three models,
measured as similarity between checkpoints: a value of 1.0 is no change.
\textbf{Left three panels:} kernel similarity between every pair of checkpoints
(160m, 1.4B, 12B), averaged over heads.
\textbf{Right:}
adjacent-checkpoint kernel similarity, with the
induction window shaded for reference.
}
\label{fig:cka}
\end{figure}

For a deeper understanding of why the model reduces to its mean field and then departs, we study the evolution of the mean field kernel $W$ (\ref{eq:W}).
Figure~\ref{fig:cka} shows how much $W$ changes across training, measured using linear CKA, averaged over heads. (Estimation details in Appendix~\ref{app:waypoint-kernel}).
The left three panels give the kernel similarity between every pair of checkpoints; the right panel gives the similarity between adjacent checkpoints, depicting the rate at which $W$ changes.

Three regimes appear in all three models, aligned with those in \S\ref{sec:phases}.
Up through step 128 the kernel barely moves: adjacent-checkpoint similarity sits at 1.0, even as the model's loss falls steeply over the same checkpoints (Figure~\ref{fig:mfgap}).
\emph{Learning in this regime is in the values, MLPs, and embeddings, not the attention pattern.}
After step 128 the kernel reorganizes, at the same point where the two departure measures begin to rise: adjacent-checkpoint similarity falls to its minimum leading up to %
the induction band, then recovers and stays high afterward.
So mean-field attention $W$ is not arranged ahead of in-context learning: it holds flat early, reorganizes rapidly leading up to and across the induction band, and stabilizes afterward.
The reorganization is much more pronounced for larger models.

The fact that mean-field attention is not changing much in the early stage of learning suggests that the attention heads are still behaving as if randomly initialized.  To confirm this, we measure how evenly each head divides its attention among the types present.
Early in training a type draws attention in proportion to how often it appears, and not otherwise: at initialization a head's attention varies by only about a factor of $1.5$ across types, which is what random projections give.
By the end of training the same heads vary by more than a factor of $20$.
Heads begin to attend selectively when the kernel reorganizes, and per-head spread increases through the rest of training.
Heads also differ sharply: some develop large spread while others barely change from their near-uniform initialization.
Appendix~\ref{app:waypoint-kernel} defines the row-spread measure, tabulates the raw values, and plots its trajectory across training for all three models (Figure~\ref{fig:spread}).

\paragraph{When the mean field holds in practice.}
The measurements in this section support the following conclusion: 
early in training a head's corpus-mean attention to a type is fixed by how often the type occurs, independent of the values those occurrences carry.
Attention and value are therefore uncorrelated across types, which is precisely Assumption~\ref{ass:value-independence}, attention-value independence for the corpus mean field.
The assumption thus holds closely early in training as an empirical fact about Pythia (not a simplification) and stops holding once the kernel reorganizes and heads acquire contextual sensitivity.

\section{In-Context Learning and Contextual Computation} \label{sec:icl} So far, \S\ref{sec:waypoint} has shown that deviation from the mean field identifies \emph{when}, during training, the model begins doing context-specific work that the mean field cannot represent.  We now apply this principle more broadly, to locate \emph{where} that extra computation lives: in which heads, on which prompts, and at which tokens of a prompt.
We take the context-conditional mean field as a \emph{null model for contextual processing}. Because a head's mean-field prediction 
models
what its output would be in the absence of contextual computation, the deviation from it 
estimates
the contextual computation present. %
This makes it a \emph{principled} null: the baseline it sets is the model's own average behavior, not an external reference.

Thus, even though to this point in the paper, the mean-field residual was considered a source of error, in this section we treat the residual as a useful signal.
The residual has two sources: the correlation between where the head attends and the values it moves, and the contextualization of those values (the attended tokens carrying content specific to this context rather than their corpus average).  We make these two concepts more mathematically precise in \eqref{eq:dev-decomp} below.  Each makes a distinct contribution to the residual.

Results in this section are based on
the deviation of a head's actual output $\v z^{\ell h}_{sX}$ from the
context-conditional prediction $\hat{\v z}^{\ell h}_{sX}$ of
\eqref{eq:ccmf}. 
We compute two measures:  the \emph{context-dependent deviation} of head
$(\ell,h)$ on $X$, and its per-context average (over all heads):
\begin{equation}
  D^{\ell h}_{sX} = 1-\cos\!\big(\v z^{\ell h}_{sX},\,\hat{\v z}^{\ell h}_{sX}\big),
  \qquad
  D_{\mathrm{MF}}(X) = \operatorname{mean}_{\ell,h}\, D^{\ell h}_{sX}.
  \label{eq:deviation}
\end{equation}

Note that $s$ does not appear in the average to produce \Dmf~in \eqref{eq:deviation}, as $X$ is associated with a unique query type $s$.

In this section, we will show the following:
\begin{description}

\item[Mean field deviation rises with in-context learning.]
We show three experiments supporting the strong association of mean-field
deviation with ICL
(\S\ref{sec:icl-correlates}).

\item[Deviation decomposes additively into two contributors.] For each context, the
deviation splits exactly into an attention-pattern covariance and a
contextualization of the attended values, each of which carries information about
ICL in both GPT-2 and Gemma-2-2B
(\S\ref{sec:mf-deviation-decomposed}).
Averaged over contexts per-head, the split characterizes every
head's deviation across those two axes,
and we find that deviation in general involves both axes.

\item[Mean field deviation organizes an unsupervised survey of head activity.
] 
Ranking GPT-2 heads by mean field deviation on generic text %
surfaces the known
in-context heads;
further analyses confirm the documented heads' roles and characterize previously
undocumented GPT-2 heads.
(Method in Appendix~\ref{sec:instrument}, findings in
\S\ref{sec:catalog-findings}.)

\item[Mean field deviation isolates function vectors.] Aggregating the mean field residual on a task set, a head's deviation is causally a function vector, and the corpus null lets it be extracted from as little as one behavior-exhibiting prompt (\S\ref{sec:fv}).

\end{description}

\subsection{Mean-Field Deviation Rises with In-Context Learning}
\label{sec:icl-correlates}

Our first set of experiments shows that in-context learning has a strong effect on mean-field deviation.  Note that we do not claim that this goes both ways;  mean-field deviation can be large for other well-understood reasons; we study examples in \S\ref{sec:catalog-findings}.

We present three experiments intended to illustrate the relationship between mean-field deviation and in-context learning at different scales.
Experiment 1 asks, within a single prompt where ICL is present, whether deviation localizes ICL to the token(s) where a known induction head fires.
Experiment 2 asks, across whole prompts, whether deviation succeeds at separating prompts by the in-context learning they do, from induction down to random text.
And Experiment 3 asks, across training at the corpus level, whether deviation emerges during training at the point where induction is known to arise.

\paragraph{Experiment 1: Mean-field deviation rises with induction as defined by \cite{olsson2022induction}.}
\label{sec:icl-meter}

In order to study a controlled in-context signal, we adapt the induction setup of \cite{olsson2022induction}: a real OpenWebText chunk $c$ of $128$ tokens is presented as $[c;c]$ (BOS-prepended), so that in the second copy every token has a distant earlier occurrence and in-context (induction) prediction becomes available.  While \citeauthor{olsson2022induction}\ score induction heads on repeated \emph{random} tokens, to isolate the mechanism from content, we use real text so that the loss drop at the copy boundary is a quantity the model would actually realize on corpus data.  (Our Experiment~3 will use their random-token form.)  We treat each sequence as 128 contexts, by considering each token as a query $s$, and we measure the context-conditional mean-field deviation $\Dmf$ \eqref{eq:deviation} for that position.  Over $500$ such sequences we record, per query position, the mean next-token loss and the mean $\Dmf$; and, for each pair of matched copy-1 / copy-2 positions of the same token, an \emph{in-context loss reduction} ($\mathrm{loss}({\text{copy }1})-\mathrm{loss}({\text{copy }2)}$) against the mean deviation ($\Dmf({\text{copy }2})-\Dmf({\text{copy }1})$).

\begin{figure}[t]
\centering
\includegraphics[width=\linewidth]{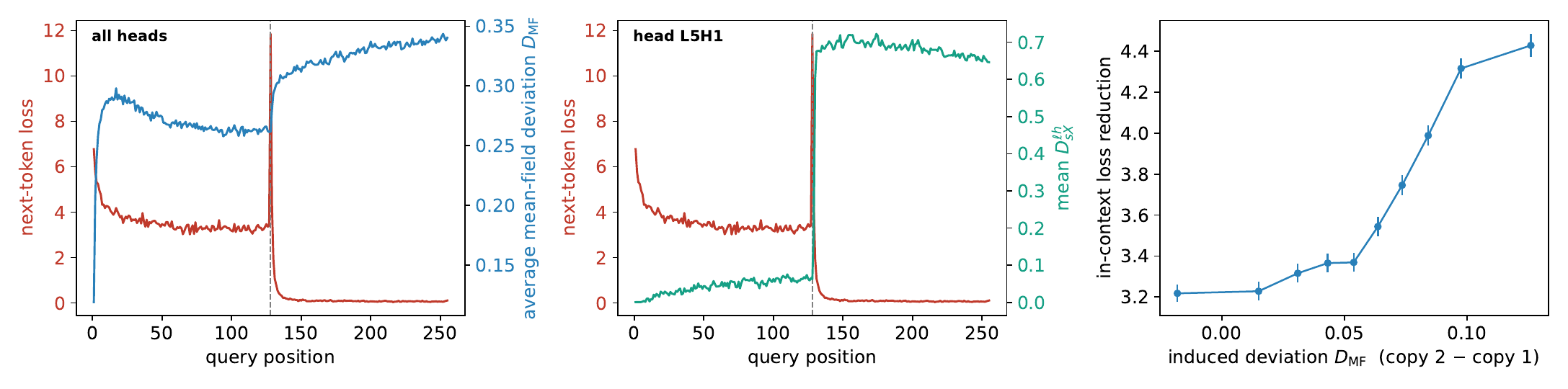}
\caption{Instance-level context-conditional mean-field deviation on GPT-2 small. \textbf{Left:} per-query-position next-token loss (red) and all-head mean-field deviation $\Dmf$ (blue) on repeated text $[c;c]$; the dashed line marks the copy boundary at position $128$, separating the two parts of the prompt. \textbf{Middle:} $D^{\ell h}_{sX}$ averaged over prompts $X$ for the single head that rises most across the boundary, GPT-2 L5H1, a known induction head. \textbf{Right:} per-token in-context loss reduction ($\mathrm{loss}_{\text{copy }1}-\mathrm{loss}_{\text{copy }2}$) binned by the induced deviation ($\Dmf({\text{copy }2})-\Dmf({\text{copy }1})$); error bars are standard errors of the mean.}
\label{fig:icl-meter}
\end{figure}

Figure~\ref{fig:icl-meter} (left) shows that at the copy boundary the next-token loss drops sharply while the mean-field deviation sharply rises. Among GPT-2 heads, deviation rise is strongest in L5H1, a known induction head (Figure~\ref{fig:icl-meter}, middle).  The right side of the figure shows that \emph{mean-field deviation has a sensitive response to induction} -- the mean-field deviation increases monotonically with induction as measured by in-context loss reduction.

\paragraph{Experiment 2: Mean-field deviation tracks the amount of ICL across prompt classes.}
Next, we ask whether mean-field deviation distinguishes different types of ICL.  We study four 
prompt classes to vary the kind of ICL expected to take place. 
\begin{itemize}
    \item \textbf{induction} (planted ICL) Each prompt consists of 12 random tokens, followed by a (random-length) repeated subsequence of those tokens, and the query is the last token in the repeated subsequence.
    \item \textbf{few-shot} (planted ICL)  Each prompt gives
five \texttt{input:output} exemplars of one task (e.g.\ \texttt{hot:cold}),
followed by a held-out query input and a colon.  The correct completion is
determined only by the rule the exemplars establish, so predicting it
requires inferring the task from the context. We use five tasks: antonym,
capital, past tense, plural, and capitalization. 
    \item \textbf{natural} (background ICL) Prompts are sequences from OpenWebText, whose last token is one of the tracked types (so that a type specific mean-field prediction can be made).  Each prompt is taken from a different OWT document.  This represents the ICL occurring naturally in typical text.
    \item \textbf{random} (matched non-ICL control) Prompts are the same random sequences and query tokens as used in `induction' but the query is restricted to only be in the first copy, so the next token is unpredictable.  This makes it a matched-control for the `induction' case.
\end{itemize}

For each prompt we compute $\Delta L$ \eqref{eq:excess-loss} and the mean-field deviation $\Dmf$.
Additionally, we shuffle the context positions using a random permutation $\pi$, and measure the resulting change in model loss ($\Delta L_{\text{perm}}=L_{\text{model}}(\pi X)-L_{\text{model}}(X)$).
To verify the mean-field's order-invariance we measure relative change in the mean-field
contribution ($\Delta\hat{\v{z}}_\text{perm} = \|\hat{\v{z}}(\pi X)-\hat{\v z}(X)\|/\|\hat{\v z}(X)\|$).

\begin{figure}[th]
\centering
\includegraphics[width=\textwidth]{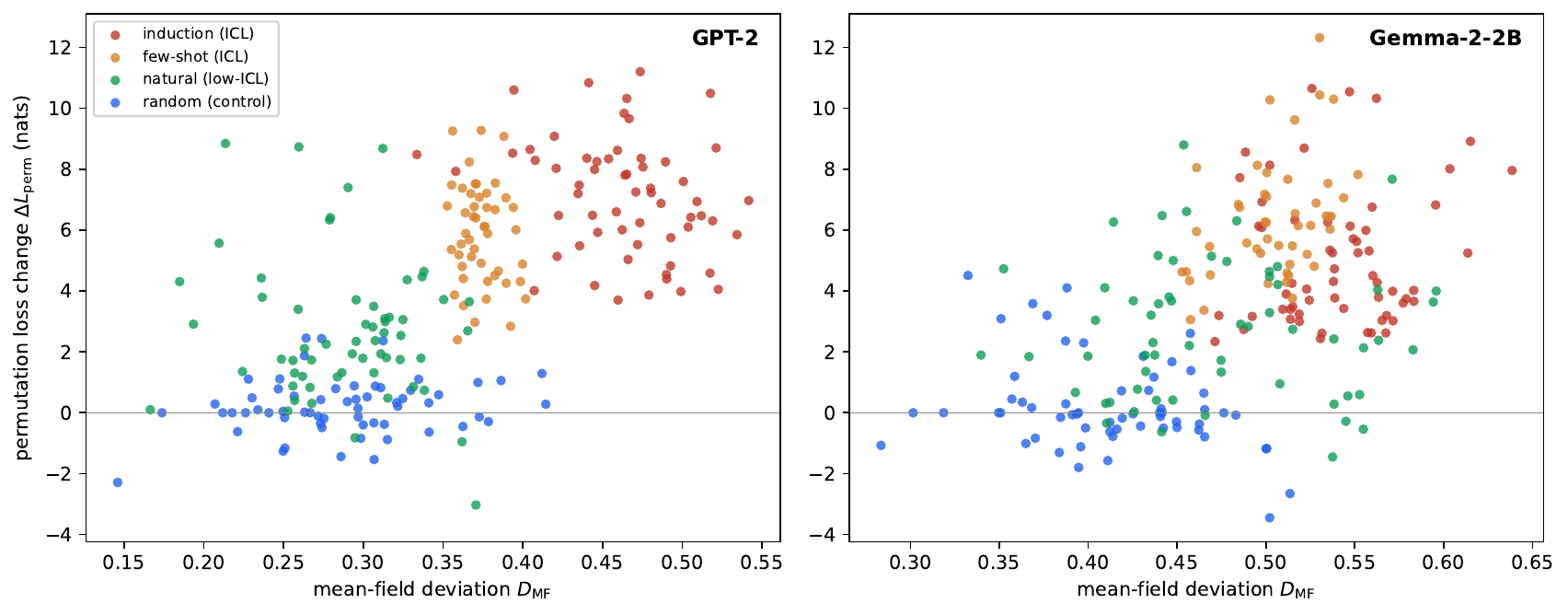}
\caption{Mean-field deviation and in-context learning: per-prompt
permutation loss change $\Delta L_{\mathrm{perm}}$ against mean-field deviation,
all $228$ prompts.  The prompt classes separate along both axes (shown in \S\ref{app:dev-corr}).}

\label{fig:icl}
\end{figure}

Table~\ref{tab:icl} reports statistics when attention contribution at every head and layer of the model is replaced with the context-conditional mean field \eqref{eq:ccmf} for each context. 
In the table, $\Delta L_\mathrm{MF}$ is the loss difference when attention is
replaced by the mean field, and $\Delta L_\mathrm{perm}$ is the loss difference when the context is randomized.
Both are computed per prompt from \eqref{eq:excess-loss} and reported as
the mean over the $n$ prompts of the class, $n$ as given in the table.%
\footnote{
Loss increases are meaningful, because the models are able to perform the tasks.  Gemma-2's top-1 accuracy is $\ge 98\%$ on all induction and few-shot prompts; GPT-2's is $95\%$ on induction and $73\%$ on few-shot, the lower figure consistent with its weaker few-shot behavior in Figure~\ref{fig:icl}.}
Figure~\ref{fig:icl} shows detailed results for every prompt in the dataset, plotting $\Delta L_\text{perm}$ against mean-field deviation.

The table and figure show that larger mean-field deviations correspond to the context categories expected to have larger ICL components, and to larger changes in loss when the context is permuted.  Both substituting the mean field and permuting the context are catastrophic when the next token is in-context-determined (induction, few-shot) and harmless when it is not (random).  These facts all suggest that mean-field deviation is responsive to the presence of ICL.  We note that this result is about class differences.  $\Dmf$ orders the classes as their permutation loss $\Delta L_{\text{perm}}$ does, a separation driven by the designed differences between classes and not by any within-class relationship.  We therefore report it as a standardized class separation (Appendix~\ref{app:dev-corr}), not as a correlation over the pooled prompts.
The random control's $\Delta L_{\text{perm}}$ is zero in expectation, its
context being exchangeable by construction, and the tabled values are consistent
with zero ($p=0.09$ / $0.29$).
The mean-field read is essentially unchanged under permutation
($\Delta\hat{\v z}_{\text{perm}}\sim10^{-7}$), confirming it supplies an order-free null.

\begin{table}[th]
\centering\small
\begin{tabular}{llrcccc}
\toprule
model & class & $n$ & $\Dmf$ & $\Delta L$ & $\Delta L_{\text{perm}}$ & $\Delta\hat\z_\text{perm}$ \\
\midrule
\multirow{4}{*}{GPT-2}
 & induction \textit{(planted ICL)}     & 60 & 0.46 & $+9.19$ & $+7.09$ & \multirow{4}{*}{$1.4\!\times\!10^{-7}$} \\
 & few-shot \textit{(planted ICL)}          & 48 & 0.37 & $+9.90$  & $+5.86$ & \\
 & natural \textit{(background ICL)}       & 60 & 0.29 & $+2.36$  & $+2.53$ & \\
 & random \textit{(control)} & 60 & 0.29 & $-0.82$  & $+0.20$ & \\
\midrule
\multirow{4}{*}{Gemma-2-2B}
 & induction \textit{(planted ICL)}     & 60 & 0.54 & $+14.01$ & $+5.03$ & \multirow{4}{*}{$2.1\!\times\!10^{-7}$} \\
 & few-shot \textit{(planted ICL)}          & 48 & 0.50 & $+18.02$ & $+6.33$ & \\
 & natural \textit{(background ICL)}       & 60 & 0.47 & $+2.36$  & $+2.72$ & \\
  & random \textit{(control)} & 60 & 0.41 & $+2.00$  & $+0.21$ & \\
\bottomrule
\end{tabular}
\caption{Per prompt class: mean-field deviation $\Dmf$; loss increase
$\Delta L$ from replacing attention by the mean field
(all layers); model loss change under context permutation
$\Delta L_{\text{perm}}$; and permutation sensitivity
$\Delta\hat{\v z}_{\text{perm}}$ (maximum over all $228$ prompts).  All columns
are averaged over heads and over the prompts of the class, $n$ per class as
shown.}
\label{tab:icl}
\end{table}

\paragraph{Experiment 3: Mean-field deviation emerges during training simultaneously with ICL.}
\S\ref{sec:waypoint} showed that the deviation, summed over heads, rises at the induction transition.
Here we look head by head: does the deviation emerge in \emph{precisely the heads} that become induction heads, and at the same checkpoint?
We test this on nineteen Pythia-160m training checkpoints (steps $0$ to $143{,}000$),
spanning the induction-formation transition that \cite{olsson2022induction} and
\cite{tigges2024consistent} place shortly after $2\times10^{9}$ tokens. At each checkpoint we
measure, for all $144$ heads, the per-head mean-field deviation 
$\bar S^{\ell h}$,
the average over query types of the per-type deviation (\S\ref{sec:ranking},
\eqref{eq:ddec}), from that checkpoint's own
corpus pass, and a standard behavioral induction score.

\begin{figure}[t]
\centering
\includegraphics[width=0.52\linewidth]{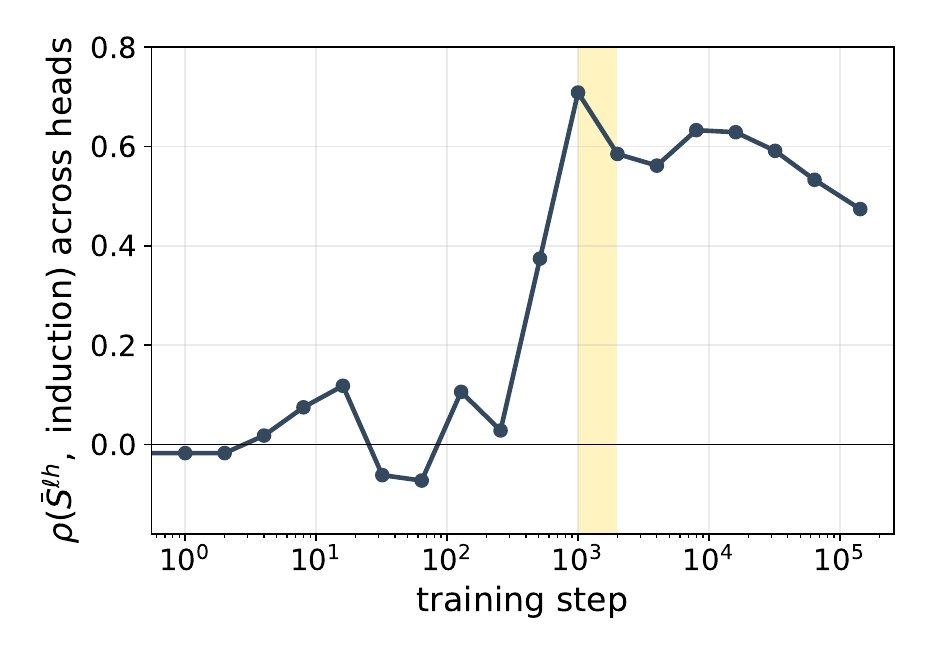}
\caption{Correlation between mean-field deviation 
$\bar S^{\ell h}$ (see Eq. \ref{eq:ddec})
and behavioral
induction score across training, Pythia-160m, over the nineteen
checkpoints of Figure~\ref{fig:mfgap}.
The yellow band marks the
induction-formation window (steps $1000$--$2000$, ${\approx}2\times10^{9}$
tokens), as in that figure.}
\label{fig:e15-correlation}
\end{figure}

The induction score is measured
independently of the mean field, from the head's attention pattern alone: we feed batches of
\emph{randomly} generated token sequences repeated back-to-back, $[\mathrm{BOS};\,c;\,c]$, and score
each head by the average attention a token in the second copy places on the position immediately
after that token's first occurrence --- the induction target~\citep{olsson2022induction}. 
In contrast to Experiment~1, this probe uses \emph{random} repeated tokens (rather than repeated text from OpenWebText). The reason for this difference is to isolate the induction circuit itself and avoid any content-based regularities. 
Figure~\ref{fig:e15-correlation} plots the cross-head Spearman correlation between $\bar S^{\ell h}$
and the induction score across the $144$ heads over training.
$\bar S^{\ell h}$ is defined in Equation \ref{eq:ddec} below, and represents the average over all types in the corpus of a head's deviation.
Among the ten checkpoints before the transition (through step 256), no correlations are significantly different from zero (all $p\ge0.16$).  It rises
at step $512$ ($0.37$), 
peaks at the documented transition ($0.71$), and then declines somewhat as later, non-induction contextual heads accumulate and dilute the pure induction signal.  Mean-field deviation
thus appears in the heads that become induction heads, at the checkpoint they do --- further evidence
that the deviation rises with the onset of in-context computation.

\subsection{Mean Field Deviation Combines Two Contributors}
\label{sec:mf-deviation-decomposed}

In order to probe the relationship between mean-field deviation and in-context learning, we decompose the deviation into its two context-dependent components, and consider how each relates to contextual processing.

\paragraph{Decomposing the deviation: covariance and contextualization terms.}
The deviation \eqref{eq:deviation} compares the actual contribution in a given context $X$ to the mean field approximation
$\hat{\v z}_X=\sum_t\hat A_{stX}\boldsymbol\mu_t$,
with $\hat{A}_{stX}$ the context-conditional kernel \eqref{eq:ctx-kernel}.

The difference splits 
into two additive terms:
\begin{equation}
  \v{z}_X-\hat{\v{z}}_X
  =\underbrace{\textstyle\sum_t (A_{stX}-\hat{A}_{stX})\,\overline{\v v}_{tX}}_{\textbf{(a) covariance term}}
  +\underbrace{\textstyle\sum_t \hat{A}_{stX}\,\big(\overline{\v v}_{tX}-\boldsymbol\mu_t\big).}_{\textbf{(b) contextualization term}}
  \label{eq:dev-decomp}
\end{equation}

Term (a) is the part of the deviation due to variation in attention: each summand is zero if $A_{stX}$, the attention the head actually places on type $t$, matches the mean-field kernel $\hat A_{stX}$.  
Because the kernel is approximately unbiased for the attention over the
ensemble of contexts sharing the query type
($\mathbb E_{st}[\hat A_{st}]\approx\mathbb E_{st}[A_{st}]$),
averaging term (a) over that ensemble gives the residual
attention--value \emph{covariance}
$\sum_t\operatorname{Cov}_{st}\!\big(A_{st}-\hat
A_{st},\,\overline{\v v}_t\big)$
(full derivation is in Appendix~\ref{app:dev-split}.)
Term (b) is the part of the deviation due to variation in the values, the
\emph{contextualization} of these values: %
each summand is zero exactly when $\overline{\v v}_{tX}$, the value type $t$ carries in this
context, matches its corpus type-mean $\boldsymbol\mu_t$.\footnote{Note that because $\hat{\v z}$ uses the measured type-mean $\boldsymbol\mu_t$ rather than the centroid value $v(\v{c}^{\ell}_t)$, the value-map Jensen gap
does not appear in the deviation at all. This is deliberate: that gap is a static property of the value map's curvature at the centroid, independent of $X$, so it cannot carry in-context information --- and in any case it is small in all of our experiments (Appendix~\ref{app:clustering}).}

These two terms describe different \emph{mechanisms} behind contextual processing.  In short, contextual processing can arise either because (a) the attention provided to certain tokens is driven to move specific values into the query, or (b) because the tokens attended to in the context are highly contextualized.  What \eqref{eq:dev-decomp} allows us to do is to separate these two sources into \emph{additive} terms that can be measured separately.

\paragraph{How do the deviation components vary across tasks?}
Since the two terms describe different mechanisms, we measure each
separately on the prompts of Table~\ref{tab:icl}, and ask how they vary across tasks.
Each contribution is formed teacher-forced, from the model's true
residuals at layer $\ell$ and the measured corpus value means
$\boldsymbol \mu^\ell_t$.
We summarize each per context as a scale-free \emph{metric}.  
Writing $\mathbf{a}$ and $\mathbf{b}$ for the two terms of
(\ref{eq:dev-decomp}), so that $\mathbf{a}+\mathbf{b}=\mathbf{z}-\hat{\mathbf{z}}$
exactly, let $\mathbf{e}=\mathbf{z}-\hat{\mathbf{z}}$ denote the total error and $\mathbf{e}_{\perp}$ be its 
component that is orthogonal to $\hat{\mathbf{z}}$, ie, $\mathbf{e}_{\perp} = \mathbf{e}-
  \frac{\mathbf{e}\cdot\hat{\mathbf{z}}}{\Vert\hat{\mathbf{z}}\Vert^2}
  \hat{\mathbf{z}}$.
We measure each term's contribution to $\mathbf{e}_{\perp}$ and normalize:
\[
\sigma_{\mathrm{cov}}
   =\frac{\mathbf{a}\cdot\mathbf{e}_{\perp}}{\lVert\mathbf{e}_{\perp}\rVert^2},
\qquad
\sigma_{\mathrm{ctx}}
   =\frac{\mathbf{b}\cdot\mathbf{e}_{\perp}}{\lVert\mathbf{e}_{\perp}\rVert^2},
\qquad
\sigma_{\mathrm{cov}}+\sigma_{\mathrm{ctx}}
   =\frac{\mathbf{e}\cdot\mathbf{e}_{\perp}}{\lVert\mathbf{e}_{\perp}\rVert^{2}}
   =1,
\]
The metrics%
\footnote{When $\mathbf{e}_{\perp}=\mathbf{0}$, $\mathbf{z}$ is collinear with $\hat{\mathbf{z}}$, so the deviation is either exactly 0 (aligned) or 2 (anti-aligned), and the perpendicular shares are undefined.  This case does not occur in our measurements.}
are
\begin{equation}
M_{\mathrm{cov}}(X)=\operatorname{mean}_{\ell,h}\,
   \sigma_{\mathrm{cov}}^{\ell h}\,D^{\ell h}_{sX},
\qquad
M_{\mathrm{ctx}}(X)=\operatorname{mean}_{\ell,h}\,
   \sigma_{\mathrm{ctx}}^{\ell h}\,D^{\ell h}_{sX}.
\label{eq:mcov-mctx}
\end{equation}
Because $\sigma_{\mathrm{cov}}+\sigma_{\mathrm{ctx}}=1$ per head, the metrics
sum to the deviation $D^{\ell h}_{sX}$ of (\ref{eq:deviation}) exactly. Note
that the shares are allocated along $\mathbf{e}_{\perp}$ only: the component of
$\mathbf{e}$ parallel to $\hat{\mathbf{z}}$ rescales the contribution without
rotating it, and the cosine-based deviation is insensitive to that rescaling,
so it is deliberately excluded from the split. The shares are signed, so one
mechanism can partially cancel the other.

\begin{table}[th]
\centering\small
\begin{tabular}{llrccccc}
\toprule
model & class & $n$ & $\Dmf$ & $M_{\text{cov}}$ & $M_{\text{ctx}}$ & $\Delta L$ & $\Delta L_{\text{perm}}$ \\
\midrule
\multirow{4}{*}{GPT-2}
 & induction \textit{(planted ICL)}     & 60 & 0.46 & 0.37 & 0.09 & $+9.19$ & $+7.09$ \\
 & few-shot \textit{(planted ICL)}          & 48 & 0.37 & 0.25 & 0.13 & $+9.90$  & $+5.86$ \\
 & natural \textit{(background ICL)}       & 60 & 0.29 & 0.18 & 0.11 & $+2.36$  & $+2.53$ \\
 & random \textit{(control)} & 60 & 0.29 & 0.22 & 0.07 & $-0.82$  & $+0.20$ \\
\midrule
\multirow{4}{*}{Gemma-2-2B}
 & induction \textit{(planted ICL)}     & 60 & 0.54 & 0.39 & 0.15 & $+14.01$ & $+5.03$ \\
 & few-shot \textit{(planted ICL)}          & 48 & 0.50 & 0.29 & 0.21 & $+18.02$ & $+6.33$ \\
 & natural \textit{(background low-ICL)}       & 60 & 0.47 & 0.29 & 0.18 & $+2.36$  & $+2.72$ \\
 & random \textit{(control)} & 60 & 0.41 & 0.33 & 0.08 & $+2.00$  & $+0.21$ \\
\bottomrule
\end{tabular}
\caption{Table~\ref{tab:icl} augmented with the covariance and contextualization metrics
$M_{\mathrm{cov}},M_{\mathrm{ctx}}$ of \eqref{eq:mcov-mctx}, which  sum to $\Dmf$.}
\label{tab:icl-terms}
\end{table}

\paragraph{Results.}

Table~\ref{tab:icl-terms} shows how the two components of mean-field
deviation vary across experiments: the covariance term $M_{\mathrm{cov}}$ (roughly, the `unusualness of attention'), and 
 the contextualization term $M_{\mathrm{ctx}}$  (roughly, the `unusualness of representations'). 
The covariance term is consistently the larger part of mean-field deviation.  But a large covariance term is not by itself an ICL signal: the random control shows high 
$M_{\mathrm{cov}}$ yet
the lowest
$M_{\mathrm{ctx}}$
of any class and near-zero $\Delta L_{\text{perm}}$.  This is because its random tokens are out-of-distribution for the corpus statistics $W_{st}$ that $\hat A$ is built from.
Among the genuine ICL classes, 
the dominant mechanism is model-dependent:
in Gemma-2 the attended tokens are more contextualized than in GPT-2 (comparing metrics to random), while GPT-2's ICL deviation is driven mainly by unusual attention.
Appendix~\ref{app:dev-corr} separates the two contributions more finely, putting them on a common scale (Table~\ref{tab:icl-corr}) as the separation of each class from the random control.  This confirms the model-dependent dominance, and shows the deviation ranks the designed conditions rather than individual prompts within them.

Overall, the picture that emerges is that ICL, as characterized by mean-field deviation, is not purely a function of covariance or contextualization alone, but in general involves both mechanisms.

\subsection{Mean Field Deviation Recovers Contextual Heads and Circuitry}
\label{sec:catalog-findings}

We next show how mean-field deviation can be used to identify heads that carry contextual
computation.  
As a first cut, we rank all heads by their context-conditional deviation, averaged across query types.
We report our findings on GPT-2 throughout, because the circuit literature has cataloged it in enough detail to inform our readings.%
\footnote{The head survey %
is repeated on Gemma-2-2B in Appendix~\ref{app:gemma}.
Some of the same machinery appears, but with no comparable catalog to check it against, our survey can only test the reproducibility of our methods, not whether its readings agree with an independent account.}
A number of heads of GPT-2 that the circuit literature has documented appear near the top of the rankings, and so do other contextual heads those methods miss, including positional and content-moving heads which we newly identify.
We additionally place heads onto the 
$M_{\mathrm{cov}}$/$M_{\mathrm{ctx}}$ plane of \S\ref{sec:mf-deviation-decomposed}, which visually 
localizes both high and low-deviation groups of heads that share a common documented role.
Finally, we investigate how known circuit pairs (e.g., a previous-token head with an induction head) align with
respect to these measures.

\paragraph{Ranking heads by corpus deviation.}
\label{sec:ranking}%
We rank all 144
GPT-2 heads by their \emph{corpus deviation}: how far the head's
corpus-average behavior departs from the mean-field account, at the level
of query types. 
For query type $s$, the head's measured contribution is
$\overline{\v z}^{\ell h}_s = \E_s\big[\sum_t A_{st}\,\overline{\v v}_t\big]$
(Eq.~\ref{eq:exact-contribution-by-type}), and its context-averaged mean-field prediction
is
$
  \overline{\hat{\v z}}^{\ell h}_s \;=\; \E_s[\hat{\v{z}}^{\ell h}_{sX}] = \textstyle\sum_t
     \E_s\big[\hat A^{\ell h}_{st}\big]\,\boldsymbol\mu_t,
$
which uses the context-averaged context-conditional kernel of
\eqref{eq:ctx-kernel}.\footnote{Both quantities are
accumulated in one corpus pass over 2M OpenWebText tokens, the same pass
that measures the kernels of \S\ref{sec:validation}. Tokens outside
the top-1000 contribute through BOS and ``other'' buckets in both terms.} 

The per-type deviation and the head's corpus score are
\begin{equation}
  S^{\ell h}_s = 1-\cos_c\!\big(\overline{\v z}^{\ell h}_s,\,\overline{\hat{\v z}}^{\ell h}_s\big),
  \qquad
  \bar S^{\ell h} = \operatorname{mean}_s\, S^{\ell h}_s,
  \label{eq:ddec}
\end{equation}
where $\cos_c$ is the centered cosine (the mean over query types is
removed from both arguments, so the score measures how the head's output
\emph{varies} with the query type, not a shared offset) and the mean is
over the types with at least 50 corpus occurrences. It is centered because across query types, the contributions share a large common offset that would dilute the deviation.  Note that it is not possible to do such a centering on a single context.

Note that $\bar S^{\ell h}$ is based on mean field deviation like $D^{\ell h},$ but unlike $D^{\ell h}$, the head's contribution is averaged over contexts
\emph{before} the deviation is taken, signified by the bar on $\overline{\hat{\v z}}^{\ell h}_s$.  Per-context
errors that are unbiased cancel in $\overline{\hat{\v z}}^{\ell h}_s$. For example, the mean field badly predicts the attention of a
previous-token head in \emph{any single context} (that attention is one-hot whereas the kernel spreads over the composition), but its errors average out against its own
bigram statistics, so by \eqref{eq:ddec} it scores low.  The score isolates the \emph{systematic},
type-conditioned part of a head's context sensitivity, which is what makes it useful
for detecting in-context learning.

\begin{figure}[t]
\centering
\includegraphics[width=0.72\linewidth]{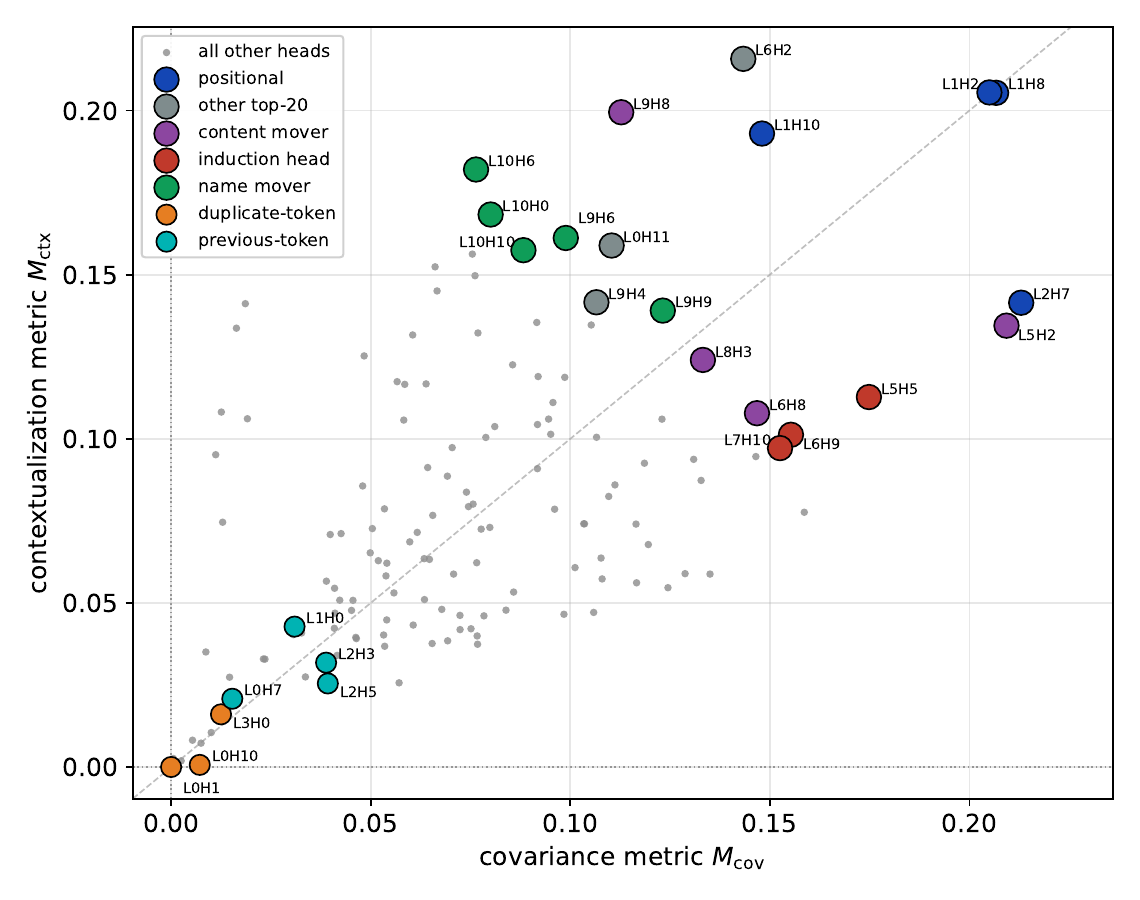}
\caption{All $144$ GPT-2 heads on the
$M_{\mathrm{cov}}$/$M_{\mathrm{ctx}}$ plane \eqref{eq:head-meters}, natural
OpenWebText. Large
markers: the top twenty heads of Table~\ref{tab:ranking-cc}, colored by functional
group, and the named heads from the \emph{bottom} of the ranking
(Table~\ref{tab:bottom}).
\hPos{L1H3} (rank 3) is omitted: at $(+0.34,+0.05)$ it is a far outlier.
The faint dashed diagonal is $y=x$: heads below it are
covariance-dominated, heads above it contextualization-dominated.}
\label{fig:ccmf-scatter}
\end{figure}

\paragraph{What the ranking recovers.}
We compute the corpus deviation
$\bar S^{\ell h}$ for all 144 GPT-2 heads and rank them.
Figure~\ref{fig:ccmf-scatter} shows all heads on the $M_{\mathrm{cov}}$ vs. $M_{\mathrm{ctx}}$ axes, and
Table~\ref{tab:ranking-cc} shows the top twenty, with deviation also broken down by 
$M_{\mathrm{cov}}$ vs. $M_{\mathrm{ctx}}$ \eqref{eq:head-meters}, along with the behavioral
induction score of \citet{olsson2022induction}.
In the final two columns, we document the role of each head from the literature and from our own studies.
Roles are attributed to the
most authoritative source that names the head: the IOI
circuit~\citep{wang2023ioi}, 
the induction-head behavioral class
defined by the test in \citet{olsson2022induction},
and this paper.
Failing all of those, we use the SAE-based head
survey~\citep{kissane2024everyhead}, whose annotations are unrefereed and
should be read as indicative; those entries are given in quotation marks
to mark them as that survey's characterizations rather than established
roles.  

We find that the top heads include many heads with a known in-context learning role. For example, \citeauthor{wang2023ioi} identify four induction heads, three of which are in the top-20 here; and they identify 11 name movers, 5 of which are in the top-20 here. 
However,
the top of the list
is not limited to ICL heads. The top twenty also include
heads that move absolute position, and heads whose role we cannot document.
We walk
through the full top-20 list of heads below, with additional per-head detail in Appendix~\ref{app:heads}.

\paragraph{Position movers (\hPos{L1H8}, \hPos{L2H7}, \hPos{L1H2}, \hPos{L1H3}, \hPos{L1H10}).}
    The first example of contextual processing we illustrate is \emph{positional} transport.  For these heads,  $90$--$99\%$ of their within-type output variance is contained in GPT-2's positional subspace. 
    We find that for each head, what is moved is the \emph{absolute position} of the attended-to key token.
    Of these heads, 
    \hPos{L1H2} attends locally (usually to the token immediately before the query), \hPos{L1H10} attends widely across the context, and the remaining three heads attend to BOS.  These latter three therefore move the position of the BOS into the query token.  We perform a detailed analysis of the mechanism, explore why this is useful and what it can accomplish in Appendix~\ref{app:positional}.

\paragraph{Induction heads (\hInd{L5H5}, \hInd{L6H9}, \hInd{L7H10}).} 
These heads show high induction scores on the metric of \citep{olsson2022induction}.  They are highly covariance-dominated, reflecting the fact that their contextual processing arises from unusual attention.  This is because an induction head attends to an earlier token that is selected based on a previous pattern in the context, rather than being predictable based on the earlier token's type.

\paragraph{Name movers (\hMover{L9H9}, \hMover{L9H6}, \hMover{L10H6}, \hMover{L10H0}, \hMover{L10H10}).}
Three name movers \hMover{L9H9}, \hMover{L9H6}, \hMover{L10H0} and backup name movers \hMover{L10H6} and \hMover{L10H10} were previously characterized as part of the IOI circuit in \citet{wang2023ioi}. These heads show intermediate induction scores (0.15-0.45), showing that they are sensitive to preceding patterns in the input, but do not perfectly copy them.  These heads are contextualization dominated; they move contextualized representations from tokens representing non-repeated names in IOI prompts, a kind of content-driven data movement.  

\paragraph{Content Movers (\hCont{L9H8}, \hCont{L5H2}, \hCont{L8H3}, \hCont{L6H8)}.} A number of heads at the top of the list are not previously characterized in the literature.  
To gain some insight into their potential role, we attempt to interpret aspects of their function. This succeeds to varying degrees, and we do not claim that these characterizations are complete or highly accurate. Since they show high mean-field deviation, we try to interpret these heads by examining contexts and in some cases decoding the content that they move.  We do this by 
extracting per-context deviation vectors $\v\Delta_i = \v z_i - \hat{\v z}_i$ over a corpus, and selecting contexts containing the highest-norm (largest deviation) instances. For each, we examine the
context around the max-deviation token, and attempt to decode the deviation vector using the Logit Lens \cite{nostalgebraist2020logitlens} and the J-lens \cite{gurnee2026workspace}.   We generically call these heads `content movers.'
\begin{description}
    \item[\hCont{L9H8}.] Decoding its high-scoring deviation vectors, we find that in all 8 out of the top 8 cases, the deviation vector decodes via J-lens to the document's topic (Table~\ref{tab:l9h8}), suggesting that what makes this head's behavior not type-predictable is its movement of topic information into the query.  
    \item[\hCont{L5H2}.] J-lens decoding suggests that this head moves an abstract completion of degree or limitation computed from the local construction (Table~\ref{tab:l5h2-occ}).
    \item[\hCont{L8H3}.] High-deviation contexts show movement of content in parallel and enumerated frames ---
    repeated constructions, itemized lists, paired names --- attending to the
    frame's anchor (Table~\ref{tab:l8h3-occ}).
    \item[\hCont{L6H8}.] High-deviation contexts show movement of content within idiomatic comparative frames (``as \ldots as one can,'' ``to say the least''; Table~\ref{tab:l6h8-occ}).
\end{description}

\paragraph{Uncharacterized heads (\hOther{L9H4}, \hOther{L6H2}, and \hOther{L0H11}).}  Three heads remain uncharacterized among the top 20.  \hOther{L9H4} is characterized in ~\citep{kissane2024everyhead} as `particle/pronoun bigram.'  We were not able to reach a confident characterization of \hOther{L6H2} or \hOther{L0H11}.

\paragraph{Heads at the bottom of the ranking.}  We study the low end of the ranking in 
Figure~\ref{fig:ccmf-scatter} and
Table~\ref{tab:bottom} (Appendix~\ref{app:bottom}).  These are heads whose corpus-average behavior is indistinguishable from a head that only uses type and co-occurrence to determine attention. We note that a number of heads with known functions show up there as well.  Four heads are duplicate-token heads (\hDup{L0H1}, \hDup{L0H10}, and \hDup{L3H0} were identified in \cite{wang2023ioi}; \hDup{L0H2} was detected by our own inspection of attention).  These heads have near-zero mean-field deviation, which is because their attention patterns are \emph{entirely type-predictable}: they always attend to themselves.  The other set is \hPrev{L0H7}, \hPrev{L1H0}, \hPrev{L2H3}, and \hPrev{L2H5}. These are previous-token heads that always attend one position back. For such a head its corpus average is just a bigram statistic -- which the mean-field kernel consequently encodes. Mean-field deviation is again small as a result.

\begin{table}[t]
\centering\footnotesize\setlength{\tabcolsep}{3.5pt}
\begin{tabular}{rlcccc l l}
\toprule
\# & head & $\bar S^{\ell h}$ & $M_{\mathrm{cov}}$ & $M_{\mathrm{ctx}}$ & induction & documented role & source \\
\midrule
1  & \hPos{L1H8}   & 0.412 & $+0.21$ & $+0.21$ & 0.00 &  positional  & this paper \\
2  & \hPos{L1H2}   & 0.411 & $+0.21$ & $+0.21$ & 0.00 &  positional  & this paper \\
3  & \hPos{L1H3}   & 0.392 & $+0.34$ & $+0.05$ & 0.00 &  positional  & this paper \\
4  & \hOther{L6H2} & 0.359 & $+0.14$ & $+0.22$ & 0.00 & -- & \\
5  & \hPos{L2H7}   & 0.354 & $+0.21$ & $+0.14$ & 0.00 &  positional  & this paper \\
6  & \hCont{L5H2}          & 0.344 & $+0.21$ & $+0.13$ & 0.00 & content mover & this paper \\
7  & \hPos{L1H10}  & 0.341 & $+0.15$ & $+0.19$ & 0.01 &  positional  & this paper \\
8  & \hCont{L9H8}  & 0.312 & $+0.11$ & $+0.20$ & 0.02 & content mover & this paper \\
9  & \hInd{L5H5}   & 0.288 & $+0.17$ & $+0.11$ & \textbf{0.93} & induction (IOI) & \cite{wang2023ioi} \\
10 & \hOther{L0H11}& 0.269 & $+0.11$ & $+0.16$ & 0.01 & -- & \\
11 & \hMover{L9H9} & 0.262 & $+0.12$ & $+0.14$ & 0.45 & name mover; letter mover & \cite{wang2023ioi,garciacarrasco2024acronyms} \\
12 & \hMover{L9H6} & 0.260 & $+0.10$ & $+0.16$ & 0.44 & name mover & \cite{wang2023ioi} \\
13 & \hMover{L10H6}& 0.258 & $+0.08$ & $+0.18$ & 0.28 & backup name mover & \cite{wang2023ioi} \\
14 & \hCont{L8H3}  & 0.257 & $+0.13$ & $+0.12$ & 0.01 & content mover & this paper \\
15 & \hInd{L6H9}   & 0.257 & $+0.16$ & $+0.10$ & \textbf{0.92} & induction (IOI) & \cite{wang2023ioi} \\
16 & \hCont{L6H8}  & 0.255 & $+0.15$ & $+0.11$ & 0.00 & content mover & this paper \\
17 & \hInd{L7H10}  & 0.250 & $+0.15$ & $+0.10$ & \textbf{0.90} & induction (non-IOI)$^\dagger$ & using metric from~\citep{olsson2022induction} \\
18 & \hMover{L10H0}& 0.248 & $+0.08$ & $+0.17$ & 0.35 & name mover & \cite{wang2023ioi} \\
19 & \hOther{L9H4} & 0.248 & $+0.11$ & $+0.14$ & 0.06 & `particle/pronoun bigram'$^\ddagger$ & \cite{kissane2024everyhead} \\
20 & \hMover{L10H10}& 0.246 & $+0.09$ & $+0.16$ & 0.15 & backup name mover & \cite{wang2023ioi} \\
\bottomrule
\end{tabular}
\caption{The twenty GPT-2 heads with the largest mean-field deviation,
rank-ordered by $\bar S^{\ell h}$ \eqref{eq:ddec}, decomposed into
$M_{\mathrm{cov}}$ vs.$\,M_{\mathrm{ctx}}$ \eqref{eq:head-meters}.
$^\dagger$induction is the behavioral class of~\citet{olsson2022induction};
$^\ddagger$unresolved by our analysis, label from the SAE survey at $5/10$
confidence.}
\label{tab:ranking-cc} 
\end{table}

\paragraph{Circuits straddle the ranking.}
A \emph{circuit} is a group of heads that compose across layers to carry
out a function: an earlier head writes some structure into the residual
stream, and a later head reads it back to produce its output.  An induction circuit
composes a previous-token head with an induction
head~\citep{Elhage2021Framework,olsson2022induction}, and the IOI circuit
composes duplicate-token heads with name movers~\citep{wang2023ioi}; in each the
first head writes and the second reads.  Both pairings straddle our ranking.  Three \hInd{induction
heads} sit in the top twenty while the
previous-token heads \hPrev{L2H5}, \hPrev{L2H3} and \hPrev{L1H0} that feed them
sit near the bottom; the \hMover{name movers}
rank high while the duplicate-token heads
\hDup{L3H0} and \hDup{L0H1} that feed them rank near last. 
We
hypothesize this as a principle needed for ICL.  In-context retrieval
is a \emph{write-then-read} computation: a first stage lays down
retrievable structure by applying the same local rule at every position
(tag each token with its predecessor; flag duplicates), and a second
stage dereferences that structure conditionally.  The first stage must be
uniform to be reliable, since it runs before the query is known, and
uniform behavior conditioned on type is what the mean-field
kernel predicts, so a tagging head's write deviation is small.  
The second stage's output is the novel
in-context binding, which corpus statistics cannot supply, so its
deviation is necessarily large.  $\Dmf$ thus meters where contingency
enters a computation, and a lookup circuit appears in the ranking twice:
its invariant half at the bottom, its contingent half at the top.
This hypothesis accords with existing two-stage readings of in-context learning.
Induction is described as a paired write/read, or associative-memory,
mechanism~\citep{olsson2022induction,bietti2023birth}.
In label-anchor accounts, shallow layers aggregate context into anchor positions that
deeper layers then read out~\citep{wang2023anchors}.
And in the same Gemma-2-2B model studied here, the contextualize-then-aggregate
circuit account has early layers write contextual information into exemplar positions
that later layers aggregate into the task
representation~\citep{bakalova2025contextualize}. 
While these accounts all describe the same 
division of labor, what mean-field analysis adds is a principle, grounded in a corpus-level statistic, that
separates the two roles without specifying a task.

\subsection{Task-Specific Deviations Are Function Vectors}
\label{sec:fv}
So far, we have been measuring a head's contribution on a specific prompt
against a corpus average, and using the deviation as a diagnostic; in \S\ref{sec:catalog-findings}, we used it
to locate, decompose, and characterize contextual-processing heads.
However, the mean field deviation becomes something we can actively inject when it is measured against a \emph{task}.

\citet{todd2024fv} showed that in large models, GPT-J among them, a small set of attention heads carries a transportable `function vector': the sum of those heads' outputs on few-shot prompts of a task, added to the residual stream of an unrelated prompt, causes the model to perform the task.
For example, an antonym vector can be extracted from few-shot prompts that
cue the task with an answer-cuing token `\texttt{:}'
(`\texttt{hot:~cold, fast:~slow, \ldots}'), then injected into a zero-shot prompt
with no examples (`\texttt{good:}'), where it triggers the model answer
`\texttt{bad}'.
\citet{todd2024fv} find attention heads through causal mediation, a supervised search over patching interventions.
We construct an alternative function vector extraction pipeline using mean-field deviation, and compare the two methods in depth.

\paragraph{Measuring deviation against a task.} 
As in \S\ref{sec:ranking}, a head's contribution is averaged over contexts \emph{before} the deviation is taken.  Here, however, we average the head's contribution over the task's few-shot prompts rather than over the corpus, and keep the difference as a vector instead of reducing it to the scalar deviation
of \eqref{eq:ddec}.
For a prompt ending in the answer-cuing token `\texttt{:}',
we define for each head its \emph{task-specific deviation}
\begin{equation}
  r^{\ell h} \;=\; \E_{X \in \text{task prompts}}\Big[\v z^{\ell h}_{\text{`:'},X}\Big]
    \;-\; \hat{\v z}^{\ell h}_{\text{`:'}},
  \label{eq:taskresid}
\end{equation}
the average over the task's prompts $X$ of the head's contribution at the final `\texttt{:}', minus the \emph{corpus} mean-field contribution $\hat{\v z}^{\ell h}_s$ evaluated at $s = \text{`:'}$.

An extraction pipeline has two components: \emph{head selection} and \emph{vector construction}.
For head selection, mean-field analysis ranks heads by the norm of the
deviation vector $\|r^{\ell h}\|$; those whose task-prompt output departs most from their corpus expectation are ranked highest.  \citet{todd2024fv}  select heads by causal mediation.
They take a head's average activation on clean task prompts, patch it into label-corrupted prompts,
and score the head by how much that restores the answer.  They call this the {\em average indirect effect} (AIE).
For vector construction, mean-field analysis injects the deviation vector
$r^{\ell h}$ itself, while \citeauthor{todd2024fv}\ take the raw sum of the chosen heads' outputs.  Unlike the patching search used by \citeauthor{todd2024fv}, the mean-field measurement needs no contrast prompts and just one forward pass per prompt.

These two differences raise two questions.  (1)~For either piece of the pipeline, does one approach dominate the other?  (2)~Which 
full pipeline transports function vectors
better:
(AIE + raw averages) or (MFA residual norms + deviation
vectors)? We answer these below, with full results in Appendix~\ref{app:fv}.

\paragraph{Setup.}  We consider Gemma-2-2B and GPT-J-6B,
with six tasks spanning a range of difficulty: the antonym, capital and
past-tense tasks of \S\ref{sec:icl-correlates}, plus
country$\to$language, country$\to$currency, and English$\to$French word
translation.
Each task draws on $60$ cue--answer pairs, split three times into different $20$ extraction
and $40$ held-out evaluation pairs.
We score a method by the fraction of the
held-out queries the model answers correctly when its extracted vector is injected
into a zero-shot prompt, averaged over the three splits ($n=120$; binomial $95\%$
confidence intervals roughly $\pm0.03$--$0.09$).
The MFA head selection is recomputed from each split's extraction prompts
(it is stable: $7$--$10$ of ten heads shared across splits), while the AIE
selection is held fixed from a single mediation sweep per task.
Injection follows \citet{todd2024fv}: the selected heads' vectors are summed
into one vector and added to the residual stream at a single layer and the final
token position (their Eqs.~5 and~14).  We fix the injection layer and scale by one
sweep and hold them fixed across all methods, so none is favored by tuning.
Appendix~\ref{app:fv} gives the sweep, the per-head-injection and GPT-J protocols,
and confirms that the protocol choice does not affect the comparison.

\begin{table}[th]
\centering\small
\begin{tabular}{lc cc @{\hskip 1.5em} cc}
\toprule
 & & \multicolumn{2}{c}{AIE-selected heads} & \multicolumn{2}{c}{MFA-selected heads} \\
\cmidrule(lr){3-4}\cmidrule(lr){5-6}
task & AIE$\cap$MFA & raw & MFA $r$ & raw & MFA $r$ \\
\midrule
antonym     & 2/10 & 0.57 & \textbf{0.82} & 0.79 & 0.76 \\
capital     & 2--3/10 & \textbf{0.91} & \textbf{0.91} & 0.01 & 0.01 \\
past-tense        & 3/10 & 0.47 & 0.87 & \textbf{0.96} & 0.93 \\
language    & 2/10 & 0.27 & \textbf{0.40} & \textbf{0.40} & 0.37 \\
currency    & 2--3/10 & 0.07 & \textbf{0.38} & 0.00 & 0.00 \\
translation & 2/10 & 0.00 & 0.10 & \textbf{0.38} & 0.26 \\
\bottomrule
\end{tabular}
\caption{Gemma-2-2B: zero-shot transfer for each combination of head
selection and vector construction.  AIE$\cap$MFA is the number of heads the selections share; a range means it varies across the three
splits.
The best result per task is in bold.
}
\label{tab:fv-matrix-body}
\end{table}

\paragraph{Results.}  Table~\ref{tab:fv-matrix-body} reports zero-shot
transfer for each combination of head selection (AIE vs.\ MFA) and vector
construction (raw sum vs.\ MFA $r$) for Gemma-2-2B.  In this model,
the overlap between the two head selection methods is low and neither consistently performs best. %
The heads selected also sit
at different depths:  AIE in a compact mid-stack group (L12--L15 with head L14H0 rank~1 on
all six tasks) and MFA in an earlier, more task-varying L6--L9 band. 
On GPT-J, the two selections more closely coincide (MFA selects
$4$--$7$ of Todd's ten heads) and transfer rates tie on each of the tasks considered
(Table~\ref{tab:fvj-matrix} in Appendix~\ref{app:fv}).

For vector construction, an initial observation is that again neither method is superior, with an even split across tasks.  But, closer inspection shows the surprising result that the best
pipeline for a task almost always \emph{mixes} the two methods, taking the head
selection from one method and the injection vector from another.
On antonym, for
instance, AIE selection coupled with the MFA deviation vector reaches $0.82$, above either
aligned pairing ($0.57$ for AIE, $0.76$ for MFA). 
Further study of these apparent complementarities could be a useful direction for future work.

\paragraph{GPT-2: category vectors.}
At GPT-2 small's scale, injecting either construction into a zero-shot prompt
never gets the model to produce that prompt's own answer,
and thus Todd et al.\ did not
evaluate function vectors in models at GPT-2's scale.
What the deviation vector transports instead is the task's \emph{category};
\S\ref{app:fv-gpt2} gives the measurements behind the following
results.
On capital, the injected model names a capital city on all $120$ held-out queries,
but essentially never the one the query asks for.
The raw average produces no task answer at all, on any query.

We can see that the deviation vector's content nonetheless indicates the whole category rather than the
remembered answers, so it is not
simply parroting.
Decoding the injected vector with the J-lens of \citet{gurnee2026workspace}, on
GPT-2's best capital head, the capitals seen in the prompts, the held-out capitals
and \emph{capitals belonging to neither set} all rank high in the vocabulary (median
ranks $9$, $37$ and $34$, against ${\sim}21{,}000$ for unrelated words).
The small advantage that observed capitals hold translates into their being output as the top prediction.
Gemma-2's vectors do not have this behavior.
At the scale of GPT-2, then, the vector fixes the \emph{kind} of answer the model should give but not the correct answer, and leans towards 
categorical responses it has previously seen.

\section{Conclusions} \label{sec:conc} 
Mean-field analysis describes a model’s average behavior relative to a chosen corpus, so our results are necessarily conditional on the corpora studied.  While we compare results across corpora (and find little difference in results among them), there are many diverse corpora that may be considered as a basis for mean-field analysis. A fruitful direction for future work is understanding how different corpora induce different attention mean-fields, and how an appropriate corpus for training the mean field can be chosen in any given analysis setting.

To summarize, we make four contributions. First, we develop the conceptual framework needed to precisely describe the average-case attention of a head.   Second, we show that, using this operator, one can make accurate predictions of average representation evolution through the layers of a model.    Third, we show that in early training stages, models behave like their mean fields. Finally, mean-field deviation provides a task-agnostic measure of context-specific computation: it recovers known ICL machinery and yields task residuals that act as transferable function vectors. 

\paragraph{Acknowledgements.}  Mark Crovella was supported by NSF grants CNS-2312711 and CNS-2319369, and by a grant from Coefficient Giving.

\label{sec:ai-disclosure}
\section*{Use of AI}

The authors used various versions of Anthropic's Claude and OpenAI's ChatGPT models throughout this work.  An explicit goal of this research project was to explore the limits of publicly available generative AI capabilities (circa 2026) for conducting  research in mechanistic interpretability.
Under the authors' guidance, Claude wrote the experimental code, ran and re-ran numerous experiments, drafted preliminary analyses of experimental results, and suggested follow-on experiments.
In many cases, validated findings from those experiments appear in this manuscript. 
In close collaboration with the authors, Claude and ChatGPT reviewed and copy-edited portions of the manuscript for correctness, readability and concision.
The research questions, experimental designs, and the interpretations reported here are the authors'.
The authors take full responsibility for the content of this paper.

\bibliography{refs}
\bibliographystyle{plainnat}
\clearpage
\appendix \section{Conditional Mean Fields with Repeated Types}
\label{app:full-conditional-meanfield}

Section~\ref{sec:context-mf} derived the $s,t$-conditional and
context-conditional mean fields under the simplifying assumption that a type occurs at
most once in a context.  This appendix removes that assumption.  The structure of the
derivation is unchanged: the same two mean fields are reached by the same two steps,
but one additional assumption is required, governing how the attention a query places
on a type grows with the number of copies of that type present.  Setting every count to
zero or one recovers Section~\ref{sec:context-mf} exactly.

\paragraph{Notation for repeated types.}
Recall from Section~\ref{sec:meanfield} that $n_t(X)$ is the number of positions
of type $t$ available to the query in $X$, and write $n_t$ for the corresponding random variable when $X$
is drawn from the corpus.  We write $\Pr_s(n_t = n)$ for the probability that a random
context with query $s$ contains exactly $n$ copies of $t$, and $\mathbb{E}_{stn}$ for
the conditional expectation over such contexts.  The expected number of type-$t$
positions in a type-$s$ query's context is
\[
\overline{n}_{st} \;=\; \mathbb{E}_s[n_t] \;=\; \sum_{n \ge 1} n \,{\Pr}_s(n_t = n).
\]
When a type can occur at most once, $n_t \in \{0,1\}$, so
$\overline{n}_{st} = \Pr_s(t)$ and $\mathbb{E}_{st1} = \mathbb{E}_{st}$; every expression
below then reduces to its counterpart in Section~\ref{sec:context-mf}.

\paragraph{The $s,t$-conditional mean field.}

Start from the exact contribution to the centroid \eqref{eq:exact-contribution-by-type}:
\[
\overline{\v z}_s = \sum_{t\in\mathcal{T}} \mathbb{E}_s[A_{st}\overline{\v v}_t]
\]

We now condition on the \emph{number} of copies of $t$ in the context rather
than merely on the presence of $t$.  Partitioning on the value of $n_t$,
\[
\mathbb{E}_s[A_{st}\overline{\v v}_t] = \sum_{n \ge 0}
   \mathbb{E}_{stn}[A_{st}\overline{\v v}_t] \,{\Pr}_s(n_t = n)
   \;=\; \sum_{n \ge 1} \mathbb{E}_{stn}[A_{st}\overline{\v v}_t] \,{\Pr}_s(n_t = n),
\]
where the $n = 0$ term drops out because $A_{st} = 0$ when $t$ is absent from
the context, exactly as before.

Denote $\mathbb{E}_{stn}[A_{st}\overline{\v v}_t]$ as
$\overline{\v z}_{stn}$.  This is the average contribution to the $s$ token centroid by
token $t$ when exactly $n$ copies of $t$ are in the context: the influence measure
$\overline{\v z}_{st}$ of Section~\ref{sec:context-mf}, resolved by multiplicity.
Then
\[
\overline{\v z}_s = \sum_{t\in\mathcal{T}} \sum_{n \ge 1}
   \overline{\v z}_{stn}\, {\Pr}_s(n_t = n)
\]
which is again an exact decomposition of the corpus contribution, now refined by
multiplicity.  
We rewrite $\overline{\v z}_{stn}$ as
\[
\overline{\v z}_{stn} = \mathbb{E}_{stn}[A_{st}]\,\mathbb{E}_{stn}[\overline{\v v}_t]
  + \operatorname{Cov}_{stn}(A_{st}, \overline{\v v}_t)
\]
and impose Assumption~\ref{ass:value-independence-st} on this finer conditioning
set, $\operatorname{Cov}_{stn}(A_{st}, \overline{\v v}_t) \approx 0$, giving
\begin{equation}
\overline{\v z}_{stn} \approx \mathbb{E}_{stn}[A_{st}]\,\mathbb{E}_{stn}[\overline{\v v}_t].
\label{eq:zbar-factored-appendix}
\end{equation}

Assumption~\ref{ass:value-query-indep} carries over with one addition: the
average value map of token $t$ is taken to be independent not only of the query type but
also of how many copies of $t$ are present.

\begin{mfass}{Value-query independence (extended to repeated types)}
\label{ass:value-context-indep-context-conditional}
$\mathbb{E}_{stn}[\overline{\v v}_t] \approx \boldsymbol\mu_t$, the corpus
average value map of token $t$, for every $s$ and every $n$.  
\end{mfass}

The quantity $\mathbb{E}_{stn}[\overline{\v v}_t]$ is an
average over type-$s$ query contexts, and $\overline{\v v}_t$ is itself
attention-weighted within the context, so it differs from $\boldsymbol\mu_t$ by the degree to which type
$t$'s representation is contextualized in the company of $s$.
The other factor of \eqref{eq:zbar-factored-appendix}, $\mathbb{E}_{stn}[A_{st}]$, is where repeated types
enter: it is a separate quantity for each multiplicity $n$, whereas
Section~\ref{sec:context-mf} had the single number $W_{st} = \mathbb{E}_{st}[A_{st}]$.
Here, we recast $W_{st}$ as a  as a rate \emph{per key}
rather than per context,
\[
W_{st} \;:=\; \frac{\mathbb{E}_s[A_{st}]}{\mathbb{E}_s[n_t]}
  \;=\; \frac{P_{st}}{\overline{n}_{st}},
\]
the total attention a type-$s$ query places on type $t$, divided by the expected
number of type-$t$ keys that mass was spread over.  This is a definition rather than an
assumption, and it is directly measurable: numerator and denominator are each a single
corpus pass.  When a type occurs at most once it reduces to
$W_{st} = P_{st}/\Pr_s(t) = \mathbb{E}_{st}[A_{st}]$, the quantity of
Section~\ref{sec:context-mf}.  Our final  assumption relates how attention mass grows with multiplicity:

\begin{mfass}{Homogeneity of attention}
\label{ass:homogeneity}
$\mathbb{E}_{stn}[A_{st}] = n\,W_{st}$.  The attention a type-$s$ query places
on type $t$ is proportional to the number of copies of $t$ present, at a per-key rate
that does not otherwise depend on the context. 
\end{mfass}

Assumption~\ref{ass:homogeneity} is purely a linearity
claim: averaging it back over $n$ returns
$\sum_{n} n\,W_{st}\,\Pr_s(n_t = n) = W_{st}\,\overline{n}_{st} = P_{st}$, which holds
identically by the definition of $W_{st}$.  What is being assumed is therefore not the
overall level of the attention but its \emph{distribution across multiplicities}:
that a context holding five copies of $t$ draws five times the mass of one holding a
single copy, rather than, say, saturating.

Under the three assumptions,
\[
\overline{\v z}_s \approx \hat{\v z}_s = \sum_{t\in\mathcal{T}}
  W_{st}\,\boldsymbol\mu_t \sum_{n \ge 1} n \,{\Pr}_s(n_t = n)
  \;=\; \sum_{t\in\mathcal{T}} W_{st}\,\boldsymbol\mu_t\,\overline{n}_{st},
\]
the repeated-type counterpart of the $s,t$-conditional mean field of
Section~\ref{sec:context-mf}: the presence probability $\Pr_s(t)$ is replaced by the
expected count $\overline{n}_{st}$, and nothing else changes.

\paragraph{The context-conditional mean field.}
We now repeat the second construction of Section~\ref{sec:context-mf} for a
particular context $X$ with query $s$, again allowing a type to occur any number of
times in $X$.  As before, $W_{stX}$ denotes the
attention paid by $s$ to $t$ in context $X$, and $\boldsymbol\mu_{tX}$ the average
value map of token $t$ in that context.

Note that $W_{stX}$ is the \emph{total} attention the
query places on type $t$, summed over all $n_t(X)$ of its positions in the prefix; it is
$A_{st}$ of \eqref{eq:exact-contribution-by-type} evaluated on this particular context.
Likewise $\boldsymbol\mu_{tX}$ is the attention-weighted mean of the value vectors at
those same $n_t(X)$ positions, which is $\overline{\v v}_t$ evaluated on this context.
Both therefore already account for every copy of $t$ in $X$.

Conditioning on $X$ fixes every count,
so $\Pr_X(t)$ is again $1$ for the tokens in $\supp(X)$ and $0$ otherwise, and
\[
\v z_{sX} = \sum_{t\in\mathcal{T}} W_{stX}\,\boldsymbol\mu_{tX}\,{\Pr}_X(t)
  \;=\; \sum_{t \in \supp(X)} W_{stX}\,\boldsymbol\mu_{tX}.
\]

Next, as before, we make the \textbf{assumption} that
$\boldsymbol\mu_{tX} = \boldsymbol\mu_t$, again noting that the difference between them is the \emph{contextualization}
of $t$ induced by its context $X$, leaving
\[
\hat{\v z}_{sX} = \sum_{t \in \supp(X)} W_{stX}\,\boldsymbol\mu_t.
\]

And as before, since $X$ is a fixed context, $W_{stX}$ is simply the attention
$A_{stX}$ that the head pays from $s$ to $t$ in this context.  To construct a
mean-field analog we form an \textbf{approximation} to $A_{stX}$ using renormalization,
now weighting each type by the number of copies it contributes:
\begin{equation}
W_{stX} \approx \hat{A}_{stX} = \frac{n_t(X)\,W_{st}}
  {\sum_{t'\in \supp(X)} n_{t'}(X)\,W_{st'}},
  \label{eq:w-def-appendix}
\end{equation}
the numerator being the mass that Assumption~\ref{ass:homogeneity} predicts for
a context holding $n_t(X)$ copies of $t$.  This yields the \emph{context-conditional}
mean field approximation for contexts with repeated types,
\[
\hat{\v{z}}_{sX} = \sum_{t \in \supp(X)}\hat{A}_{stX}\, \boldsymbol\mu_t,
\]
which reduces to the expression of Section~\ref{sec:context-mf} when every
$n_t(X)$ equals $1$.  Note that the multiplicities do not cancel from the
renormalization: they reweight the types relative to one another, so the prediction
depends on $X$ through the full multiset of types it contains and not merely through
$\supp(X)$.
\section{Analysis of the Context-conditional Mean Field}
\label{app:ccmf-analysis}

\subsection{The Kernel \texorpdfstring{\Ahat}{Ahat} as a Mean Field}
The context-conditional mean field underlies \S\ref{sec:waypoint} and \S\ref{sec:icl}. Its kernel
renormalizes the corpus kernel over the types a context actually contains \eqref{eq:w-def-appendix}:
\begin{equation}
  \Ahat_{stX} \;=\; \frac{n_t(X)\,W_{st}}{N_X},
  \qquad
  N_X \;=\; \sum_{t'\in\supp(X)} n_{t'}(X)\,W_{st'},
  \label{eq:ahat}
\end{equation}
where $A_{stX}$ is the \emph{total} attention the head pays from a query of type
$s$ to all $n_t(X)$ positions of type $t$, and $W_{st}$ defined as $P_{st}/\bar n_{st}$
(Appendix \ref{app:full-conditional-meanfield}) with $P_{st}=\E_s[A_{st}]$ and $\bar n_{st}=\E_s[n_t(X)]$.

Renormalization is a natural way to turn a corpus-average propensity into
something that sums to one over a particular context, but it introduces a nonlinearity into 
the definition of \Ahat.  Nonetheless, the choice of \Ahat\ can be justified if 
$\Ahat$ is an
\emph{unbiased estimator},
that is, if
\begin{equation}
  \E_{st}\!\left[A_{stX}-\Ahat_{stX}\right] = 0.
  \label{eq:21}
\end{equation}

Here $\Ahat_{stX}$ varies with the context through its token counts, while
$\E_{st}[A_{stX}]$ is fixed, so \eqref{eq:21} asks the random quantity to match
the constant in expectation.

The remainder of this section shows the following:
\begin{enumerate}
    \item Equation \eqref{eq:21} does \emph{not} hold in general; we show why and we show how to estimate its error.
    \item Equation \eqref{eq:21} \emph{does} hold approximately in practice for the data we study in this paper.  
    \item It is possible to define an \emph{unbiased} replacement for $W$. Specifically, we show how to construct a matrix $U$, such that if we use $U$ in place of $W$ in the definition of \Ahat, then \eqref{eq:21} holds by construction. When we replace $W$ with its unbiased version $U$ in our experiments, \emph{our key results do not change.}
    \item Since Equation \eqref{eq:21} holds approximately in practice for our datasets, that justifies the splitting of context-conditional mean field deviation into the components given in \eqref{eq:dev-decomp}, which we call \emph{covariance} and \emph{contextualization}. 
\end{enumerate}

In the remainder of this appendix we typically omit the context $X$ and
abbreviate the per-context quantities $A_{stX}$, $\Ahat_{stX}$ and $n_t(X)$ as
$A_{st}$, $\Ahat_{st}$ and $n_t$.  We restore the explicit $X$ (as in $A_{stX}$,
$n_t(X)$, $N_X$) at the steps where the dependence on the particular context is
the point.  The expectations $\E_s$ and $\E_{st}$ average over the contexts $X$,
and the corpus quantities $P_{st}$, $W_{st}$, $\bar n_{st}$ and $\pi_{st}$ carry
no context argument.

\subsection{Analysis of Bias in \texorpdfstring{\Ahat}{Ahat}}

First, we give some insight into the nature of the bias in \Ahat.
Because $A_{stX}=0$ and $n_t(X)=0$ together whenever $t$ is absent, writing
$\pi_{st}=\Pr_s(t\in\supp(X))$ gives $\E_s[A_{st}]=\pi_{st}\E_{st}[A_{st}]$ and
$\E_s[n_t]=\pi_{st}\E_{st}[n_t]$. The presence probability therefore cancels
from the definition of $W$:
\begin{equation}
  W_{st}\;=\;\frac{P_{st}}{\bar n_{st}}
  \;=\;\frac{\E_s[A_{st}]}{\E_s[n_t]}
  \;=\;\frac{\E_{st}[A_{st}]}{\E_{st}[n_t]}
\end{equation}
which means that
\begin{equation}
  \E_{st}[A_{st}]\;=\;W_{st}\,\E_{st}[n_t] .
  \label{eq:target}
\end{equation}
From \eqref{eq:ahat}, $\E_{st}[\Ahat_{st}]=W_{st}\,\E_{st}[n_t/N_X]$.
Subtracting from \eqref{eq:target},
\begin{equation}
  \E_{st}\!\left[A_{st}-\Ahat_{st}\right]
  \;=\;W_{st}\Big(\E_{st}[n_t]\;-\;\E_{st}\!\left[n_t/N_X\right]\Big).
  \label{eq:bias}
\end{equation}
Writing $1/N_X = 1+(1-N_X)/N_X$ in \eqref{eq:bias},
\begin{equation}
  \E_{st}[n_t/N_X]-\E_{st}[n_t]
  \;=\;-\,\E_{st}\!\left[\frac{n_t\,(N_X-1)}{N_X}\right],
  \label{eq:corr}
\end{equation}

More precisely, the deficit splits into two terms of opposite
sign,
\begin{equation}
  \E_{st}[n_t/N_X]-\E_{st}[n_t]
  \;=\;\E_{st}[n_t]\big(\E_{st}[1/N_X]-1\big)
  \;+\;\operatorname{Cov}_{st}\!\big(n_t,\,1/N_X\big).
  \label{eq:twoterm}
\end{equation}
The first is a Jensen term: $1/N_X$ is convex, so $\E_s[1/N_X]>1/\E_s[N_X]\approx1$,
pushing toward
\emph{over}-prediction. For example, in GPT-2 measured $\E_s[1/N_X]$ is $1.36$ while $\E_s[N_X]$ is $0.99$.
The second is negative: both $n_t(X)$ and $N_X$ grow
with the length of the context, so $n_t$ and $1/N_X$ are anticorrelated,
pushing toward \emph{under}-prediction. For every bucket whose multiplicity
grows with length (all tracked types, and the ``other'' bucket), the
covariance term is larger, and $\Ahat_{st}$ under-predicts. The one exception is the
position-$0$ bucket, whose multiplicity is identically $1$: its covariance
term vanishes exactly, leaving the pure Jensen term, and it is
over-predicted.  

Both $A$ and $\Ahat$ are
distributions over the same context: $\sum_t A_{stX}=\sum_t\Ahat_{stX}=1$, hence
\begin{equation}
  \sum_t \E_s\!\left[A_{st}-\Ahat_{st}\right]\;=\;0
  \label{eq:sumzero}
\end{equation}
exactly, for every $s$. It's useful to define the per-cell \emph{relative} bias
\begin{equation}
   r_{st}=1-\E_{st}[\Ahat_{st}]/\E_{st}[A_{st}].
\end{equation}
This establishes a connection with $P$. The identity \eqref{eq:sumzero} then reads
$\sum_t P_{st}\,r_{st}=0$, since
$\E_s[A_{st}-\Ahat_{st}]=\pi_{st}\E_{st}[A_{st}]\,r_{st}=P_{st}\,r_{st}$.
In other words, the relative biases cancel when each is weighted by the average attention
$P_{st}$ its pair carries.

The bias is thus a redistribution of attention mass among key types, not
a net distortion: because it sums to zero under the same $P$-weighting that
governs the mean-field prediction, it cannot accumulate in the aggregate
quantities the paper reports.  \S\ref{sec:route1} shows the per-pair bias is
also small empirically, and \S\ref{sec:route2} removes it entirely with a
calibrated estimator.

\subsection{The Measured Bias of \texorpdfstring{\Ahat}{Ahat} is Small}
\label{sec:route1}

To assess the empirical bias of $\Ahat_{st}$, we proceed as follows.  For each corpus position
whose query type is tracked, and each head, we compute $N_X$ over the inclusive
causal prefix and accumulate
per (head, type) the conditional means $\E_{st}[n_t]$ and $\E_{st}[n_t/N_X]$.\footnote{Corpus: $389$ chunks of $512$ OpenWebText tokens ($\sim$200k).}
The quantity  $\E_{st}[n_t/N_X]\big/\E_{st}[n_t] = \E_{st}[\Ahat_{st}]/\E_{st}[A_{st}]$ by \eqref{eq:bias}; a ratio
$1$ is unbiased.
The results are in Table~\ref{tab:ahat-empirical-bias}.  The table shows that 
the unweighted cell average is $0.946$ and $0.933$, a $5$--$7\%$
under-prediction for the typical pair, with over $95\%$ of pairs within
$20\%$ of unbiased in both models. $\E_s[N_X]\approx1$ is a consistency check on $P$ and $\bar n$.

\begin{table}[h]
\centering\small
\begin{tabular}{lrrrr}
\toprule
model & $\E_s[N_X]$ & $\E_{st}[n_t]$ &
 $\E_{st}[n_t/N_X]\big/\E_{st}[n_t]$ & cells within 20\% of 1.0 \\
\midrule
GPT-2       & 0.988 & 1.468 & \textbf{0.946} & 95.5\% \\
Gemma-2-2B  & 0.995 & 1.466 & \textbf{0.933} & 96.8\% \\
\bottomrule
\end{tabular}
\caption{Observed bias of (head,type) pairs.  ``Cells within 20\%'' counts well-sampled (head, type) pairs
whose ratio lies in $[0.8,1.2]$: 129k cells for GPT-2, 187k for Gemma.
}
\label{tab:ahat-empirical-bias}
\end{table}

Empirically, in GPT-2 we observe that correlation between $n_t$ and $(N_X - 1)/N_X$ is quite similar for 
different token types $t$.  
Binning types into quartiles by presence frequency (the number of scored query
positions whose causal prefix contained the type), with the top quartile split
at the $90$th percentile, every bin's mean ratio lies in $[0.93,0.95]$, with
${\approx}11\%$ of cells above $1$ in each. The compensation that
\eqref{eq:sumzero} requires therefore sits in the untracked bucket: the
position-$0$ bucket has $n=1$ by definition (hence no covariance term by
\eqref{eq:twoterm}), ratio $\E[1/N_X]=1.36$, and ${\approx}40\%$ of the
attention mass.  Thus, \Ahat\ has one over-predicted sink balancing a uniformly
under-predicted field. The same structure holds in Gemma-2-2B: tracked types
uniform at $0.93$--$0.94$ across every frequency quartile, 
and the position-0 sink over-predicted at $1.22$ on $40\%$ of the
mass.  A separate run comparing $\Ahat$ against the model's
measured attention reproduces the deficit, confirming that \eqref{eq:bias} is
the whole story.

This small bias has negligible effect on our results.  We demonstrate this as follows.  In the next section we show how to construct a truly unbiased estimator of \Ahat.  We then show that, using this unbiased estimator, the key results in the paper are unchanged.

\subsection{Using an Unbiased Estimator of \texorpdfstring{\Ahat}{Ahat} Does Not Change Key Results}
\label{sec:route2}

The bias in \eqref{eq:bias} exists because $W$ is obtained in closed form, as
the ratio $P/\bar n$ of two \emph{marginal} means, while it is \emph{used}
inside a nonlinear per-context renormalization. 

\paragraph{Constructing an unbiased estimator of \texorpdfstring{\Ahat}{Ahat}.} 
Instead of using $W$ in the definition of \Ahat, we may instead find 
a matrix $U$ that makes \eqref{eq:21} true by construction.
Multiplying \eqref{eq:21} by
$\pi_{st}$ to remove the conditioning, the requirement is
\begin{equation}
  \E_s\!\left[\Ahat_{st}(U)\right] \;=\; \E_s[A_{st}] \;=\; P_{st}
  \qquad\text{for every } (s,t),
  \label{eq:fixedpoint}
\end{equation}
i.e.\ \emph{the kernel must reproduce the measured attention marginals when run
through its own renormalization.} Any $U$ satisfying \eqref{eq:fixedpoint}
gives $\E_{st}[A_{st}-\Ahat_{st}]=0$ exactly, for every pair, by construction.

We note that, for each query type $s$, achieving \eqref{eq:fixedpoint} is a marginal-matching problem.
To see this, observe that for each context $X$ containing $s$, \eqref{eq:ahat} makes
$\Ahat_{s\cdot X}(U)$ a probability vector over the key types: type $t$ is offered
$n_t(X)$ times, each occurrence carries weight $U_{st}$, and the vector is the
resulting share.  Averaging those per-context vectors over the corpus gives
$Q_{s\cdot}=\E_s[\Ahat_{s\cdot}(U)]$, again a probability vector over key types ---
the model's \emph{predicted type marginal} for query type $s$. The measured
$P_{s\cdot}$ is the corresponding \emph{observed type marginal}. So
\eqref{eq:fixedpoint} asks for the weights $U$ whose induced marginal, after
per-context normalization and averaging, reproduces the observed one.

Problems of this marginal-matching form are typically solved by Sinkhorn-type iterations \citep{sinkhorn1967concerning,idel2016review}, i.e., iterative
proportional fitting (IPF).
IPF is appropriate because per-context normalization
$\sum_t\Ahat_{stX}=1$ is one constraint family, satisfied automatically by
construction; the target $Q_{s\cdot}=P_{s\cdot}$ is the other. We have
exactly one free parameter $W_{st}$ per constraint, so the system is square. 
The update is correspondingly multiplicative: if the current kernel puts too little mass on
$(s,t)$, scale $U_{st}$ up in proportion to the shortfall and let the
renormalization redistribute the rest.  Applying IPF to this problem is shown as Algorithm~\ref{alg:sinkhorn}. 

\begin{algorithm}[t]
\caption{Unbiased calibration of $W$: the estimator $U$ (run per head; heads are independent)}
\label{alg:sinkhorn}
$U^{(0)} \gets W = P/\bar n$ \tcp*{the paper's estimator: a good warm start}
\For{$k=0,1,2,\dots$}{
  $Q_{st} \gets 0$ for all $(s,t)$;\quad $m_s \gets 0$ for all $s$ \tcp*{corpus accumulators}
  \ForEach{chunk, each query position $p$ with tracked type $s$}{
    $n \gets \big(n_{t}(X)\big)_{t}$ \tcp*{multiplicities of the causal prefix, incl.\ buckets}
    $a_t \gets n_t\, U^{(k)}_{st}$ for all $t$ \tcp*{numerators of \eqref{eq:ahat}}
    $N \gets \sum_{t} a_{t}$ \tcp*{the normalizer $N_X$}
    $Q_{st} \gets Q_{st} + a_t/N$ \quad for all $t$ \tcp*{accumulate $\Ahat_{stX}(U^{(k)})$ into row $s$ only}
    $m_s \gets m_s + 1$\;
  }
  \ForEach(\tcp*[f]{normalize each row by its own occurrence count}){query type $s$}{
    $Q_{st} \gets Q_{st}/m_s$ \quad for all $t$ \tcp*{now $Q_{st}=\E_s[\Ahat_{st}(U^{(k)})]$}
  }
  \lIf{$\max_{s,t}\,|Q_{st}-P_{st}| < \varepsilon$}{\textbf{stop}}
  \ForEach(\tcp*[f]{one multiplicative correction per entry}){pair $(s,t)$}{
    $U^{(k+1)}_{st} \gets U^{(k)}_{st}\cdot
      \Big(\dfrac{P_{st}}{\max(Q_{st},\,\delta)}\Big)^{\alpha}$\;
  }
}
\end{algorithm}

\paragraph{Algorithm.} 

Applying IPF to this problem is shown as Algorithm~\ref{alg:sinkhorn}.
Applying a damping exponent $\alpha < 1$ in
Algorithm~\ref{alg:sinkhorn} is the standard remedy if the residual were to oscillate rather than decrease. Empirically
it is not needed here --- with $\alpha=1$ the residual converges in three iterations with no oscillation,
which is the behavior expected of a contraction.

\paragraph{Using an unbiased estimator of \texorpdfstring{\Ahat}{Ahat} does not change key results.} 
At a fixed point of Algorithm~\ref{alg:sinkhorn}, $Q=P$ so
\eqref{eq:fixedpoint} holds and the bias \eqref{eq:bias} is zero for every
$(s,t)$. Unbiasedness of $\Ahat(U)$ is then true by construction.
Consequently, we can recompute any body result
that uses the context-conditional kernel with a calibrated $U$ in place of
$W=P/\bar n$, holding everything else fixed.
We now do
that for four key results: the head ranking of \S\ref{sec:ranking}, the training-dynamics
gap of \S\ref{sec:waypoint}, the ICL prompt measurements, and the instance-level meter.

\label{sec:U2M}

We first build a calibrated $U$ at $2$M tokens. GPT-2 was calibrated at its published weights and
Pythia-160m at training step $1000$, each against its own $P$ and $\bar n$.  Table~\ref{tab:calib2M} shows the convergence properties of Algorithm~\ref{alg:sinkhorn}, noting that good convergence was achieved in 5 iterations for both cases.
We note that the ratio $U/W^{(0)}$ over the $67.6$M fitted GPT-2 cells has a $5$--$95\%$ range of $[0.76,\,1.29]$ and a median of $1.005$.

\begin{table}[h]
\centering\small
\begin{tabular}{llrrrrr}
\toprule
model & residual & iter 0 & 1 & 2 & 3 & 4 \\
\midrule
\multirow{2}{*}{GPT-2}
  & median $|Q-P|/P$ & 0.1178 & 0.0103 & 0.0014 & 0.0002 & \textbf{0.00003} \\
  & $\max|Q-P|$      & 0.346  & 0.127  & 0.090  & 0.071  & \textbf{0.057} \\
\midrule
\multirow{2}{*}{Pythia-160m (step $1000$)}
  & median $|Q-P|/P$ & 0.1261 & 0.0145 & 0.0028 & 0.0006 & \textbf{0.0001} \\
  & $\max|Q-P|$      & 0.217  & 0.107  & 0.081  & 0.064  & \textbf{0.052} \\
\bottomrule
\end{tabular}
\caption{Calibration at $2$M tokens ($3914$ chunks of $512$), five iterations,
$\alpha=1$. Iteration $0$ is the residual of the present estimator $W = P/\bar n$.
}
\label{tab:calib2M}
\end{table}

Recomputing key results, we find no change to any of our findings.  Specifically:
\begin{description}
    \item[Training Dynamics of \S\ref{sec:waypoint}:] We chose the data point for Pythia-160m at step $1000$ as this lies just at the point where mean-field model performance begins to deviate from true model performance, and applied the calibrated $U$ for that model. At that point, $L_\text{MF}$ is unchanged to three significant figures (invisible on the plot), and $\Delta L$ and $\Dmf$ are also unchanged to three significant figures.
    \item[Experiment 1 of \S\ref{sec:icl-correlates}:] Under the calibrated kernel $U$, every element of the figure survives: the deviation rises across the copy boundary, the single head whose deviation rises most is L5H1, and the in-context loss reduction increases monotonically with the induced deviation from the first decile to the tenth.
    \item[Experiment 2 of \S\ref{sec:icl-correlates}:] All values in Table~\ref{tab:icl} are unchanged to two significant figures.
    \item[Head Ranking of \S\ref{sec:ranking}:] The context-averaged context-conditional mean field was recomputed using the calibrated kernel $U$ for all $144$
GPT-2 heads.  Comparing to Table~\ref{tab:ranking-cc}, we find the new ranking has Spearman correlation of 0.9967, the maximum shift in rank over all heads is 4, and 19 out of the top 20 heads as shown in the table are present in the re-ranked top 20.
\end{description}

The paper's other results (\S\ref{sec:centroid-rollout} and \S\ref{sec:fv}) use the kernel $P$, so are not affected by bias in \Ahat.
We thus conclude that not only is the bias of $\Ahat$ small, as shown here, but that small bias does not affect any of the key results in the paper.

\subsection{Derivation of the Covariance / Contextualization Split}
\label{app:dev-split}
Next we justify the split of the context-conditional mean field into its covariance and contextualization terms \eqref{eq:dev-decomp}.

Fix a head $(\ell,h)$ and a context $X$ with query type $s$, and write $A_{st}$ for $A_{stX}$ as elsewhere in this appendix; all sums run over the present types
$t\in \supp(X)$.
 The head's actual contribution to the query
is $\v z=\sum_t A_{st}\,\overline{\v v}_t$, with $A_{st}$ the attention mass on type $t$ and
$\overline{\v v}_t$ the in-context attention-weighted mean value of the type-$t$ keys. The
context-conditional mean-field replaces the attention by the
renormalized corpus kernel $\hat A_{st}$ \eqref{eq:ahat} and
the in-context value by the corpus type-mean $\boldsymbol\mu_t$
(which Assumption~\ref{ass:value-query-indep} equates, to good
approximation, with $\mathbb E_{st}[\overline{\v v}_t]$), giving
$\hat{\v z}=\sum_t\hat A_{st}\,\boldsymbol\mu_t$. The prediction thus
differs from the actual output in two factors at once: the attention weights
($A_{st}$ versus $\hat A_{st}$) and the values ($\overline{\v v}_t$ versus
$\boldsymbol\mu_t$).  To separate the two changes, insert the \emph{mixed}
term $\sum_t\hat A_{st}\,\overline{\v v}_t$ --- predicted attention on
actual values --- which changes one factor at a time.  Adding and
subtracting this mixed term,
\begin{align}
  \v z-\hat{\v z}
  &=\sum_t A_{st}\,\overline{\v v}_t-\sum_t\hat A_{st}\,\boldsymbol\mu_t \notag\\
  &=\sum_t A_{st}\,\overline{\v v}_t
    \;\underbrace{-\,\sum_t\hat A_{st}\,\overline{\v v}_t
    \;+\,\sum_t\hat A_{st}\,\overline{\v v}_t}_{=\,0}
    \;-\,\sum_t\hat A_{st}\,\boldsymbol\mu_t \notag\\
  &=\underbrace{\Big(\sum_t A_{st}\,\overline{\v v}_t-\sum_t\hat A_{st}\,\overline{\v v}_t\Big)}_{\text{covariance term}}
   +\underbrace{\Big(\sum_t\hat A_{st}\,\overline{\v v}_t-\sum_t\hat A_{st}\,\boldsymbol\mu_t\Big)}_{\text{contextualization term}} \notag\\
  &=\sum_t (A_{st}-\hat A_{st})\,\overline{\v v}_t
   +\sum_t \hat A_{st}\,\big(\overline{\v v}_t-\boldsymbol\mu_t\big).
  \label{eq:dev-split-app}
\end{align}

\paragraph{Why we refer to the first term as a contribution to covariance.} Written for a single
context $X$ it is $\sum_t (A_{st}-\hat A_{st})\overline{\v v}_t$, a
realization rather than a covariance.  Its meaning is fixed by the ensemble
the kernel is calibrated on: all contexts having query type $s$.

As shown above, Equation~\ref{eq:21} approximately holds for our data, and 
the empirical deviations from it do not affect our key results. 
Hence, taking the expectation of
a term of the first group over the contexts with query type $s$ in which
$t$ appears,
\begin{equation}
  \mathbb E_{st}\!\big[(A_{st}-\hat A_{st})\,\overline{\v v}_t\big]
  =\operatorname{Cov}_{st}\!\big(A_{st}-\hat A_{st},\,\overline{\v v}_t\big)
  +\underbrace{\mathbb E_{st}\big[A_{st}-\hat A_{st}\big]}_{\approx\,0}
   \,\mathbb E_{st}\big[\overline{\v v}_t\big]
  \;\approx\;\operatorname{Cov}_{st}\!\big(A_{st}-\hat A_{st},\,\overline{\v v}_t\big),
  \label{eq:cov-identity}
\end{equation}
and the corpus constant $\boldsymbol\mu_t$ may equally be subtracted from
$\overline{\v v}_t$ inside the covariance, for the same reason.  So the
first group is, in expectation over the query-type ensemble, the per-type
covariance between the head's \emph{residual} attention --- the part of
$A_{st}$ that presence and corpus propensity, i.e.\ the kernel, do not
predict --- and the value it thereby reads.  This is the quantity
Assumption~\ref{ass:proportionality} sets to zero: under proportionality
$A_{st}=\hat A_{st}$ identically, and the covariance vanishes term by
term.  More generally it vanishes iff, across contexts sharing the query
type, how much \emph{more or less than predicted} the head attends to
type $t$ is uncorrelated with \emph{what} it reads there: i.e.\ the head
transports content fixed by token type, not selected by context.  A single
context cannot exhibit a covariance, but a large first term signals that
$X$ sits in the tail of this ensemble covariance, which is why we describe
its magnitude as the context's contribution to covariance.

The second group (contextualization) instead varies the \emph{values} at fixed kernel weights: it is nonzero when
the in-context type-$t$ value $\overline{\v v}_t$ departs from its corpus average
$\boldsymbol\mu_t$, i.e.\ when the tokens the head reads for type $t$ carry context-specific
content. Had we instead used the paper's full mean field
$\hat{\v z}'=\sum_t\hat A_{st}\,v(\v c_t)$ of \eqref{eq:mean-field}, a third value-map (Jensen)
group $\sum_t\hat A_{st}\big(\boldsymbol\mu_t-v(\v c_t)\big)$ would appear; using the measured
$\boldsymbol\mu_t$ removes it, and it is small in any case (Appendix~\ref{app:clustering}). The
metrics $M_{\mathrm{cov}},M_{\mathrm{ctx}}$ of \eqref{eq:mcov-mctx} resolve the two groups of
\eqref{eq:dev-split-app} along the component of the error orthogonal to $\hat{\v z}$ ---
signed shares that sum to one --- and scale by the deviation $D^{\ell h}_{s,X}$, so per head
the two meters sum to the deviation exactly; the network-level meters average these per-head
quantities over heads and layers.  The weighting of the value drift is by $\hat A_{st}$,
as shown in \eqref{eq:dev-split-app}.

\section{Impact of the Value-Map Mean-Field Approximation}
\label{app:clustering}

This appendix asks what is lost by evaluating the value map at a type's
centroid rather than averaging it over the type's occurrences.  We find that little
is lost: occurrences of a type sit close enough to their centroid that the two agree.

The argument has two parts.
The error is exactly the Jensen gap of the attention block's input normalizer, so it
is governed by the within-type covariance; and that covariance is small enough, at
every layer of both models, that $\bar{\v{v}}_t \approx v(\v{c}_t)$.

The corpus mean field (Section~\ref{sec:meanfield}) replaces the per-occurrence value
$v(\v{x}_t)$ by the value of the type centroid $v(\v{c}_t)$, i.e.\ it assumes
$\bar{\v{v}}_t = \mathbb{E}[v(\v{x}_t)] \approx v(\v{c}_t)$. Because the per-head value map
$v(\v{x}) = \mathrm{ln}_1(\v{x})\,W_V + \v{b}_V$ has the input normalizer $\mathrm{ln}_1$ (LayerNorm in
GPT-2, RMSNorm in Gemma-2) as its only nonlinearity (and $W_V$ is linear, so it commutes with the
expectation), the error is exactly the Jensen gap of $\mathrm{ln}_1$ over the type-$t$ occurrence cloud,
\[
  \bar{\v{v}}_t - v(\v{c}_t) = \big(\mathbb{E}[\mathrm{ln}_1(\v{x}_t)] - \mathrm{ln}_1(\v{c}_t)\big)\,W_V .
\]
A second-order (delta-method) expansion~\citep{oehlert1992delta,gao2019jensen} gives
$\mathbb{E}[\mathrm{ln}_1(\v{x}_t)] - \mathrm{ln}_1(\v{c}_t) \approx
\tfrac12\,\mathrm{tr}\!\big(\nabla^2\mathrm{ln}_1(\v{c}_t)\,\mathrm{Cov}(\v{x}_t)\big)$, where the
LayerNorm Jacobian and Hessian follow from $\mathrm{ln}_1(\v{x})=\sqrt{d}\,P\v{x}/\|P\v{x}\|$ with the
mean-centering projection $P=I-\tfrac1d\v{1}\v{1}^\top$~\citep{ba2016layernorm,brody2023layernorm}
(RMSNorm omits $P$, $\mathrm{ln}_1(\v{x})=\sqrt{d}\,\v{x}/\|\v{x}\|$, but the argument is unchanged).
The gap is therefore governed by the \emph{within-type covariance} $\mathrm{Cov}(\v{x}_t)$: if
occurrences of a type cluster tightly around their centroid, the value-map mean field is exact to
second order. We verify this clustering directly.

Within-type clustering is the Euclidean counterpart of the contextual self-similarity of
\citet{ethayarajh2019contextual}, who studied mean pairwise cosine across a word's occurrences, which they found decreases
with depth. 

\paragraph{Metrics.} Write $\{\v{x}_{t,i}\}_{i=1}^{n_t}$ for the residual-stream
representations of the $n_t$ occurrences of type $t$ at a given depth, with type centroid
$\v{c}_t=\tfrac1{n_t}\sum_i\v{x}_{t,i}$ and global centroid
$\bar{\v{c}}=\tfrac1{\sum_t n_t}\sum_{t,i}\v{x}_{t,i}$. We report three quantities per layer.
\emph{(i) Relative RMS deviation:}
$r_t = \sqrt{\tfrac1{n_t}\sum_i\|\v{x}_{t,i}-\v{c}_t\|^2}\,/\,\|\v{c}_t\|$, the root-mean-square spread
of a type's occurrences about its centroid, relative to the centroid magnitude.
\emph{(ii) Variance explained by type, $\eta^2$:} a one-way variance decomposition (the intraclass
correlation). With total variance
$V_{\mathrm{tot}}=\tfrac1{\sum n_t}\sum_{t,i}\|\v{x}_{t,i}-\bar{\v{c}}\|^2$ and pooled within-type
variance $V_{\mathrm{w}}=\tfrac1{\sum n_t}\sum_{t,i}\|\v{x}_{t,i}-\v{c}_t\|^2$,
\[
  \eta^2 = 1 - V_{\mathrm{w}}/V_{\mathrm{tot}},
\]
the fraction of variance attributable to type identity ($\eta^2\!\to\!1$ means occurrences collapse
onto their centroids; $\eta^2\!\to\!0$ means type identity carries no information).
\emph{(iii) Value-map Jensen gap:} the relative magnitude
$\|\mathbb{E}[\mathrm{ln}_1\v{x}_t]-\mathrm{ln}_1\v{c}_t\|/\|\mathrm{ln}_1\v{c}_t\|$ and the direction
$1-\cos(\mathbb{E}[\mathrm{ln}_1\v{x}_t],\,\mathrm{ln}_1\v{c}_t)$ --- the actual size and orientation
of the gap that the second-order argument above bounds. All quantities are accumulated in a single
pass over 2M OpenWebText tokens, per type and per residual depth $E[\ell]$ (the input to each block's
attention).

\paragraph{Rogue-dimension control.} As discussed in more detail in Appendix~\ref{app:rogue}, a handful of coordinates --- ``rogue dimensions''
--- carry a disproportionate share of the residual variance. 
\citet{timkey2021rogue} showed that for GPT-2's final layer a single
dimension accounts for ${\sim}76\%$ of the expected cosine between two random
tokens. A variance-weighted statistic, including the $\eta^2$ above (which
sums variance across coordinates), is therefore dominated by whether those few dimensions separate
types, and is nearly blind to the structure in the remaining ${\sim}760$. We accordingly also report
a per-dimension-standardized version: writing $\sigma^2_d$ and $w_d$ for the total and pooled-within
variance of coordinate $d$, we average the \emph{per-coordinate} ratio,
\[
  \eta^2_z = \tfrac1d\textstyle\sum_{d}\big(1 - w_d/\sigma^2_d\big),
\]
which weights every dimension equally --- equivalent to z-scoring each coordinate to unit variance
before the decomposition --- so that no single rogue dimension dominates. (The raw
$\eta^2=1-\sum_d w_d/\sum_d\sigma^2_d$ is its variance-weighted counterpart.) This is the
standardization control recommended by \citet{timkey2021rogue}.

\begin{figure}[t]
\centering
\includegraphics[width=\linewidth]{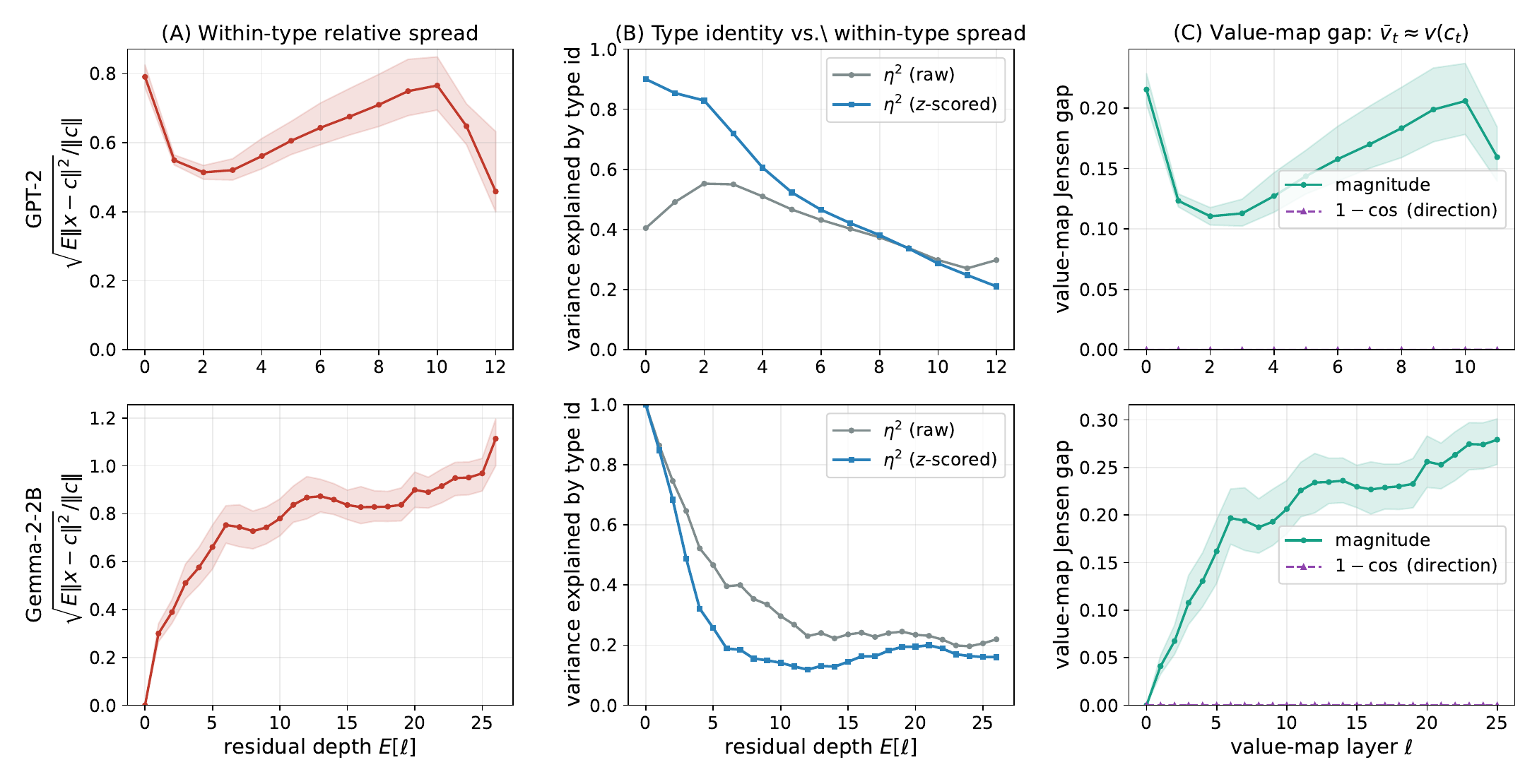}
\caption{Same-type token representations and their type centroids
(2M OpenWebText tokens, top-1000 types). \textbf{Top row:} GPT-2; \textbf{bottom row:} Gemma-2-2B.
\textbf{(A)} relative RMS deviation of occurrences around their type centroid,
$\sqrt{\mathbb{E}\|\v{x}-\v{c}_t\|^2}/\|\v{c}_t\|$ (median; IQR band over types).
\textbf{(B)} fraction of variance explained by type identity, $\eta^2$: raw (gray) and per-dimension
z-scored (blue).
\textbf{(C)} the value-map Jensen gap
$\|\mathbb{E}[\mathrm{ln}_1\v{x}]-\mathrm{ln}_1\v{c}_t\|/\|\mathrm{ln}_1\v{c}_t\|$ (green) and its
direction $1-\cos$ (purple).}
\label{fig:clustering}
\end{figure}

\paragraph{Results.} Figure~\ref{fig:clustering} reports the results for both models. In GPT-2 the
rogue-dimension-robust measure shows type identity explaining $\eta^2_z=0.90$ of the variance at the
model input: occurrences are very tightly clustered around their type centroid relative to the
between-type spread. The raw $\eta^2$ \emph{understates} this at $0.40$ --- a few rogue dimensions
obscure the within-type structure, as documented by \citet{timkey2021rogue}. Both measures decline with depth (to $\eta^2_z\approx0.25$ by
layer~11), the geometric form of the increasing context-specificity of
\citet{ethayarajh2019contextual}.

Gemma-2 behaves similarly: $\eta^2_z=0.85$ at the first block input falling to
${\approx} 0.15$ in the deep layers.  There are two aspects specific to Gemma-2. First, its input residual
$E[0]$ is \emph{exactly} degenerate ($\eta^2=1$, zero within-type spread): RoPE injects position
inside the attention computation rather than adding it to the residual, so a type's input
representation is identical across occurrences, whereas GPT-2's additive positional embedding already
spreads a type's occurrences at the input. Second, for Gemma the z-scored $\eta^2_z$ sits slightly
\emph{below} the raw $\eta^2$ rather than above it (e.g.\ $0.32$ vs $0.52$ at $E[4]$): its
high-variance coordinates are themselves type-discriminative, so equalizing dimensions removes
information rather than noise --- the opposite of GPT-2 --- though the clustering stays strong under
both measures.

The impact of this tight clustering for the mean field is the same in both models: the value-map Jensen gap is
negligible at every layer (right-hand side of the figure). In terms of cosine distance, in GPT-2 (using LayerNorm), 
$1-\cos(\mathbb{E}[\mathrm{ln}_1\v{x}],\mathrm{ln}_1\v{c}_t)<3\times10^{-5}$, and in Gemma-2 (using RMSNorm),
$<3\times10^{-4}$.  (These are invisible on the plots.)  Thus, in both models, $\bar{\v{v}}_t$ and $v(\v{c}_t)$ are essentially parallel, and have a residual magnitude gap of order $10$--$28\%$. The clustering is thus tight enough that
replacing $\bar{\v{v}}_t$ by $v(\v{c}_t)$ is an excellent approximation, particularly in direction
--- the empirical condition behind the corpus mean field.

\section{Mean-Field Kernel Details}
\label{app:modeling-details}

The mean-field rollout tracks the $N$ most frequently occurring types,
$\mathcal{T}_N\subset\mathcal{T}$.
So that each query's
attention remains normalized, the kernel for each head carries two key columns
beyond the $N$ tracked types: a \textbf{BOS/position-0} column and an
\textbf{other} column that aggregates every type not in $\mathcal{T}_N$. Thus $P^{\ell h}$ is
an $N\times(N{+}2)$ row-stochastic matrix: for a query of type $s$,
\[
\sum_{t\in\mathcal{T}_N} P^{\ell h}_{st} \;+\; P^{\ell h}_{s\bullet} \;+\;
P^{\ell h}_{s\circ} \;=\; 1,
\]
where $P^{\ell h}_{s\bullet}$ and $P^{\ell h}_{s\circ}$ are the
corpus-averaged attention masses a type-$s$ query places on position~$0$
and on positions holding non-tracked types, respectively, defined exactly as
the tracked masses $P^{\ell h}_{st}=\mathbb{E}_{\mathrm{occ}}[A_{st}]$. The
per-head contribution of \eqref{eq:mean-field}
(with $v(\cdot)$ the head's value map) is then
\[
\v{z}^{\ell h}_s = \sum_{t\in\mathcal{T}_N} P^{\ell h}_{st}\,v(\hat{\v{c}}^\ell_t)
\;+\; P^{\ell h}_{s\bullet}\,v(\v{b}^\ell)
\;+\; P^{\ell h}_{s\circ}\,v(\bar{\v{c}}^\ell).
\]

\paragraph{BOS/position-0 column.} Position~$0$ of every input holds the
BOS token, and under the causal mask it attends only to itself, so its
representation has no occurrence-to-occurrence variation.
$\v{b}^\ell$ is that layer-$\ell$ representation: a deterministic function of the
model weights, obtained from a single forward
pass on the one-token input $[\mathrm{BOS}]$ (equivalently, the corpus mean of
position-0 representations). It is therefore not a measured contextual
quantity.
Position~$0$ is also where the models place a large share of their attention,
the attention sink of Appendix~\ref{app:coverage}, which is why this column is kept
separate rather than folded into ``other''.
Every input is prepended with the model's BOS token --- or, for models
with no dedicated BOS such as GPT-2, its surrogate \texttt{<|endoftext|>} ---
following the TransformerLens convention. In models with a true BOS the token
can be frequent enough to enter $\mathcal{T}_N$ itself (Gemma-2's
\texttt{<bos>} does), and it must then be excluded as a \emph{query} type
from the $P$ and $W$ matrices: it occurs only at position~$0$, where the
causal prefix contains nothing else to attend to, so its kernel row is empty
and its renormalized attention undefined.

\paragraph{Other column.} The non-tracked types are collapsed into a single
column whose value we set to $\bar{\v{c}}^\ell=\tfrac1N\sum_{t\in\mathcal{T}_N}\hat{\v{c}}^\ell_t$,
the running mean of the predicted \emph{tracked} centroids at layer~$\ell$. This is
internal to the rollout and requires no measurement. 
Consequently the rollout consults no measured residual at any layer: the
tracked centroids $\hat{\v{c}}^\ell_t$ evolve from the input embeddings, the BOS
value $\v{b}^\ell$ is a weight-derived constant, and the ``other'' value
$\bar{\v{c}}^\ell$ is the running centroid mean.

\paragraph{Held-out evaluation.} All comparisons in Section~\ref{sec:validation} are out-of-sample. We split OpenWebText into two disjoint halves by document parity (even-indexed documents versus odd), $5$M tokens each, and estimate every corpus quantity the model uses --- the mean-field attention kernel $P^{\ell h}$, the co-occurrence statistics, and the per-type value and centroid statistics --- on the \emph{train} half, while measuring the true per-layer centroids $\v{c}^\ell_t$ that the rollout is scored against on the disjoint \emph{test} half. The rollout therefore never sees the evaluation centroids during estimation.

\paragraph{Estimation of $W$.}
Because $W_{st}=P_{st}/\bar n_{st}$ is a ratio of estimated quantities, its entries are noise-dominated — and can be arbitrarily large — wherever the pair $(s,t)$ is rarely observed. We therefore restrict $W$ to its well-estimated support: entries whose co-occurrence count $C_{st}$ (the number of $s$-occurrences with at least one $t$ in the causal prefix) falls below $20$ are set to zero, and $\Ahat_{stX}$ renormalises over the remaining support. The mask discards $53\%$ of $(s,t)$ pairs but only $3\%$ of attention mass (Gemma-2: $41\%$ and $5\%$), and all context-conditional results below are computed under it.
\section{Coverage of the Tracked Type Set}
\label{app:coverage}

Every construction in the body of this paper tracks the $N=1000$ most frequent token
types, with everything else collapsed into a position-$0$ bucket and an ``other''
bucket (Appendix~\ref{app:modeling-details}).
How much of the corpus that set covers depends on
what is counted: about two thirds of corpus positions carry a tracked type, while the
share of attention mass landing on tracked types is far lower, $18$--$42\%$, almost
all of the difference attributable to the attention sink at position~$0$.
The choice of $N$ itself matters less than either figure suggests:
Appendix~\ref{app:waypoint-vocab} repeats the training-waypoint substitution at
$N=2000$ and $N=4000$ and finds the phase structure unchanged.

\paragraph{Position coverage.}
Since each position defines one context (Section~\ref{sec:notation}),
corpus positions whose type is tracked equate to 
contexts for which the mean field has a tracked query type.  The fraction of positions covered is the relevant
measure wherever the substitution is applied only at tracked query types.  The
remaining positions keep the model's own computation.  Measured over $1$M OpenWebText
tokens with each model's own tokenizer and its own top-1000 set
(Table~\ref{tab:position-coverage}):

\begin{table}[h]
\centering\small
\caption{Position coverage: the fraction of corpus positions ($1$M
OpenWebText tokens) whose type is among the model's top-$1000$ set.}
\label{tab:position-coverage}
\begin{tabular}{lrr}
\toprule
model & vocabulary & position coverage \\
\midrule
GPT-2 small        & $50{,}257$  & $0.663$ \\
Pythia             & $50{,}277$  & $0.655$ \\
Gemma-2-2B         & $256{,}000$ & $0.684$ \\
Llama-3.2-3B / 3.1-8B & $128{,}256$ & $0.643$ \\
Qwen-3-14B         & $151{,}669$ & $0.660$ \\
\bottomrule
\end{tabular}
\end{table}

Coverage of positions sits at $64$--$68\%$ across a fivefold spread in vocabulary size.
$1000$ types out of Gemma-2's $256{,}000$ cover as much text as $1000$ out of GPT-2's
$50{,}257$.  Token frequency is Zipfian, so the head of the distribution is the same
text however finely the tail is divided.  The two Llamas share a tokenizer, and so
share a set and a coverage.

\begin{figure}[h]
\centering
\includegraphics[width=0.66\linewidth]{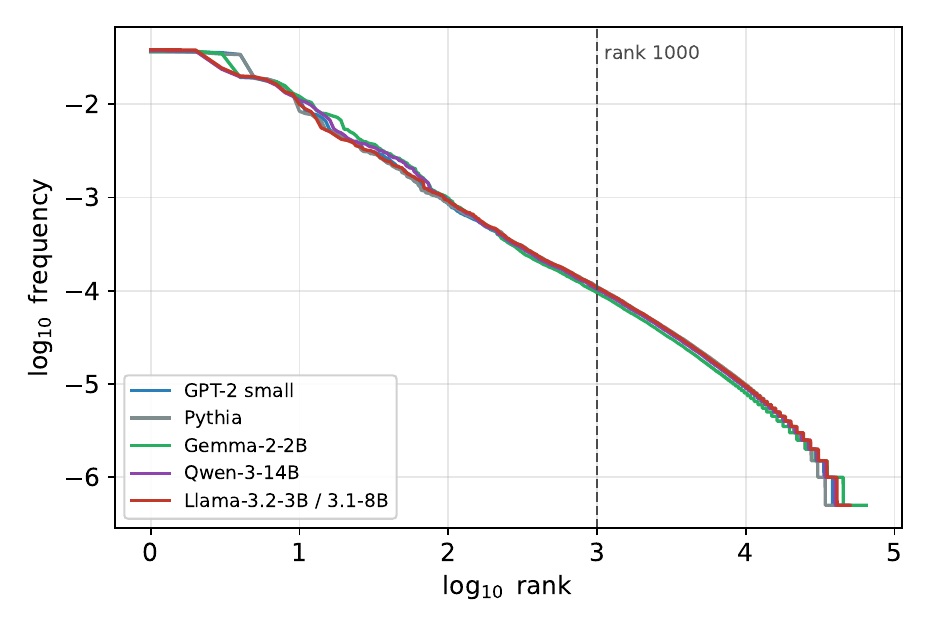}
\caption{Rank--frequency curves for the five tokenizers on $2$M tokens
of OpenWebText, log--log, one curve per tokenizer; the two Llamas share a
tokenizer and so a curve.  Dashed line: the rank-$1000$ cutoff used throughout.
The curves are nearly indistinguishable up to the cutoff and for a decade beyond
it, diverging only in the far tail where vocabulary size decides how finely rare
text is split.}
\label{fig:zipf}
\end{figure}

Figure~\ref{fig:zipf} shows why.  The rank--frequency curves of the five
tokenizers, measured on the same $2$M tokens of OpenWebText, lie almost on top of one
another over the tracked range, and the rank-$1000$ cutoff falls at nearly the same
frequency in every model ($0.96$--$1.10\times10^{-4}$).  They separate only beyond
rank ${\sim}10^{4}$, where the larger vocabularies keep resolving types the smaller
ones have merged, which is the region the tracked set excludes.  Taking the top $1000$
types therefore selects the same text in each model, not just the same number of
types.

\paragraph{Attention-mass coverage.}
The second quantity is how much of the attention a head places lands on
tracked types, as opposed to the position-$0$ bucket or the ``other'' bucket. 
This is the row sum $\sum_{t\in\mathcal{T}_N} P^{\ell h}_{st}$, the share of query
type $s$'s attention that reaches tracked key types, averaged over layers, heads and
tracked query types (Table~\ref{tab:attn-coverage}).

\begin{table}[h]
\centering\small
\caption{Attention-mass coverage: average per-query attention landing on
tracked types, the position-$0$ sink, and the ``other'' bucket.  The final
column is the tracked share of non-sink attention, computed from the row
entries; it is the quantity comparable to the position coverage of
Table~\ref{tab:position-coverage}.}
\label{tab:attn-coverage}
\begin{tabular}{lrrrr}
\toprule
model & tracked types & position 0 & other & tracked / (tracked $+$ other) \\
\midrule
GPT-2 small   & $0.419$ & $0.399$ & $0.182$ & $0.697$ \\
Gemma-2-2B    & $0.417$ & $0.405$ & $0.178$ & $0.701$ \\
Qwen-3-14B    & $0.370$ & $0.488$ & $0.142$ & $0.723$ \\
Llama-3.2-3B  & $0.231$ & $0.654$ & $0.115$ & $0.668$ \\
Llama-3.1-8B$^\dagger$ & $0.181$ & $0.654$ & $0.165$ & $0.523$ \\
\bottomrule
\end{tabular}
\end{table}

$^\dagger$The Llama-3.1-8B row is measured on the reduced-vocabulary
extraction of Appendix~\ref{app:llama}, which excludes $67$ structural types
that are both frequent and heavily attended, so its tracked share is
understated.  Measured on that same reduced basis the Llama-3.2-3B figure falls
from $0.231$ to $0.182$, against the 8B's $0.181$, so on a like-for-like
footing the two models agree, and the 8B's full-vocabulary value should be read
as ${\approx}0.23$.  The final column inherits the same understatement, so the
8B's $0.523$ is not comparable to the full-vocabulary rows above it.

The largest single destination in every model is the sink: position~0 takes
$40\%$ of all attention in GPT-2 and Gemma-2.  It takes $49\%$ in Qwen-3 and $65\%$ in
both Llamas, consistent with the attention-sink
literature~\citep{xiao2024streaming,gu2025sink}.
The sink also accounts for nearly all of the gap between the two tables.
Raw attention-mass coverage sits far below position coverage, $18$--$42\%$ against
$64$--$68\%$, but the raw number charges the tracked set with sink mass that no choice
of tracked types could capture.
The comparable quantity is the tracked share of \emph{non-sink} attention, the final
column of Table~\ref{tab:attn-coverage}, labeled tracked/(tracked+other), which is $67$--$72\%$ for every model
measured on its full vocabulary.
So when heads attend to token content at all, they place attention on tracked types at
slightly above the tracked set's share of positions, and the shortfall in the raw
number belongs to the sink.

\paragraph{What the untracked mass costs.}
Attention that leaves the tracked set still enters the
prediction, through the two bucket values $v(\v b^\ell)$ and $v(\bar{\v c}^\ell)$ of
Appendix~\ref{app:modeling-details}, and thus the kernel remains row-stochastic.
However, some detail is lost. Attention to untracked types is answered by one shared
vector rather than by a value per type, so a model with a lower tracked share has more
of its heads' behavior compressed into that single average.  
\section{Mean-Field Predictions Across Corpora}
\label{app:corpus}

As described in the body, mean field analysis concerns the average behavior of a model over a given corpus.  The results in the body of the paper use OpenWebText as the reference corpus.  Here we describe how mean-field prediction of centroids varies across four different corpora.   We demonstrate that, across the corpora we study, differences between predictions are very small, and that centroid predictions generalize out of distribution (i.e., cross-corpora).

\subsection{Corpora Studied}
\label{sec:corpora}

\paragraph{Models.} We evaluate corpus differences using the same models as studied in the paper body: GPT-2 small and Gemma-2 2B.   Statistics are
re-extracted from scratch for each model, and each
model's shared vocabulary is re-derived as \emph{its} 1000 most frequent
OpenWebText types.

We choose corpora to reflect a diversity of ordinary English prose:
\begin{description}
\item[OpenWebText (OWT)~\citep{gokaslan2019openwebtext}.] The
  reference corpus, used in the paper body.  This corpus consists of web pages linked from Reddit, collected to reproduce
  GPT-2's training data~\citep{radford2019gpt2}. 
\item[FineWeb~\citep{penedo2024fineweb}.] This is a recent, heavily filtered
  crawl of the general web. We use the 10-billion-token sample.  Like OWT, 
  this corpus is general web text, but collected separately
  with different filtering.
\item[Wikipedia (WikiText-103)~\citep{merity2017pointer}.] This consists of `featured'
  and `good' Wikipedia articles, in raw form. This corpus is stylistically different from web text in that it is more formal and more encyclopedic in subject matter.
\item[Pre-1919 books~(PG-19) \citep{rae2020pg19}.] This set consists entirely of books
  published before 1919 from Project Gutenberg. As such, it has a literary style, uses older vocabulary and
  constructions, and has no web artifacts. 
\end{description}

All corpora are prepared identically:
text is tokenized with the model's tokenizer and packed into 512-token
windows (windows may span document boundaries), and statistics are
accumulated over 2 million tokens per (corpus, half) sample using
TransformerLens~\citep{nanda2022transformerlens}. Throughout, the
vocabulary is fixed to the 1000 most frequent types of the OWT
reference sample, so that all results are over the same set
of types.  

\subsection{Mean Field Predictions Are Stable Across Corpora and Generalize Out of Distribution}
\label{sec:matrix}

For each corpus $X$ we estimate the 
mean-field kernel and roll out predicted centroids on $X$'s
statistics half. For each corpus $Y$ we measure true centroids on $Y$'s
evaluation half. For each pair $(X, Y)$ we initialize
the rollout from $X$'s layer-0 centroids, roll it out with $X$'s kernel
through the model's weights, and score the predicted centroids at every layer
against $Y$'s true centroids. With four corpora this gives a $4 \times 4$
matrix of runs, which we ran in full for both models.

\begin{figure}[t]
\centering
\includegraphics[width=\linewidth]{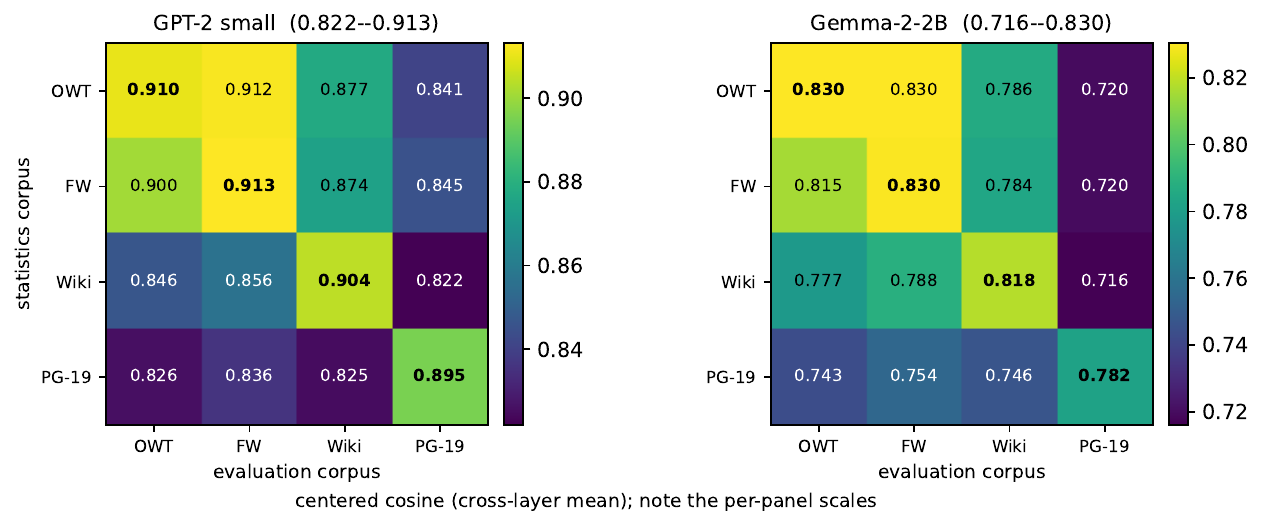}
\caption{Average cosine similarities of centroid predictions. Rows: corpus supplying the mean-field
statistics. Columns: corpus supplying the evaluation centroids. Cell values
are the cross-layer mean centered cosine; diagonal (matched) cells in bold.
}
\label{fig:heat}
\end{figure}

\begin{figure}[t]
\centering
\includegraphics[width=\linewidth]{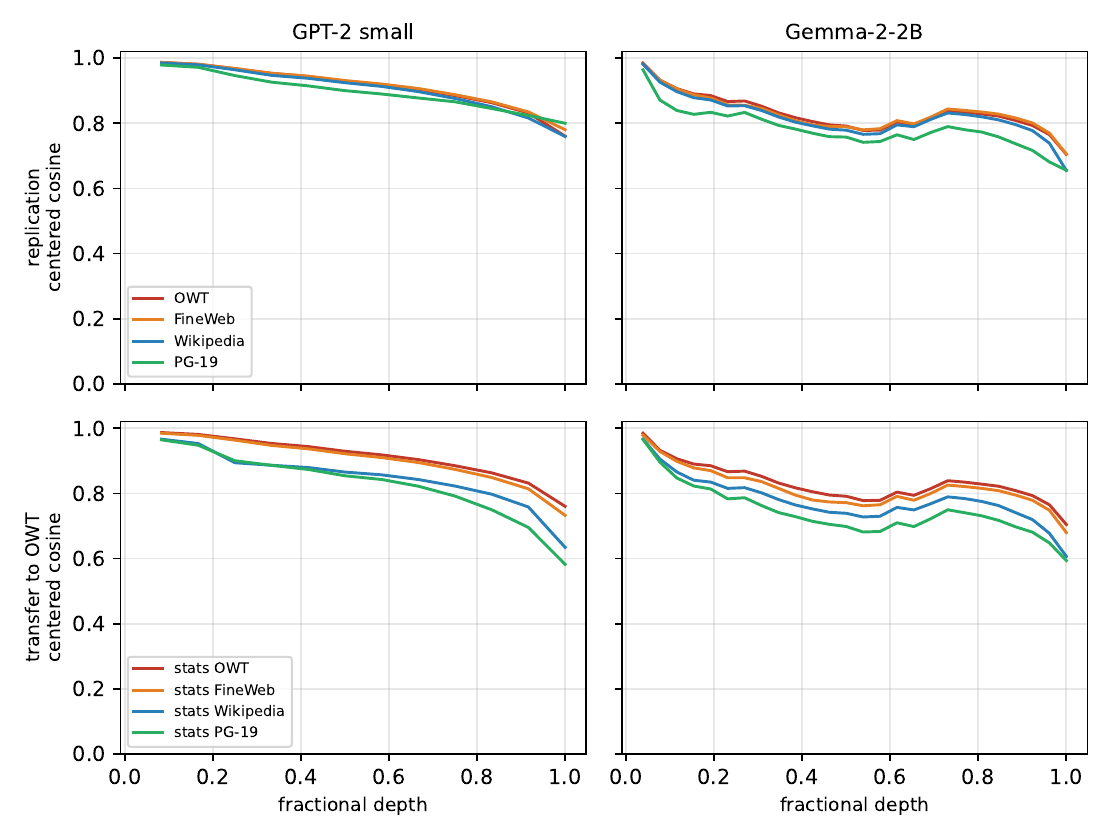}
\caption{Per-layer rollout quality by fractional depth, one column
per model. Top row: replication (statistics and evaluation on the same
corpus), one curve per corpus. Bottom row: generalization (statistics from each
corpus, evaluated against OpenWebText centroids).}
\label{fig:curves}
\end{figure}

Figure~\ref{fig:curves} shows centroid prediction accuracy using the same open-loop prediction as Section~\ref{sec:validation}.  We summarize those figures in Figure~\ref{fig:heat}, where the diagonal cells
($X = Y$) show the average cosine similarity of mean-field prediction of centroids for that corpus (average over curve values in Figure~\ref{fig:curves}). 
The off-diagonal cells assess generalization: whether statistics measured on one kind of text predict behavior on another.

Table \ref{tab:matrix} shows bootstrapped 95\% confidence intervals for Figure~\ref{fig:heat}, and additionally the context-free baselines.  The context-free baseline runs
each type through the model alone, as the two-token input
$[\textrm{BOS}, t]$, and scores the resulting per-layer residual against the
true centroids exactly as a prediction is scored. Using no corpus statistics
beyond the layer-0 initialization, it is the floor that any corpus-derived
kernel must clear. The context-free forward does not depend on a corpus, so a
single set of residuals serves every column and only the evaluation centroids
change.

\begin{table}[t]
\centering\small
\caption{Values from Figure~\ref{fig:heat} showing confidence intervals as well as context-free baseline}
\label{tab:matrix}
\begin{tabular}{llcccc}
\toprule
model & statistics & eval OWT & eval FineWeb & eval Wikipedia & eval PG-19 \\
\midrule
GPT-2 & OWT & \textbf{0.910} [.907,.913] & 0.912 [.910,.915] & 0.877 [.873,.881] & 0.841 [.835,.846] \\
 & FineWeb & 0.900 [.897,.904] & \textbf{0.913} [.911,.916] & 0.874 [.870,.878] & 0.845 [.839,.850] \\
 & Wikipedia & 0.846 [.838,.855] & 0.856 [.848,.865] & \textbf{0.904} [.901,.907] & 0.822 [.812,.831] \\
 & PG-19 & 0.826 [.818,.833] & 0.836 [.828,.842] & 0.825 [.817,.832] & \textbf{0.895} [.889,.900] \\
\cmidrule(lr){2-6}
 & \emph{context-free} & \emph{0.799} & \emph{0.807} & \emph{0.781} & \emph{0.745} \\
\midrule
Gemma-2 & OWT & \textbf{0.830} [.826,.834] & 0.830 [.826,.834] & 0.786 [.781,.791] & 0.720 [.713,.726] \\
 & FineWeb & 0.815 [.810,.820] & \textbf{0.830} [.826,.833] & 0.784 [.779,.789] & 0.720 [.712,.727] \\
 & Wikipedia & 0.777 [.767,.786] & 0.788 [.779,.797] & \textbf{0.818} [.814,.822] & 0.716 [.705,.728] \\
 & PG-19 & 0.743 [.734,.752] & 0.754 [.745,.762] & 0.746 [.738,.754] & \textbf{0.782} [.775,.788] \\
\cmidrule(lr){2-6}
 & \emph{context-free} & \emph{0.686} & \emph{0.699} & \emph{0.681} & \emph{0.607} \\
\bottomrule
\end{tabular}
\end{table}

We make three observations.
\begin{enumerate}
    \item Mean-field centroid prediction replicates well on each corpus. Diagonal cells are all 
    at or near the model's published OpenWebText level, including for pre-1919 books. 
    \textbf{Thus the conclusions of Section~\ref{sec:validation} are not specific to the OWT corpus used.} 
    \item \textbf{OOD generalization is strong} (off-diagonal cells in Figure~\ref{fig:heat}). Training the mean-field on one corpus and predicting centroid evolution on a different corpus yields accuracy generally close to the same-corpus evaluation.  The worst mismatched cell clears the context-free baseline in every column and in
    both models. Statistics from century-old books still carry
    usable information about how the model treats modern web text.
    \item OOD generalization degrades according to expected dissimilarity of corpora, with generalization between OWT and FineWeb best, and pre-1919 novels least.  All within-column differences are statistically distinguishable %
\end{enumerate}

\section{Full Mean Field Results}
\label{app:full-mean-field-results}
The four metrics compare, at each layer $\ell$, the mean-field
predictions $\hat{\v{c}}^{\,\ell}_t$ of Algorithm~\ref{alg:rollout}
with the true centroids $\v{c}^{\,\ell}_t$ measured on the held-out
corpus half, over the evaluation set $\mathcal{T}$ of the $1000$ most
frequent types (experimental settings in \S\ref{sec:validation}).
Because the representation cloud is strongly anisotropic --- it sits far
from the origin relative to its own size
(Appendix~\ref{app:anisotropy}) --- every metric is computed after
mean-centering each cloud by its own mean: writing
$\bar{\v{c}} = \tfrac{1}{|\mathcal{T}|}\sum_{t\in\mathcal{T}}\v{c}^{\,\ell}_t$
and $\bar{\hat{\v{c}}}$ analogously for the predictions, we score the
centered quantities
$\tilde{\v{c}}_t = \v{c}^{\,\ell}_t - \bar{\v{c}}$ and
$\tilde{\hat{\v{c}}}_t = \hat{\v{c}}^{\,\ell}_t - \bar{\hat{\v{c}}}$.
Without centering, the shared offset would dominate every similarity.
The four metrics are then:
\begin{itemize}
\item \emph{Centered cosine similarity} --- the mean over types of the
per-type cosine,
$\tfrac{1}{|\mathcal{T}|}\sum_{t}
\cos\!\big(\tilde{\hat{\v{c}}}_t,\ \tilde{\v{c}}_t\big)$:
per-type directional agreement, aggregated by a simple mean.
\item \emph{Representation similarity analysis (RSA)} --- we compute the
pairwise cosine similarities among the centered true centroids
$\{\tilde{\v{c}}_t\}$, the same among the centered predictions
$\{\tilde{\hat{\v{c}}}_t\}$, and report the Pearson correlation
between the two lists of off-diagonal entries.  RSA scores whether the
predicted cloud has the right \emph{shape} --- which types lie near which
--- independently of any global rotation of the cloud.
\item \emph{Centered relative error} --- the ratio of means
$\sum_{t}\big\|\tilde{\hat{\v{c}}}_t-\tilde{\v{c}}_t\big\| \,\big/\,
\sum_{t}\big\|\tilde{\v{c}}_t\big\|$.
A ratio of means rather than a mean of per-type ratios, so types lying
near the cloud center do not arbitrarily inflate the metric; this is the
one metric jointly sensitive to direction and magnitude.
\item \emph{Centered norm ratio} ---
$\sum_{t}\big\|\tilde{\hat{\v{c}}}_t\big\| \,\big/\,
\sum_{t}\big\|\tilde{\v{c}}_t\big\|$, the predicted cloud's mean
radius relative to the true cloud's, isolating magnitude: $1$ is ideal,
values above $1$ indicate overshoot.
\end{itemize}

\begin{figure}[p]
    \centering
    \includegraphics[width=0.32\linewidth]{figures/gemma_cos_source.pdf}
    \includegraphics[width=0.32\linewidth]{figures/gpt2_cos_source.pdf}
    \includegraphics[width=0.32\linewidth]{figures/qwen14b_cos_source.pdf}\\
    \includegraphics[width=0.32\linewidth]{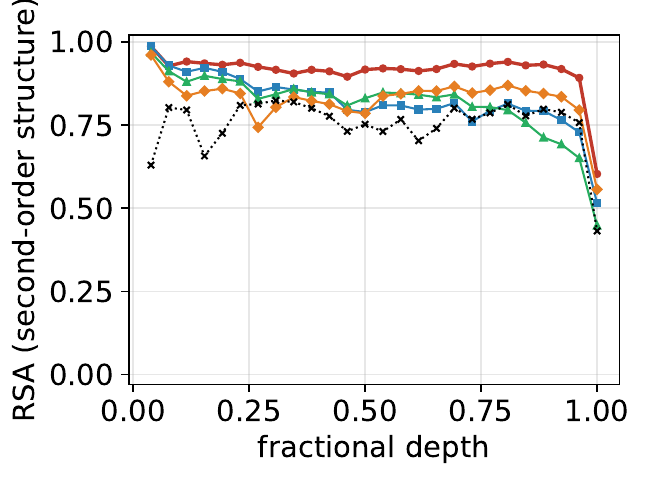}
    \includegraphics[width=0.32\linewidth]{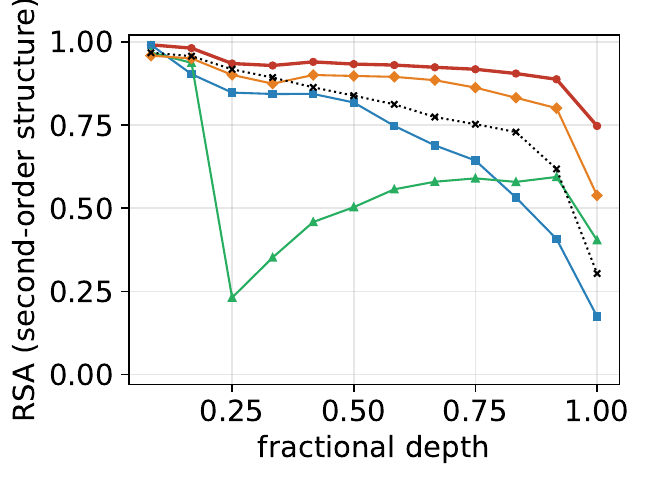}
    \includegraphics[width=0.32\linewidth]{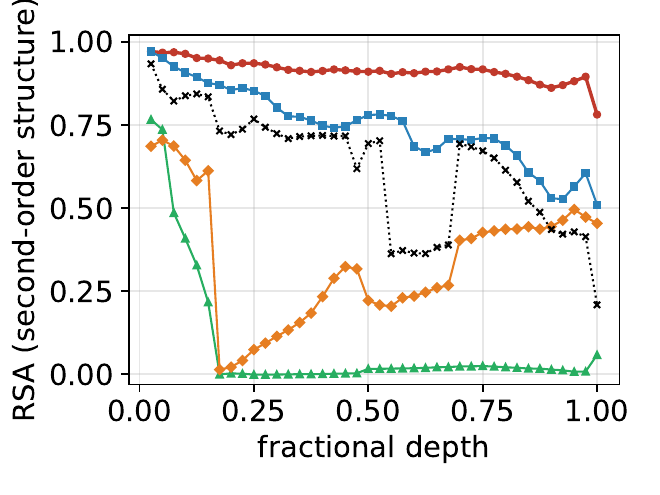}\\
    \includegraphics[width=0.32\linewidth]{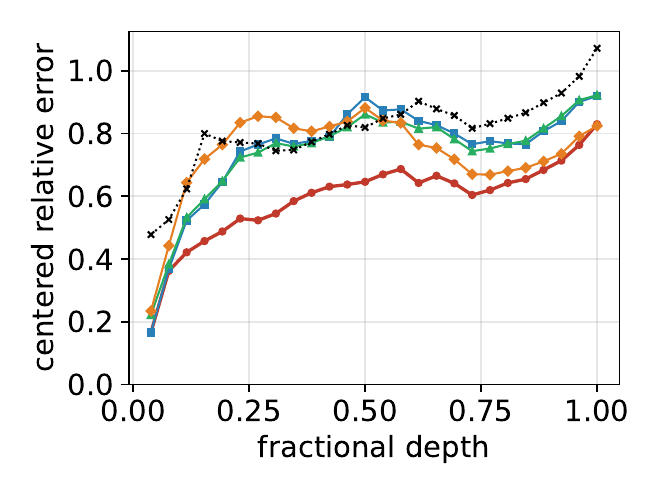}
    \includegraphics[width=0.32\linewidth]{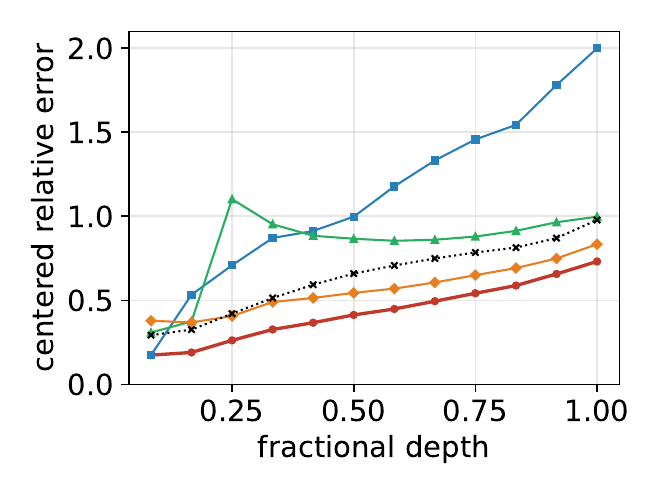}
    \includegraphics[width=0.32\linewidth]{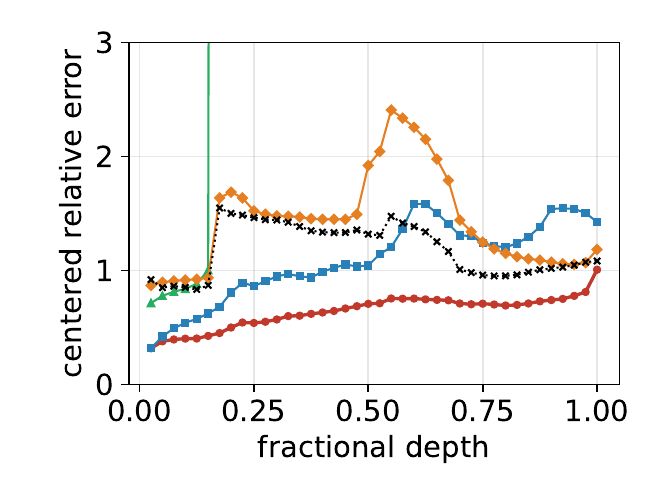}\\
    \includegraphics[width=0.32\linewidth]{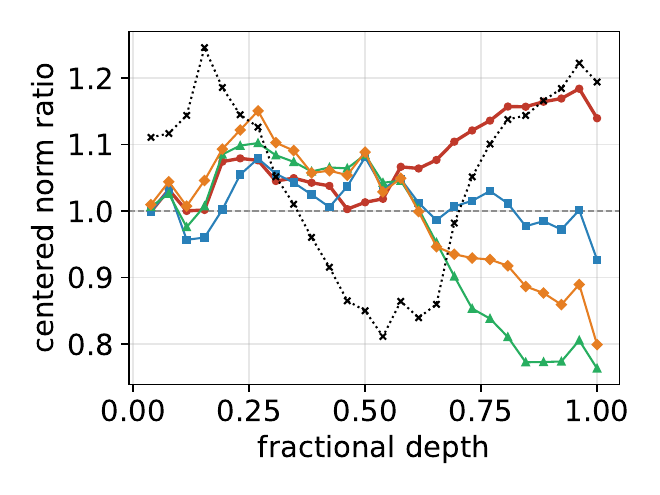}
    \includegraphics[width=0.32\linewidth]{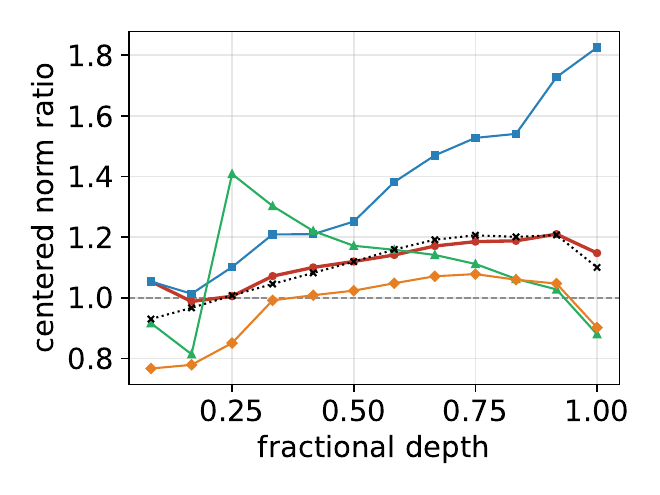}
    \includegraphics[width=0.32\linewidth]{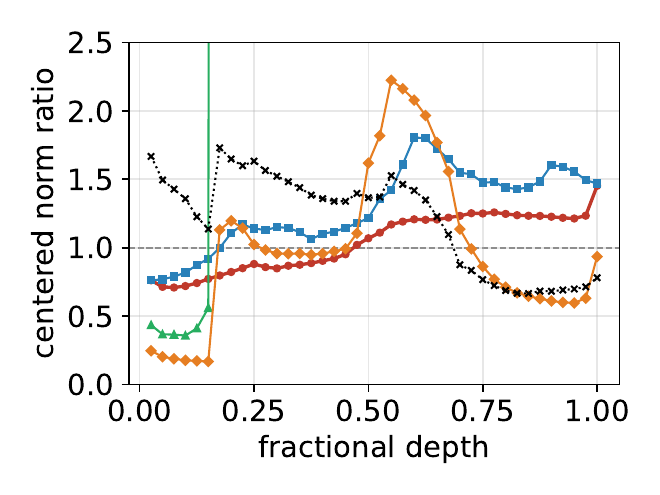}\\[4pt]
    \includegraphics[width=0.85\linewidth]{figures/legend_source.pdf}
    \caption{Mean-field predictions are accurate and superior to alternative kernels (kernel-source substitutions). Left: Gemma-2; middle: GPT-2; right: Qwen-3-14B. Rows (top to bottom): centered cosine similarity, representation similarity analysis (RSA), centered relative error, and centered norm ratio. For Qwen the \texttt{uniform} and \texttt{Pavg} substitutions diverge in magnitude; their relative-error and norm-ratio curves leave the plotted range.}
    \label{fig:mf-vs-alternative-kernels-full}
\end{figure}

\begin{figure}[p]
    \centering
    \includegraphics[width=0.42\linewidth]{figures/gemma_cos_recipe.pdf}
    \includegraphics[width=0.42\linewidth]{figures/gpt2_cos_recipe.pdf}\\
    \includegraphics[width=0.42\linewidth]{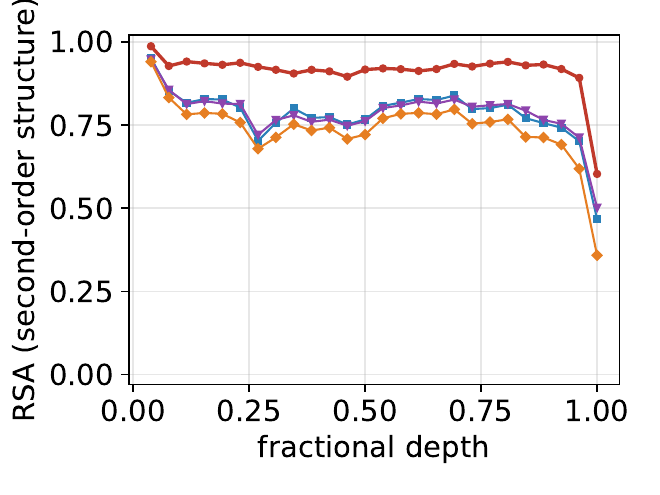}
    \includegraphics[width=0.42\linewidth]{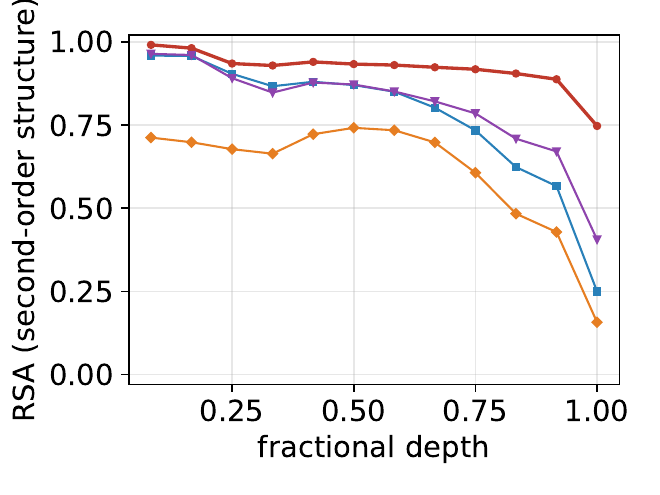}\\
    \includegraphics[width=0.42\linewidth]{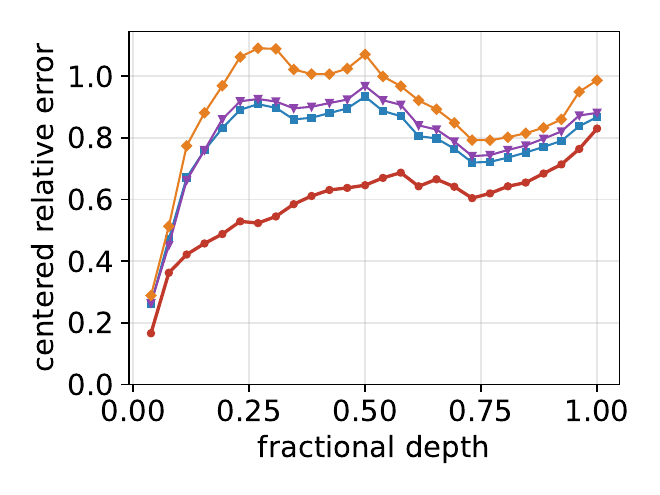}
    \includegraphics[width=0.42\linewidth]{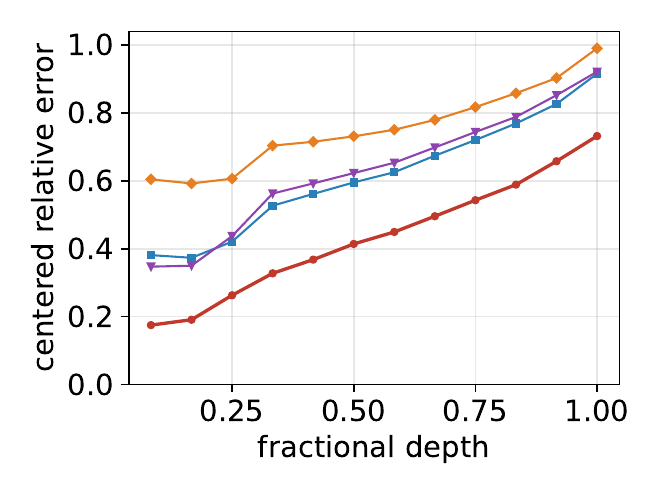}\\
    \includegraphics[width=0.42\linewidth]{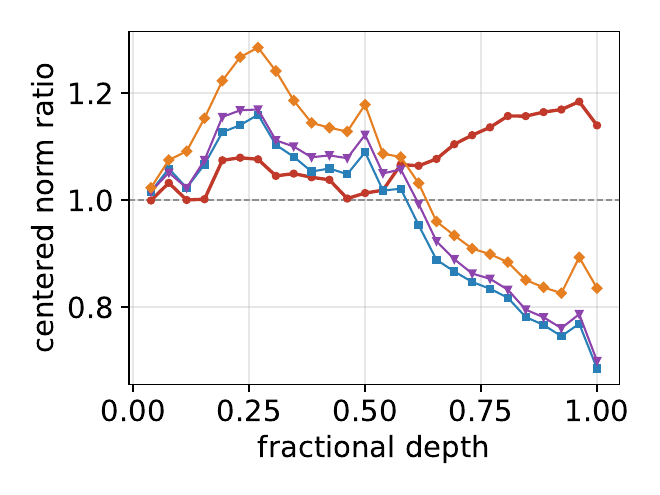}
    \includegraphics[width=0.42\linewidth]{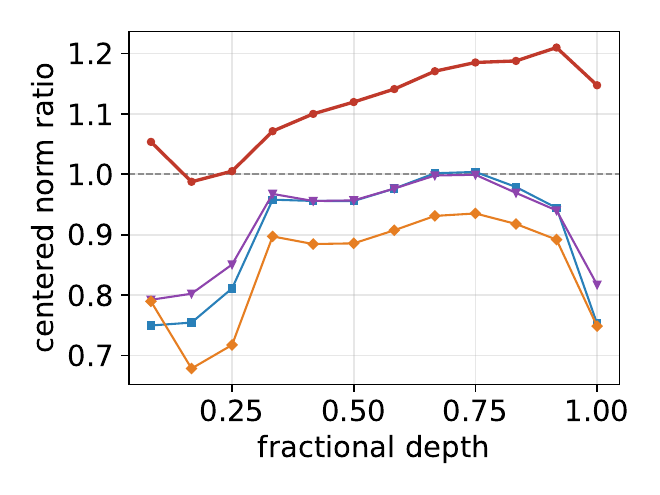}\\[4pt]
    \includegraphics[width=0.85\linewidth]{figures/legend_recipe.pdf}
    \caption{Mean-field predictions are more accurate than the attention-recipe ablations. Left: Gemma-2, Right: GPT-2. Rows (top to bottom): centered cosine similarity, representation similarity analysis (RSA), centered relative error, and centered norm ratio.}
    \label{fig:mf-vs-query-options-full}
\end{figure}

\S\ref{sec:validation} reports only the first of these four metrics.
Figure~\ref{fig:mf-vs-alternative-kernels} in \S\ref{sec:validation} shows centered
cosine for the kernel-source substitutions and
Figure~\ref{fig:mf-vs-query-options} shows it for the attention-recipe ablations;
Figures~\ref{fig:mf-vs-alternative-kernels-full}
and~\ref{fig:mf-vs-query-options-full} below repeat both comparisons on all four
metrics, for Gemma-2,
GPT-2 and Qwen-3-14B.

\section{Anisotropy of Representations}
\label{app:anisotropy}
As noted in Section~\ref{sec:validation}, in both models the cloud of type centroids sits far from the origin relative to its own size. 
In particular, the all-type centroid $\bar{\v{c}}$ grows in magnitude from $35$ to $379$ across GPT-2's layers and from $25$ to $424$ across Gemma-2's, while the mean Euclidean distance of a type centroid to $\bar{\v{c}}$ stays a fraction of that magnitude --- around $0.8\,\|\bar{\v{c}}\|$ in GPT-2's early and middle layers and falling to $0.3\,\|\bar{\v{c}}\|$ at the last layer;  in Gemma-2, it starts above 1.0 but settles around $0.6$--$0.7\,\|\bar{\v{c}}\|$ at deeper layers.
Table~\ref{tab:anisotropy} reports, per layer for both models, the magnitude of the all-type centroid $\bar{\v{c}}=\tfrac1N\sum_t \v{c}_t$, the mean Euclidean distance $\|\v{c}_t-\bar{\v{c}}\|$ of a type centroid from the cloud center, and their ratio $\text{mean dist}/\|\bar{\v{c}}\|$. The table shows that representations occupy a compact region offset from the origin, the motivation for mean-centering before computing cosine similarities.

\begin{table}[H]
\centering\small
\caption{Anisotropy of centroid representations, per layer: the magnitude of the
all-type centroid $\|\bar{\v{c}}\|$, the mean Euclidean distance of a type centroid
from the cloud center, and their ratio.  Gemma-2's layers appear in two columns.}
\label{tab:anisotropy}
\setlength{\tabcolsep}{4pt}
\begin{tabular}{rccc@{\hskip 2.5em}rccc@{\hskip 1.25em}rccc}
\toprule
\multicolumn{4}{c}{GPT-2} & \multicolumn{8}{c}{Gemma-2} \\
\cmidrule(lr){1-4}\cmidrule(lr){5-12}
$\ell$ & $\|\bar{\v{c}}\|$ & dist. & ratio & $\ell$ & $\|\bar{\v{c}}\|$ & dist. & ratio & $\ell$ & $\|\bar{\v{c}}\|$ & dist. & ratio \\
\midrule
0 & 2.4 & 2.7 & 1.12 & 0 & 25.5 & 78.3 & 3.07 & 14 & 114.5 & 72.3 & 0.63 \\
1 & 34.8 & 26.6 & 0.76 & 1 & 33.8 & 66.8 & 1.97 & 15 & 120.7 & 77.9 & 0.65 \\
2 & 39.5 & 31.8 & 0.81 & 2 & 50.6 & 61.8 & 1.22 & 16 & 132.8 & 90.5 & 0.68 \\
3 & 44.9 & 35.2 & 0.78 & 3 & 43.7 & 60.2 & 1.38 & 17 & 154.4 & 101.4 & 0.66 \\
4 & 47.8 & 38.3 & 0.80 & 4 & 49.9 & 55.3 & 1.11 & 18 & 157.0 & 110.0 & 0.70 \\
5 & 50.5 & 40.8 & 0.81 & 5 & 47.0 & 56.1 & 1.19 & 19 & 177.7 & 127.1 & 0.72 \\
6 & 54.1 & 44.1 & 0.81 & 6 & 50.2 & 54.8 & 1.09 & 20 & 185.3 & 139.0 & 0.75 \\
7 & 58.9 & 48.1 & 0.82 & 7 & 51.4 & 55.8 & 1.09 & 21 & 213.9 & 153.5 & 0.72 \\
8 & 67.0 & 53.7 & 0.80 & 8 & 64.7 & 56.8 & 0.88 & 22 & 248.8 & 174.0 & 0.70 \\
9 & 77.4 & 59.9 & 0.77 & 9 & 68.8 & 58.1 & 0.84 & 23 & 279.4 & 187.1 & 0.67 \\
10 & 99.1 & 67.9 & 0.69 & 10 & 78.6 & 60.5 & 0.77 & 24 & 314.6 & 203.1 & 0.65 \\
11 & 166.3 & 78.6 & 0.47 & 11 & 82.8 & 62.0 & 0.75 & 25 & 361.4 & 222.1 & 0.61 \\
12 & 379.1 & 114.8 & 0.30 & 12 & 96.2 & 64.7 & 0.67 & 26 & 424.3 & 280.6 & 0.66 \\
 & & &  & 13 & 94.5 & 65.5 & 0.69 &  & & &  \\
\bottomrule
\end{tabular}
\end{table}

 \section{Impact of Rogue Dimensions on Results}
\label{app:rogue}

In every model, a handful of the $D$ dimensions of the representation $\x_i\in\mathbb{R}^D$
carry magnitudes far above the rest, and through the block's normalization a
small error in one perturbs every dimension on the next rollout step. 
This appendix examines
the effect of those ``rogue'' dimensions. Setting them to their corpus value (which we call `pinning') 
is a small, principled
correction that modestly improves most models (\S\ref{app:rogue-pin}).  Note, though, that the
results in the body all use the unmodified (un-pinned) representations, and the effects of the correction are
examined here in the appendix only. As an aside, we note that these rogue dimensions also seem to
play a role in the instability results  for Llama-3.2-3B that we analyze in Appendix~\ref{app:llama}.

\subsection{Rogue-Dimension Pinning: a Modest Correction}
\label{app:rogue-pin}

\paragraph{The phenomenon.} Every transformer we examine carries a handful of
dimensions one to two orders of magnitude above the median: referred to as
``rogue dimensions'' of GPT-2 and
BERT~\citep{timkey2021rogue,kovaleva2021bert,puccetti2022outliers}, 
``outlier features'' that emerge with scale~\citep{dettmers2022int8}, and 
``massive activations'' documented across the LLaMA family and other modern
models~\citep{sun2024massive}.  The same dimensions underlie the representation
anisotropy of \citet{gao2019degeneration} and \citet{mu2018allbutthetop}.
Because each transformed block normalizes its input by a scale computed over all $D$
dimensions (root-mean-square for the RMSNorm models, standard deviation for
GPT-2's LayerNorm), a rogue dimension dominates that scale, so a small relative
error in its predicted value perturbs every other dimension on the following
step. Pinning these dimensions to their corpus value removes that coupling.

\paragraph{Why removing them is justified.} These dimensions are approximately
\emph{context-invariant}, taking nearly the same value whatever the input token,
and act as an implicit bias or attention-sink mechanism rather than as carriers
of token values~\citep{sun2024massive}. These dimensions have a direct link to attention sinks (the
tendency of heads to deposit excess probability mass on a few uninformative
positions)~\citep{xiao2024streaming,gu2025sink}.
\citet{bondarenko2023quantizable} show that this ``do-nothing'' routing of attention is what
produces the outlier activations. Holding these dimensions at their corpus value during the
rollout thus mainly removes a nuisance term, rather than a signal the mean field is asked
to reproduce.

\paragraph{Identifying the set.} We pin a dimension if at some layer its
per-dimension RMS over the token centroids exceeds $15\times$ that layer's
median, a threshold set by the wide magnitude gap separating these dimensions
from the bulk; it selects at or below $1\%$ of $D$ in each model
(Table~\ref{tab:rogue-dims}). The specific indices are in the code release.

\begin{table}[t]
\centering\small
\caption{Number of dimensions pinned during the rollout %
by model, and against model width $D$.}
\label{tab:rogue-dims}
\begin{tabular}{lcc}
\toprule
model & $D$ & pinned dimensions \\
\midrule
GPT-2         & $768$  & $10$ \\
Gemma-2-2B    & $2304$ & $25$ \\
Llama-3.2-3B  & $3072$ & $10$ \\
Llama-3.1-8B  & $4096$ & $14$ \\
Qwen-3-14B    & $5120$ & $53$ \\
\bottomrule
\end{tabular}
\end{table}

\paragraph{Effect of pinning.} Pinning is a refinement that produces a modest improvement 
in accuracy of rollout prediction. Pinning leaves the
teacher-forced prediction and the early rollout layers unchanged, and it
improves the final-layer rollout for the models whose rollout error runs through
these dimensions: for GPT-2 the centered cosine rises $0.76\!\to\!0.84$
(relative error $0.73\!\to\!0.57$), for Gemma-2 $0.70\!\to\!0.82$
($0.83\!\to\!0.56$), and for Qwen-3-14B $0.72\!\to\!0.82$. For the two Llama
models it is neutral to slightly negative (Llama-3.1-8B $0.72\!\to\!0.71$,
Llama-3.2-3B $0.60\!\to\!0.58$), because their rollout error is not dominated by
these dimensions, but by the mechanism in Appendix~\ref{app:llama}.

\subsection{Rogue Dimensions and the Qwen-3-14B Ablations}
\label{app:rogue-qwen}

Qwen-3-14B is the one model where pinning changes the
\emph{interpretation} of a result rather than just a quantitative improvement. The
mean-field rollout of Qwen-3-14B is barely affected by pinning.  However,
three substitute kernels whose rollouts
drifted far off course (\texttt{Pavg}, \texttt{cooc}, and \texttt{rowshuffle})
as shown in Figure~\ref{fig:mf-vs-alternative-kernels-full} (right-hand column),
are dramatically improved (and the remaining four improve marginally)
(Table~\ref{tab:qwen-pinned}). The rogue dimensions are a conditional
amplifier: with the mean-field kernel the predicted norm never leaves the stable
regime, whereas some substitute kernels drive the trajectory off course enough for the
mispredicted rogue values, amplified through the shared RMSNorm scale to produce a
discrete blow-up.  In those cases, pinning removes the blowup. 
Rogue dimensions therefore do not penalize every
kernel uniformly; they engage only once a prediction has already drifted, which
is why the mean-field kernel escapes their effect.

\begin{table}[t]
\centering\small
\caption{Qwen-3-14B rollout under each kernel, unpinned versus pinned:
cross-layer mean centered cosine (standard error over the 40 layers) and mean
centered relative error.}
\label{tab:qwen-pinned}
\begin{tabular}{lcccc}
\toprule
 & \multicolumn{2}{c}{centered cosine} & \multicolumn{2}{c}{relative error} \\
\cmidrule(lr){2-3}\cmidrule(lr){4-5}
kernel & unpinned & pinned & unpinned & pinned \\
\midrule
Mean field & 0.816 (0.009) & 0.826 (0.009) & 0.640 & 0.721 \\
\midrule
\texttt{cooc} & 0.258 (0.020) & 0.498 (0.008) & 1.429 & 0.954 \\
\texttt{Pavg} & 0.114 (0.036) & 0.622 (0.009) & 24.42 & 0.892 \\
\texttt{rowshuffle} & 0.245 (0.009) & 0.456 (0.006) & 1.042 & 0.969 \\
\texttt{meanq} & 0.430 (0.014) & 0.503 (0.007) & 0.949 & 0.928 \\
\texttt{P0only} & 0.591 (0.024) & 0.628 (0.021) & 1.091 & 1.027 \\
\texttt{uniform} & 0.061 (0.027) & 0.089 (0.029) & 96.11 & 31.91 \\
\bottomrule
\end{tabular}
\end{table}

The sensitivity of Qwen-3-14B to rogue dimensions is almost entirely a property
of a single layer. Applying a substitute kernel to a chosen subset of
layers and the mean-field kernel everywhere else localizes the damage
(Table~\ref{tab:qwen-layers}): substituting at layer~0 alone incurs
$0.279$ of the $0.327$ all-layers total (an $85\%$ share), while substituting at
all thirty-nine \emph{other} layers costs only $0.065$ (a $20\%$ share). The
two shares sum to $105\%$, so the decomposition is nearly clean, and layer~0
carries more than four times the damage of the other thirty-nine combined. The
same measurement on GPT-2 spreads the damage across depth instead, as one
expects of a rollout whose error accumulates layer by layer. Qwen's apparent
hypersensitivity to kernel substitution is therefore an early-layer property
rather than depth compounding, in the same family as the layer-1 amplification
that destabilizes Llama-3.2-3B's rollout (Appendix~\ref{app:llama}), though by
a different mechanism.

\begin{table}[t]
\centering\small
\caption{Localizing the damage from a kernel substitution in
Qwen-3-14B, pinned. The substitute (\texttt{cooc}) is applied only to the
listed layers, with the measured kernel used everywhere else; entries are
cross-layer mean centered cosine over the 40 layers and the shortfall
against the measured kernel.}
\label{tab:qwen-layers}
\begin{tabular}{lcc}
\toprule
layers given the substitute kernel & centered cosine & damage \\
\midrule
none (measured kernel) & 0.826 (0.009) & --- \\
all 40 layers & 0.498 (0.008) & 0.327 \\
layer 0 only & 0.547 (0.008) & 0.279 \\
all layers except 0 & 0.761 (0.011) & 0.065 \\
\bottomrule
\end{tabular}
\end{table}

Pinning is not uniformly beneficial: the mean field's own relative error
\emph{worsens} slightly under it ($0.640\to0.721$) even as its cosine improves,
and \texttt{uniform} stays broken when pinned, so the amplifier is not the only
thing defeating that kernel.

\section{The Llama Models: Accurate Per-Layer Prediction, Failed Rollout}
\label{app:llama}

We extended the mean-field validation of Section~\ref{sec:validation} to
Llama-3.2-3B and Llama-3.1-8B (both RMSNorm, RoPE, grouped-query attention;
statistics from a 2M-token OpenWebText pass over the top-1000 Llama token
types). We find that per-layer accuracy matches the other models (Figure \ref{fig:tf-attention}).
However, the open-loop rollout does not, and 
exposes a failure mode specific to Llama.  We work out this dependency here, 
through a sequence of targeted experiments.  While our experiments serve
to \emph{eliminate} a number of natural hypotheses, we ultimately lack a crisp explanation. 
We report the eliminations along with the
final characterization, since several of the eliminated explanations could 
have been reasonable hypotheses \emph{a priori}.

\paragraph{Summary of results.} First, we affirm that the mean-field yields reasonably accurate predictions layer-by-layer; it is only when errors accumulate through the open-loop rollout that predictions diverge.
We show using teacher-forced predictions that the mean field predicts each
layer's centroid update for both Llamas about as well as for the other models:
cross-layer mean centered cosine $0.858$ ($\pm 0.017$) for Llama-3.2-3B and
$0.837$ ($\pm 0.019$) for Llama-3.1-8B, with the familiar mid-stack dip
(minima $0.69$ and $0.62$) and recovery. 

\begin{figure}[t]
    \centering
    \includegraphics[width=0.42\linewidth]{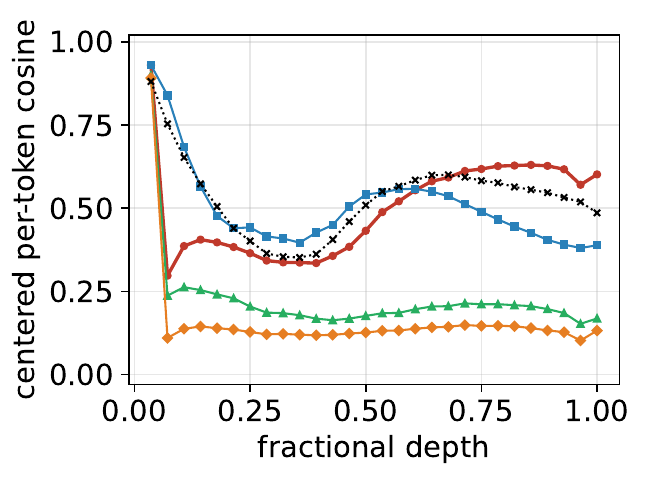}
    \includegraphics[width=0.42\linewidth]{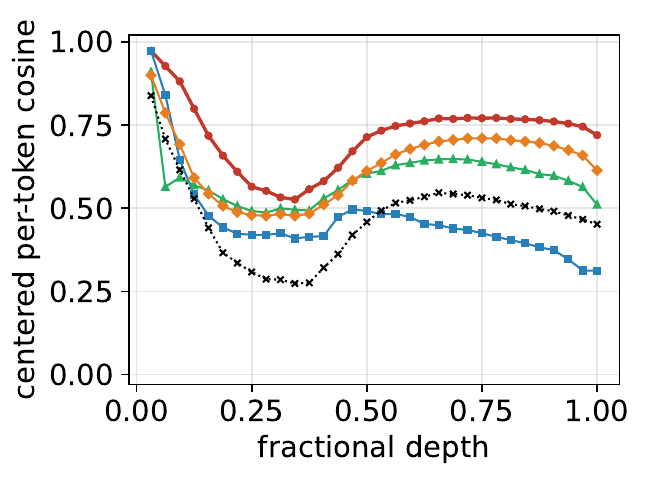}\\
    \includegraphics[width=0.85\linewidth]{figures/legend_source.pdf}
    \caption{Kernel-source substitutions (as Figure~\ref{fig:mf-vs-alternative-kernels})
    for the Llama models. Left: Llama-3.2-3B; right: Llama-3.1-8B. Centered
    per-token cosine of the open-loop rollout by fractional depth.}
    \label{fig:llama-kernels}
\end{figure}

\begin{figure}[t]
    \centering
    \includegraphics[width=0.42\linewidth]{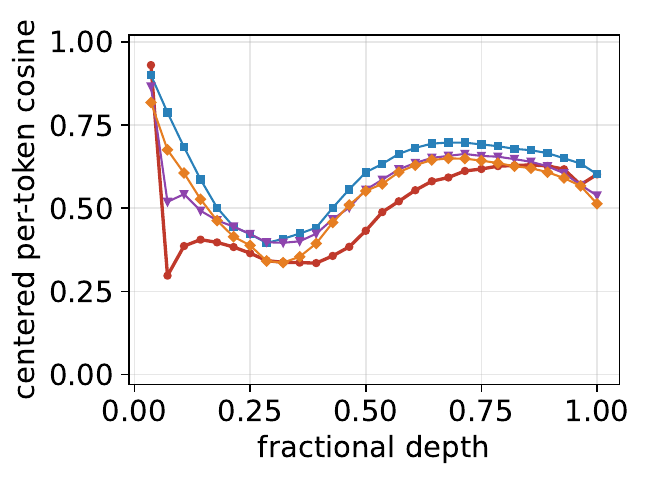}
    \includegraphics[width=0.42\linewidth]{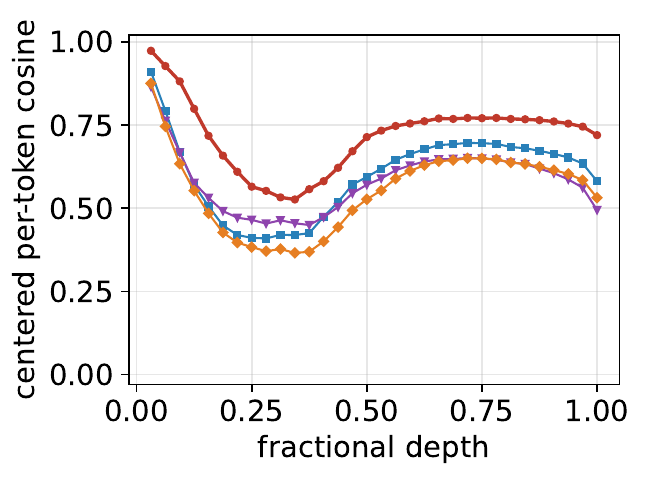}\\
    \includegraphics[width=0.85\linewidth]{figures/legend_recipe.pdf}
    \caption{Attention-recipe ablations (as Figure~\ref{fig:mf-vs-query-options})
    for the Llama models. Left: Llama-3.2-3B; right: Llama-3.1-8B.}
    \label{fig:llama-recipe}
\end{figure}

\begin{figure}[t]
    \centering
    \includegraphics[width=0.85\linewidth]{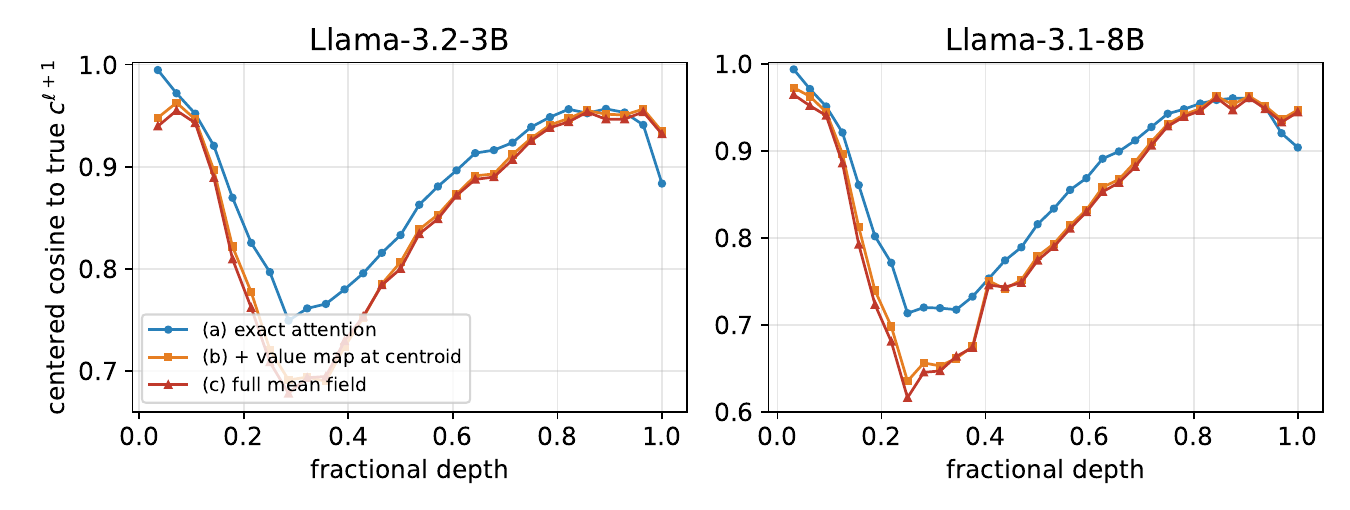}
    \caption{Per-layer (teacher-forced) error decomposition (as
    Figure~\ref{fig:error-decomp}) for the Llama models: (a) exact attention
    with the MLP at the centroid; (b) additionally the value map at the
    centroid; (c) the full mean field. Left: Llama-3.2-3B; right:
    Llama-3.1-8B. Note the lower vertical scale than
    Figure~\ref{fig:error-decomp}.}
    \label{fig:llama-decomp}
\end{figure}

Figures~\ref{fig:llama-kernels}--\ref{fig:llama-recipe} give the Llama
counterparts of Figures~\ref{fig:mf-vs-alternative-kernels}
and~\ref{fig:mf-vs-query-options}, and Figure~\ref{fig:llama-decomp} the
counterpart of Figure~\ref{fig:error-decomp}. 
Iterated open-loop, Llama-3.1-8B
behaves approximately like GPT-2, Gemma-2, and Qwen-3: cross-layer mean rollout cosine
$0.719$ ($\pm 0.019$) with cosine similarity at the final-layer of $0.72$. 
Notably, Llama-3.2-3B instead
\emph{diverges}: the predicted centered magnitude inflates to $4.9\times$ the
true value, the rollout cosine collapses to $0.30$ by layer~2 and averages
$0.499$ ($\pm 0.028$), and the centered relative error exceeds $1$ from
mid-stack on. 

The teacher-forced prediction is not severely affected (Figure~\ref{fig:llama-decomp}).  
These per-layer measurements remain the important mean-field fidelity measure for this
model.  The rest of this section
concerns why the \emph{iteration} fails.

\paragraph{The divergence is a single-step amplification in the
layer-1 MLP.} Tracking the rolled-out and true trajectories together, the
error entering the layer-1 step is amplified $17\times$ in that one step and
thereafter merely persists (${\sim}1.05\times$ per later step). We note three properties of the error. (i) It is a \emph{curvature} effect and not a
large derivative. Rescaling the incoming error to $\alpha\delta$ and re-running
the block, the amplification grows nonlinearly with the amount that we scale the error.
So the block's derivative at the true
centroid is benign in the direction the error actually points --- at
small amplitude the real error and a random one are amplified almost equally
($1.6\times$ versus $1.2\times$). What matters is that when the error is large
enough, $36\%$ of the centroid norm,  it leaves the region over which the MLP is
affine. 
(ii) It is \emph{in the MLP}: decomposing the step, the
attention update expands the error by $1.06\times$; the subsequent MLP takes it
to $17\times$. (iii) It is \emph{model-specific}: the same measurement gives
${\sim}1\times$ at every layer for GPT-2, Gemma-2, Qwen-3-14B, and
Llama-3.1-8B. Further, the rogue-dimension pinning of Appendix~\ref{app:rogue} does not
affect the error (final rollout cosine $0.602 \to 0.584$ for the 3B, $0.720 \to 0.708$
for the 8B): the amplified error is spread broadly across coordinates, with the
largest massive-activation channel contributing only $1.2\%$ of the error.

Further investigation identifies that only a few gates (approximately 25) in the first MLP are responsible
for the great majority of the added error.
Those few gates account for the amplification outright. 
Because the MLP
output is exactly linear in the post-activation, the response decomposes exactly
over the MLP's $8192$ gates; restricting it to the top 25 recovers $94.7\%$ of
the \emph{nonlinear excess}, while an equal number of randomly chosen gates recovers $0.1\%$.

\paragraph{The amplified error concentrates on closed-class tokens.} The
per-token amplification is heavy-tailed: the median token's error grows
$12\times$, the $99$th percentile $143\times$, the maximum $323\times$, and the
ten most-amplified tokens carry $53\%$ of the total output-error energy. Of the
fifty most-amplified, forty-three are newlines, punctuation, or
clause-boundary tokens such as parentheses or ``The.''.  These are all `closed-class' tokens: 
high-frequency function words and structural markers rather than content words.
Many start at early layers with tiny errors; 
the extreme case is token \texttt{`).\textbackslash n\textbackslash n'} whose centroid error norm is  $0.49$ going into transformer block 1 and error $159$ coming out of that block.

\paragraph{Exclusion experiment: structural tokens are not the cause.} The
concentration on boundary tokens suggests an obvious hypothesis: the structural
tokens themselves drive the failure. We tested it by rerunning the entire
pipeline --- token selection, corpus statistics, and rollout, on a fresh
1M-token extraction --- with every newline, whitespace, and pure-punctuation
token excluded from the tracked vocabulary ($67$ of the original top-1000;
capitalized words that initialized a clause were retained deliberately, as a probe). The
divergence is unchanged: peak magnitude ratio $5.2$ versus $4.9$, mean rollout
cosine $0.426$ versus $0.499$, teacher-forced accuracy unchanged ($0.852$).
Token amplification simply \emph{migrates} to remaining closed-class tokens: conjunctions (\texttt{but}, \texttt{ and}, \texttt{ or}),
auxiliaries (\texttt{is}, \texttt{ are}, \texttt{ has}, \texttt{ was}),
negation stems 
(\texttt{wasn}, \texttt{ didn}, \texttt{ doesn}), determiners
(\texttt{the}, \texttt{ a}), and sub-word prefixes (\texttt{p},
\texttt{ de}) --- only one of the top twenty is a capitalized clause-initial
word --- with the same concentration (top ten $=54\%$ of error energy).
The same rerun for Llama-3.1-8B likewise changes nothing material (mean
rollout cosine $0.615$ versus $0.719$, peak magnitude ratio $1.45$ versus
$1.70$, teacher-forced unchanged at $0.827$).

\paragraph{Attention-sink accounts, tested.} The coincidence of boundary
tokens, massive-activation channels, and layer~1 (where sink structure first
forms) suggests the attention sink~\citep{xiao2024streaming,sun2024massive} as
a common cause. However, three versions of this idea all give insufficient explanations. 

The first version is that boundary tokens develop large errors because they are local attention sinks. 
In fact, boundary tokens genuinely are \emph{local} sink targets --- they receive
$1.4$--$1.7\times$ more attention per occurrence than content tokens.
However, the queries that feed them are not the queries that feed BOS, so boundary tokens are not
standing in for the sink.  Further, across tokens, how strong a ``magnet'' a token is does not predict its error amplification.

The second version is that boundary tokens develop large errors because they carry the 
massive activations.  However, we find that boundary tokens' median magnitude in the
massive-activation dimensions is 0.71, vs 0.73 for content tokens.   So they are not
unusual in their massive activation dimensions.

The third version is that boundary tokens are a mixture of tokens with high values of the massive
activation dimension and lower values of the massive activation dimension.  This would cause the MLP at the centroid to make strongly incorrect predictions.   However, we find that the massive-activation dimension values are unimodel across boundary tokens.   What survives is only a graded boundary signal  in the massive activation dimension ($1.2$--$1.8\times$ higher for phrase- and sentence-initial or -final
occurrences), too weak and too continuous to account for the amplification.

\paragraph{Capitalization pairs: position is not the driver.} As a final
positional test, we compared mean-field deviations across all $77$
capitalized/lowercase pairs in the vocabulary (\texttt{ The}/\texttt{ the},
\texttt{ In}/\texttt{ in}, \ldots), capitalization being a near-perfect marker
of sentence-initial position. There is no systematic effect: the paired median
detonation ratio is $1.00$ (Wilcoxon $p=0.37$), and teacher-forced error is
actually slightly \emph{lower} for the capitalized forms ($0.81\times$,
$p<10^{-4}$). The tails are large in both directions --- \texttt{ The}
detonates $3.6\times$ its lowercase twin and paragraph-initial \texttt{The}
$7.1\times$, but \texttt{ and} detonates $50\times$ more than \texttt{ And} ---
so the well-known \texttt{ The} case is atypical rather than representative.

\paragraph{What remains.} The Llama-3.2-3B divergence is fully characterized: it
is confined to a single step in the layer-1 MLP, specific to this model, and
heavy-tailed over closed-class tokens, with its output concentrated in the
massive-activation dimensions.
The natural explanations are eliminated: it is not the structural tokens
(exclusion experiment), not rogue-dimension coupling (pinning inert), not
sink-magnet status, not a sink-writing gate, and not syntactic position
(capitalization pairs). What is left is a curvature effect in this one
model's layer-1 MLP, and it is now accounted for: the layer-0 mean-field error
of high-frequency, closed-class tokens is concentrated enough to drive a handful
of SwiGLU gates across their nonlinearity, and those gates carry essentially all
of the resulting amplification. This is the MLP-at-centroid (Jensen)
approximation, identified in Section~\ref{sec:validation} as the leading
per-step error source in every model, meeting an error large enough to leave
the affine region. Why this model, and this layer, place their gates so close to
the knee where the others do not, we leave open.

\section{Impact of the MLP on Mean-Field Approximation}
\label{app:mlp-gap}

The last step of Algorithm~\ref{alg:rollout}
applies the layer's MLP once, to the mean intermediate representation $\v{m}_s$
($\mathrm{MLP}(\mathrm{LN}(\v{m}_s))$), in place of averaging the MLP over the
individual intermediate tokens of type $s$
($\mathbb{E}_{\mathrm{occ}}[\mathrm{MLP}(\mathrm{LN}(\v{m}))]$). To quantify the
resulting error we accumulate both quantities per type and layer over a corpus
and compare them.
Table~\ref{tab:mlp-gap} reports, per layer, the mean-centered cosine and relative error (ratio of means) between the two quantities, averaged over types. For both GPT-2 and Gemma-2, the behavior follows a similar pattern.  The approximation is near-exact at the first layers (cosine $\approx 0.99$ for both models), weakest in the middle of the network (cosine $\approx 0.65$ for GPT-2 at layers 4--5, $\approx 0.73$ for Gemma-2 at layer 11), and improved somewhat at later layers.  Overall, the centered cosine averaged over layers is $0.77$ (GPT-2) and $0.86$ (Gemma-2).

These numbers are what make the MLP step the dominant term in the error
decomposition of \S\ref{sec:nae}, yet a cosine of $0.65$ (the bark) sounds far worse than the
per-layer shortfall of $0.02$--$0.03$ (the bite) reported there.
Two things account for the difference.
The cosines above measure the MLP contribution against itself, whereas the
decomposition measures the predicted centroid, which carries the attention output
alongside the MLP contribution, so a given disagreement in the MLP term costs less in
the centroid than it does on its own.
And most of the disagreement is a shared offset in \emph{magnitude} rather than an
error in \emph{direction}.

\begin{table}[t]
\centering\small
\caption{MLP mean-field approximation, per layer: mean-centered cosine and relative
error (ratio of means) between the occurrence-mean MLP contribution for type $s$ and
the MLP applied to the mean intermediate token $\v{m}_s$, averaged over types.
Gemma-2's layers continue in the right-hand block. }
\label{tab:mlp-gap}
\begin{tabular}{rcc@{\hskip 3em}rcc@{\hskip 1.5em}rcc}
\toprule
\multicolumn{3}{c}{GPT-2} & \multicolumn{6}{c}{Gemma-2} \\
\cmidrule(lr){1-3}\cmidrule(lr){4-9}
$\ell$ & cos & rel.\ err & $\ell$ & cos & rel.\ err & $\ell$ & cos & rel.\ err \\
\midrule
0 & 0.995 & 0.333 & 0 & 0.983 & 0.190 & 13 & 0.806 & 0.866 \\
1 & 0.966 & 0.360 & 1 & 0.952 & 0.411 & 14 & 0.856 & 0.636 \\
2 & 0.828 & 0.717 & 2 & 0.947 & 0.381 & 15 & 0.894 & 0.591 \\
3 & 0.742 & 0.865 & 3 & 0.892 & 0.606 & 16 & 0.887 & 0.613 \\
4 & 0.650 & 1.073 & 4 & 0.877 & 0.662 & 17 & 0.913 & 0.555 \\
5 & 0.650 & 1.118 & 5 & 0.815 & 0.970 & 18 & 0.894 & 0.643 \\
6 & 0.686 & 1.045 & 6 & 0.860 & 0.609 & 19 & 0.868 & 0.668 \\
7 & 0.752 & 0.903 & 7 & 0.791 & 1.025 & 20 & 0.880 & 0.595 \\
8 & 0.778 & 0.856 & 8 & 0.778 & 0.917 & 21 & 0.885 & 0.563 \\
9 & 0.748 & 0.859 & 9 & 0.765 & 0.965 & 22 & 0.861 & 0.637 \\
10 & 0.773 & 0.783 & 10 & 0.766 & 0.898 & 23 & 0.860 & 0.609 \\
11 & 0.697 & 0.909 & 11 & 0.726 & 0.917 & 24 & 0.871 & 0.656 \\
 &  &  & 12 & 0.795 & 0.754 & 25 & 0.829 & 0.756 \\
\bottomrule
\end{tabular}
\end{table}

\section{Replication on OLMo-2, and Evidence of a Shared Mean-Field Ceiling}
\label{app:waypoint-olmo}

\begin{figure}[htbp]
\centering
\includegraphics[width=0.7\textwidth]{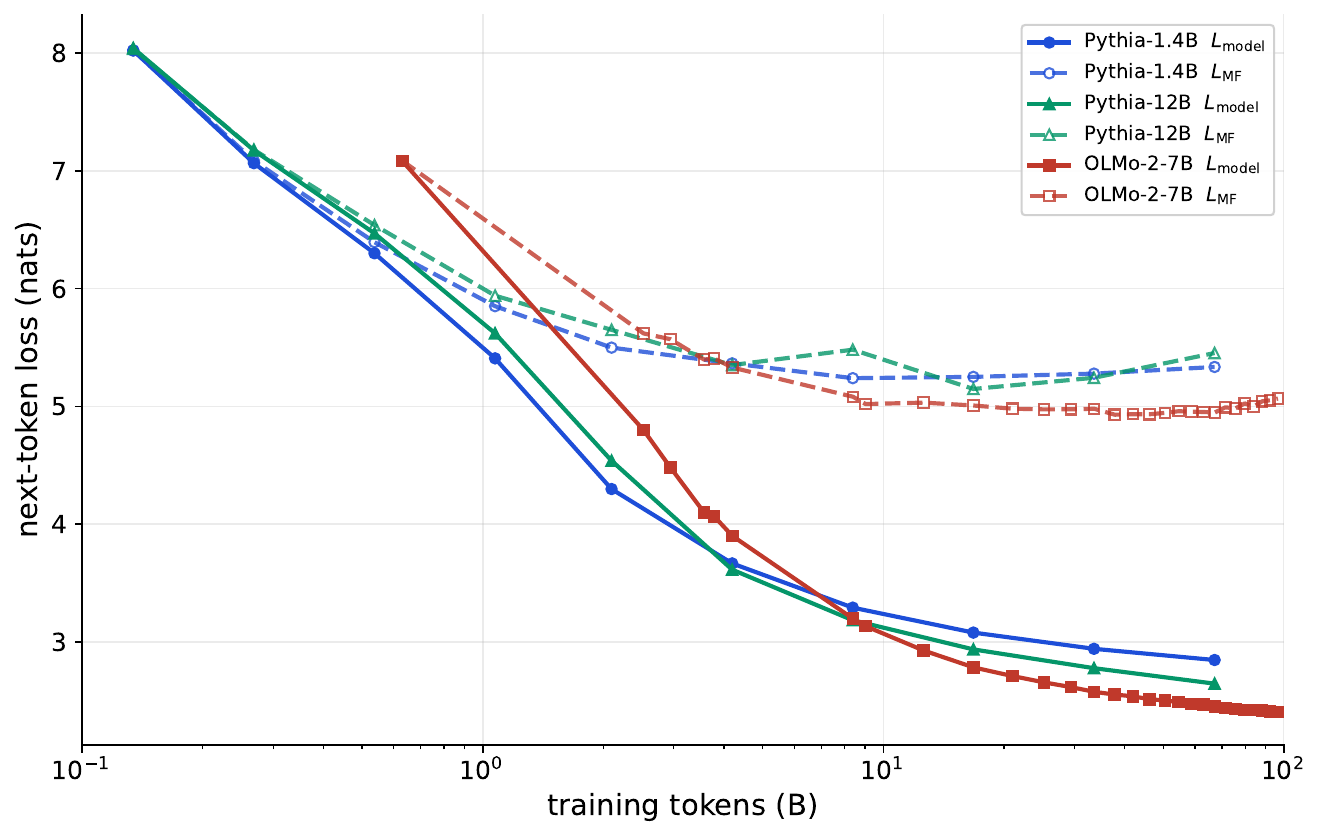}
\caption{Pythia-1.4B, Pythia-12B and OLMo-2-7B on a common axis of
training tokens in billions. 
Solid lines with filled markers are $L_{\mathrm{model}}$;
dashed lines with hollow markers are $L_{\mathrm{MF}}$, the loss when every
head's read is replaced by its per-context mean-field read. Color and marker
shape both identify the model. 
}
\label{fig:crossmodel}
\end{figure}

Our waypoint results in \S\ref{sec:waypoint} were measured on Pythia. To ask whether our findings
were a property
of that suite or of language-model training generally, we repeated the
\S\ref{sec:waypoint} substitution on OLMo-2-7B \citep{olmo2024olmo2}
across its 44 stage-1 checkpoints, and plotted all three models on a common axis
of training \emph{tokens} in Figure~\ref{fig:crossmodel}. A common token axis is
necessary because one step is not the same amount of training in the two suites:
a Pythia step is $2.097$M tokens and an OLMo step is $4.2$M.
OLMo is an
independent architecture, applying RMSNorm to the attention \emph{output} rather
than its input, so that the residual entering attention is unnormalized, and it is
trained on a different corpus.
We measured its statistics exactly as
Pythia's (a 2M-token OpenWebText pass over the top $1000$ types), giving
substitution coverage $0.65$ against Pythia's $0.64$.

At the left of the figure all three models fail to improve over their 
respective mean
fields: $\Delta L = -0.001$ for Pythia-1.4B and $-0.006$ for Pythia-12B at
$0.13$B tokens, and $+0.003$ for OLMo at $0.63$B. A model reducing to its mean field early in training is therefore not unique to Pythia; a
second architecture trained on different data exhibits the same behavior, and to the
same degree: OLMo's deviation there is $\Dmf = 0.0713$, against $0.0720$ for
Pythia-160m at initialization.

The rest of the figure shows that as
training proceeds, the three $L_{\mathrm{model}}$ curves separate and keep
falling, reaching $2.85$, $2.65$ and $2.41$ nats by the right-hand edge. The
three $L_{\mathrm{MF}}$ curves do not: they flatten into a narrow band and stay
there, spanning only $4.93$--$5.45$ nats over the whole decade from $10$B to
$100$B tokens.
The three models differ by a factor of eight in
parameters, and OLMo in training data as well, yet the gap $\Delta L$ at the
right-hand edge is nearly the same for all three: $2.49$, $2.81$ and $2.66$ nats.
A plausible reading is that the loss a
mean field can reach is set by the corpus statistics it is built from rather than by
the model whose attention it replaces: the ceiling is fixed, and the widening gap is
the model moving away from it.

The comparison has limits. OLMo publishes no checkpoint before $0.63$B tokens
and none at initialization, so its curve begins where Pythia's has  already begun to depart from its mean field. Thus, the left-hand decade is Pythia-only. In this window there are ten Pythia
points against twenty-nine OLMo points, so the Pythia curves are the coarser of the
two. All three rest on 2M-token statistics, whose sensitivity we examine next.

\section{Sensitivity to the Top-\texorpdfstring{$N$}{N} Truncation}
\label{app:waypoint-vocab}

To assess the sensitivity to truncation to the top- $N$ types, we re-ran the substitution of
\S\ref{sec:waypoint} for Pythia-160M at $N=2000$ and $N=4000$ types, which raised position
coverage, the quantity of Appendix~\ref{app:coverage}, from $65\%$ to $72\%$ and
$80\%$, respectively.
That substitution resolves the tracked types explicitly and collapses everything else
into the position-$0$ and ``other'' buckets of
Appendix~\ref{app:modeling-details}, each of which carries the measured attention
mass but a single shared value vector; queries outside the tracked set keep the
model's unmodified computation.

\paragraph{Controlling estimation quality.} The kernel entry
$W_{st}=P_{st}/\bar n_{st}$ is estimated from a corpus pass, and its support is
the co-occurrence count $C_{st}$. Enlarging $N$ enlarges the pair set
quadratically while leaving the corpus fixed, so support per pair falls sharply.  Matching $N=1000$'s support at larger $N$ requires roughly $4\times$
the corpus at $N=2000$ and $15$--$20\times$ at $N=4000$, by three separate
criteria (pair coverage, mass coverage, and the worst-supported corner). We
therefore repeated every extraction at \textbf{40M tokens}, twenty times the
original pass and enough to cover the worst case, using the same corpus for all
three arms so that corpus size is not itself a variable.

\begin{figure}[htbp]
\centering
\includegraphics[width=\textwidth]{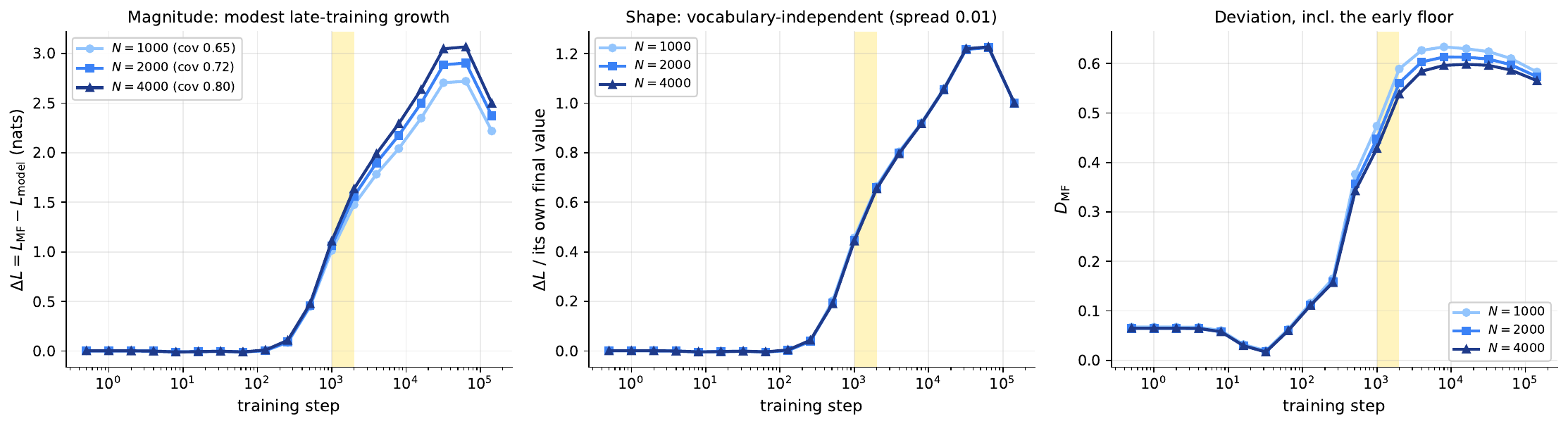}
\caption{The waypoint experiment at $N=1000$, $2000$ and $4000$ keys,
all with 40M-token statistics. \textbf{Left:} $\Delta L$; the arms separate only
late in training, by an amount tracking substitution coverage. \textbf{Centre:}
$\Delta L$ normalized to each arm's own final value, whose shape agrees to
within $0.01$. \textbf{Right:} $\Dmf$, including the early minimum, which is
within $0.003$ across the three. The band marks the induction window as in
Figure~\ref{fig:mfgap}.}
\label{fig:vocab-robustness}
\end{figure}

\paragraph{At matched estimation quality,
the truncation has little effect.}
Over steps $0$--$64$ the three vocabularies give $\Delta L$ within $0.001$ nats of one
another (Figure~\ref{fig:vocab-robustness}). 
When each arm's $\Delta L$ curve is
normalized to its own final value, the three agree in shape 
to within $0.01$
across a fourfold
change in $N$. The phase structure detailed in \S\ref{sec:waypoint} is unchanged.

\begin{figure}[htbp]
\centering
\includegraphics[width=\textwidth]{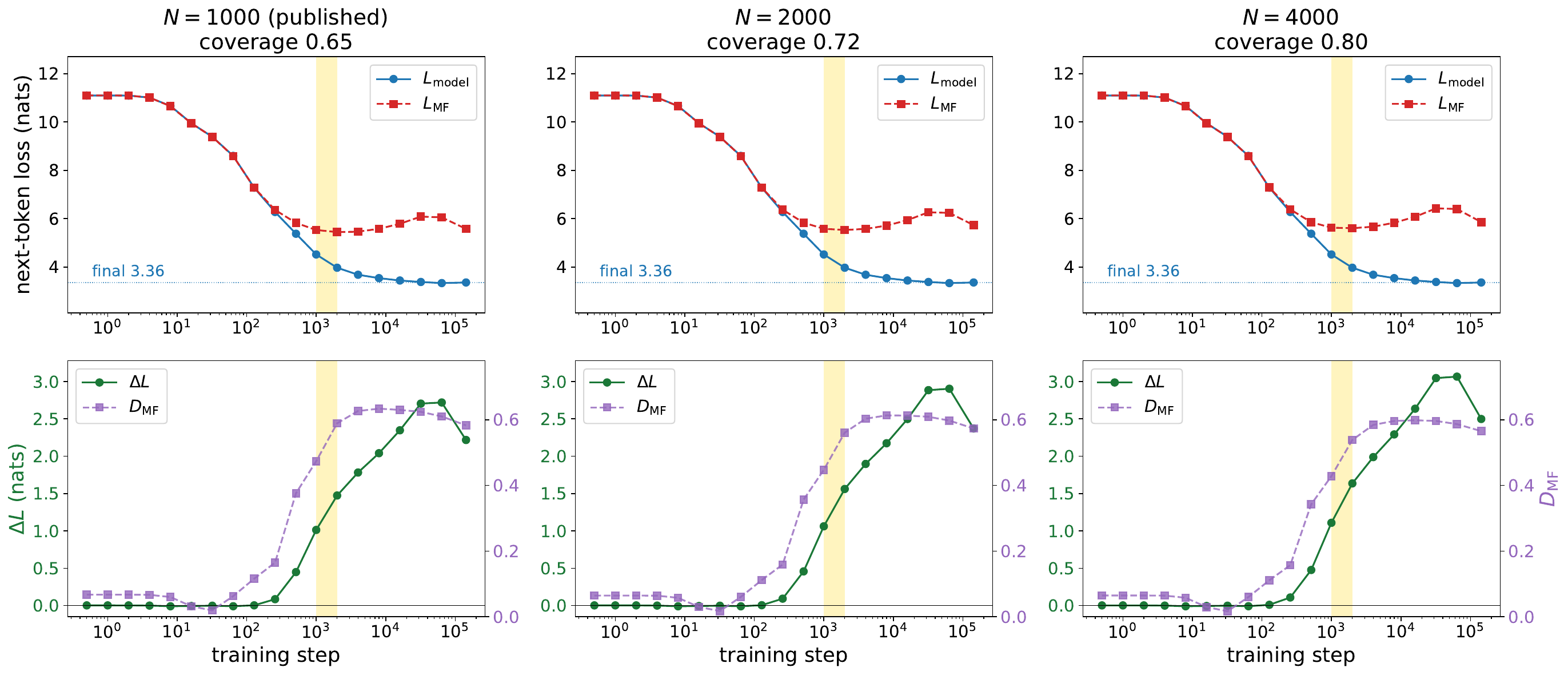}
\caption{The waypoint experiment at $N=1000$, $2000$ and $4000$ keys
(columns), 40M-token statistics throughout, drawn in the layout of
Figure~\ref{fig:mfgap} so the two may be compared panel for panel.
\textbf{Top:} next-token loss of the model ($L_{\mathrm{model}}$, solid) and of
the context-conditional mean field ($L_{\mathrm{MF}}$, dashed); the dotted line
marks the model's final loss, and each column is titled with its substitution
coverage. \textbf{Bottom:} $\Delta L = L_{\mathrm{MF}} - L_{\mathrm{model}}$
(left axis) with the mean per-head deviation $\Dmf$ (right axis). The band marks
the induction window, as in Figure~\ref{fig:mfgap}. $L_{\mathrm{model}}$ is
common to the three columns by construction; the shape of the remaining curves is
likewise unchanged, and only the magnitude of $\Delta L$ grows with coverage.}
\label{fig:vocab-mfgap}
\end{figure}

\paragraph{The full trajectories.} Figure~\ref{fig:vocab-mfgap} redraws the
three arms in the layout of Figure~\ref{fig:mfgap}, with $L_{\mathrm{model}}$ and
$L_{\mathrm{MF}}$ above, $\Delta L$ and $\Dmf$ below, against training step with
the induction window shaded, so each column can be read directly against the
per-model panels there.  We note, first, that 
$L_{\mathrm{model}}$ is the same curve in all three columns, agreeing to
$6\times10^{-7}$ nats at every checkpoint: the substitution vocabulary is a
property of the null model, not of the network, so the model's own loss is a
fixed reference and only $L_{\mathrm{MF}}$ moves. This is a check that the three
arms differ in nothing but the intended variable. Second, every landmark of the
trajectory falls at the same checkpoint irrespective of $N$. $L_{\mathrm{MF}}$
stays within $0.01$ nats of $L_{\mathrm{model}}$ through step $64$ and detaches
between steps $128$ and $256$; it reaches its minimum at step $2000$, rises to a late maximum at step
$32000$, and falls back by step $143000$. 

\section{Kernel Comparison: Estimation Details}
\label{app:waypoint-kernel}

The kernel dynamics of Section~\ref{sec:kernel-dynamics} compare the
per-key kernel $W^{(\tau)} = P^{(\tau)}/\overline{n}$
(Appendix~\ref{app:full-conditional-meanfield}) at checkpoint $\tau$ between every pair of
checkpoints.  The divisor $\overline{n}$ is a pure token count and so is
model-free and identical at every checkpoint; dividing by it removes the
frequency weighting carried by $P$'s entries, leaving changes in the attention
pattern itself.  Similarity is
linear CKA, computed per head and averaged over heads, with each kernel
column-centered beforehand and replaced by a rank-96 SVD approximant (the
CKA is computed exactly from the 96-dimensional factors, and agrees with
the exact value to within 0.001).  All kernel-level comparisons are
restricted to type pairs with context count $\ge 20$; without this support
mask a small number of near-zero-support pairs, whose ratios are unstable,
dominate the quadratic CKA.  Every checkpoint's statistics are extracted
from the same corpus sample with the same chunking, so entries are
directly comparable across checkpoints.

\paragraph{Numerical values.} Adjacent-checkpoint similarity is
0.99--1.00 through the frozen stretch (steps 0--32 at 1.4B and 12B, 0--64 at
160m).  It then falls progressively across the reorganization: at 160m
$0.99 \to 0.86 \to 0.76 \to 0.71$ over steps 32--512, at 1.4B $0.99
\to 0.97 \to 0.79 \to 0.67$, and most steeply at 12B $0.93 \to 0.65 \to 0.45
\to 0.40$, with the minimum near steps 256--2000 in every model and deepest at
12B ($0.40$).  After the transition the two smaller models recover to
${\sim}0.8$ by step 4000 and ${\sim}0.9$ by step 16000; 12B recovers more
slowly ($0.66$ then $0.75$), consistent with its deeper reorganization.

\paragraph{Row uniformity.}
\begin{figure}[t]
\centering
\includegraphics[width=\textwidth]{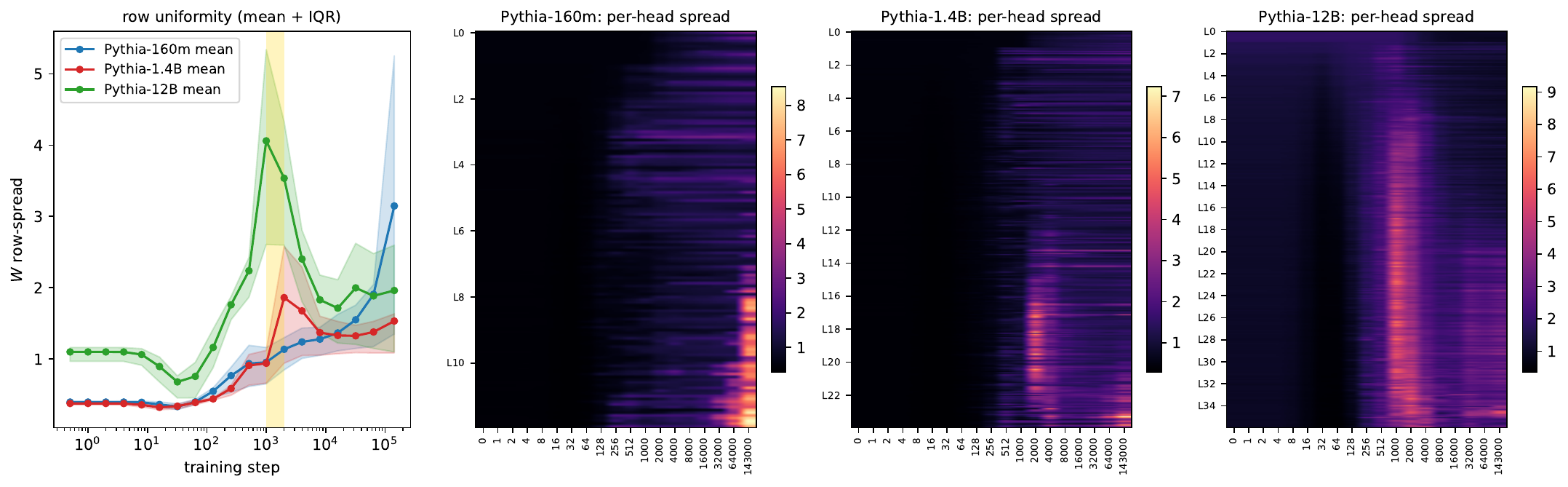}
\caption{How each head divides its attention across token types,
all three models, plotted as fold-variation: a value of one is uniform attention across tokens.  \textbf{Left:} mean and interquartile band across heads.
\textbf{Right three panels:} The same, head by head (160m, 1.4B, 12B).  Head
differentiation begins around step 128; a subset of
near-uniform heads persists throughout, while the mid-training peak grows with
scale (mean fold-variation ${\approx}4$ at 12B).}
\label{fig:spread}
\end{figure}
The row-spread measure is
$\operatorname{mean}_s \operatorname{std}_t(\log W[s,t])$ over the masked
key types, computed per head.  At initialization the mean spread is 0.40
(160m) and 0.38 (1.4B), with all heads within ${\sim}15\%$ of one another;
it dips slightly at the $\Dmf$-minimum checkpoints before head
differentiation begins.

Figure~\ref{fig:spread} plots this measure across training.  The left panel shows the mean spread with its interquartile band across heads; the right three panels show the individual spread of each head, grouped by layer, for the 160m, 1.4B, and 12B models.  Early in training the division is close to even, so a token draws attention in proportion to how often it appears and not otherwise; heads begin to attend selectively when the kernel reorganizes, and the per-head spread then generally increases through the remainder of training.  The heads also differ sharply from one another: some develop large spread while others barely change from their near-uniform initialization.

\section{How Reliably Each Deviation Metric Separates the Prompt Classes}
\label{app:dev-corr}

\begin{table}[th]
\centering\small
\begin{tabular}{llccc}
\toprule
 & & \multicolumn{3}{c}{effect size $g$ against the random control ($n=60$)}\\
\cmidrule(lr){3-5}
model & metric & induction ($n{=}60$) & few-shot ($n{=}48$) & natural ($n{=}60$)\\
\midrule
\multirow{3}{*}{GPT-2}
 & $\Dmf$            & $+3.60$ & $+2.06$ & $+0.04$\rlap{$^{\dagger}$}\\
 & $M_{\text{cov}}$  & $+2.98$ & $+0.67$ & $-0.88$\\
 & $M_{\text{ctx}}$  & $+1.49$ & $+6.84$ & $+2.14$\\
\midrule
\multirow{3}{*}{Gemma-2-2B}
 & $\Dmf$            & $+2.93$ & $+2.21$ & $+1.06$\\
 & $M_{\text{cov}}$  & $+1.21$ & $-0.86$ & $-0.61$\\
 & $M_{\text{ctx}}$  & $+2.36$ & $+7.41$ & $+2.86$\\
\bottomrule
\end{tabular}
\caption{Separation of each prompt class from the random control, as a
standardized mean difference (Hedges' $g$).
Positive means the class exceeds the control; negative means it runs the other
way.  All entries are significant at $p<0.05$ on a Welch test, most far below it,
except the one marked $\dagger$.}
\label{tab:icl-corr}
\end{table}

Section~\ref{sec:icl-correlates} claims that the mean-field deviation $\Dmf$, and its
covariance and contextualization components, respond to in-context learning.  We
test this against four designed prompt classes, induction, few-shot, natural
text, and a random control, asking whether each class separates from the control.
Table~\ref{tab:icl-corr} reports the separation as Hedges' $g$, the gap between a
class's mean and the control's in units of their pooled within-class spread; a
large $g$ means the two are reliably apart, not that the deviation is large.
The deviation $\Dmf$ separates the planted classes from the control in
both models (induction and few-shot $g$ of $+3.60, +2.06$ for GPT-2 and
$+2.93, +2.21$ for Gemma-2).  The two components carry the separation
differently: in GPT-2, $M_{\mathrm{cov}}$ separates induction ($+2.98$) and
$M_{\mathrm{ctx}}$ separates few-shot ($+6.84$), while in Gemma-2,
$M_{\mathrm{ctx}}$ separates every class and $M_{\mathrm{cov}}$ is weak or negative.
This matches the model-dependent split reported in \S\ref{sec:icl}.

Two points qualify the table.  First, the separation is
\emph{cross-class}: $g$ measures how well the deviation ranks these designed
conditions, not individual prompts within a condition.  Within a class the
metrics carry almost no information about $\Delta L_{\mathrm{perm}}$.  Prompt
class alone explains it at $R^2=0.69$ for GPT-2 and $0.58$ for Gemma-2, and
adding any metric raises $R^2$ by at most $0.004$.  A correlation pooled over
prompts would look strong while measuring only the class separation, so we do not
report one.

Second, the separation is uneven across classes.  GPT-2's $\Dmf$ does not
distinguish natural text from the control ($g=+0.04$, $p=0.84$), even though
permuting that text raises its loss ($g=+1.22$).  In GPT-2, then, the deviation
registers planted ICL but not the background ICL of ordinary text.  Gemma-2's
$\Dmf$ separates natural text as well ($+1.06$).

\section{Provenance of the Heads' Roles and  Ranking}
\label{app:heads}

This appendix supplies the provenance of Table~\ref{tab:ranking-cc}: how each
head's documented role was established, and what the readings show at both ends of
the ranking.

\subsection{Lenses for Mean Field Analysis}
\label{sec:instrument}

The lenses below all read the per-occurrence deviation $D^{\ell h}_{sX}$, whose
type-averaged form is the ranking score $\bar S^{\ell h}$ of \eqref{eq:ddec}; each
notes where the per-occurrence and ranking orderings disagree.
The first lens, the covariance/contextualization plane, is the one the body uses
throughout (\S\ref{sec:mf-deviation-decomposed} - \S\ref{sec:catalog-findings}); the
other lenses are instruments for the per-head readings collected here.

\tool{Covariance/contextualization plane.}
\label{tool:axes}
Terms (a) and (b) of \eqref{eq:dev-decomp} separate the deviation by which
factor varies: where a head attends, versus what values the tokens it
attends to carry.
Averaging the per-context shares of \eqref{eq:mcov-mctx} over occurrences rather
than over heads places every head on two axes,
\begin{equation}
  M_{\mathrm{cov}}^{\ell h}=\E_{\text{contexts } X}\!\big[\,\sigma_{\mathrm{cov}}\,
      D^{\ell h}_{sX}\big],
  \qquad
  M_{\mathrm{ctx}}^{\ell h}=\E_{\text{contexts } X}\!\big[\,\sigma_{\mathrm{ctx}}\,
      D^{\ell h}_{sX}\big],
  \label{eq:head-meters}
\end{equation}
with $\sigma_{\mathrm{cov}}$ and $\sigma_{\mathrm{ctx}}$ the two shares of the
error defined at \eqref{eq:mcov-mctx}.
The shares sum to $D^{\ell h}_{sX}$ per occurrence, so the axes and the ranking
score measure the same deviation, split by mechanism rather than two separately
normalized magnitudes.
The meters are signed, so a component that partially cancels the other appears as
a negative coordinate, and the value function includes the head's output
projection, so all quantities are in residual-stream units.

\tool{Top-deviation context mining.}
\label{tool:contexts}
The per-occurrence deviation $D^{\ell h}_{sX}$ can be computed for every
context $X$ in a corpus stream. For each head we keep the top-$K$ results
and record the surrounding text, the head's top attended
keys, and the actual next token.
We filter out occurrences whose deviation
is an artifact of vocabulary truncation rather than of the head's
function; \S\ref{app:protocol} gives the criteria.

\tool{Vocabulary-space decode.}
\label{tool:readout}
To read the deviation vector
$
  \Delta_i \;=\; \v z^{\ell h}_{s,X_i} \;-\; \hat{\v z}^{\ell h}_{s,X_i}
$
in vocabulary space, we project it using
the logit lens~\cite{nostalgebraist2020logitlens}.
To improve middle-layer readouts, we use  the additional J-lens refinement of \citet{gurnee2026workspace},
which replaces $W_U$ with $W_U J_\ell$, where $J_\ell$ is the model's
average linear account of how a vector inserted at layer $\ell$ reaches
the output.  \S\ref{app:protocol} gives the estimation and the per-occurrence decoding.

\subsection{Measurement Protocol}
\label{app:protocol}

\paragraph{The repeated-text condition.}

Several readings below compare a head's per-occurrence deviation on natural text
with its deviation on \emph{repeated} text.
The construction is that of Experiment~1 (\S\ref{sec:icl-correlates}): a real
OpenWebText chunk $c$ is presented as $[c;c]$, so that each position in the second
copy has an exact earlier copy of its own context.
We sample query occurrences only from the second copy, since the survey reports one
deviation per head per condition rather than a matched copy-1 against copy-2
difference; sampling from the second copy is also what leaves induction and
duplicate detection something to act on.
And the pass covers $300{,}000$ tokens at stride 8, with the same position and
vocabulary filters as above and against the same mean field.
Rise ranks quoted below are over all $144$ heads.

\paragraph{Occurrence mining for Lens~\ref{tool:contexts}.}

Lens~\ref{tool:contexts} keeps the top-$K$ occurrences per head by
per-occurrence deviation.  We apply two additional filters:
occurrences where the head places more than $10\%$ of its attention on
out-of-vocabulary tokens are dropped, since the single averaged ``other''
value is then carrying much of the contribution, and windows containing
garbled or rare-Unicode text are dropped for the same reason.
The vector decoded is $\v\Delta_i = \v z_i - \hat{\v z}_i$, the residual against the
context-conditional mean-field prediction, the same vector whose angle defines the
deviation used for the mining (note the contrast with \S\ref{app:positional}, whose
$\v\Delta$ is centered on the corpus type mean instead).
We decode each occurrence separately rather than averaging first, since averaging
$\Delta_i$ over occurrences mixes distinct entities and topics, which could either create
or wash out clusters.

\paragraph{The readout of Lens~\ref{tool:readout}.}

The logit-lens readout of a vector $\v u$ inserted after attention at
layer $\ell$ is $\mathrm{softmax}(W_U\,\mathrm{norm}(\v u))$, with
$\mathrm{norm}$ the final-layer normalization.  This ignores the mediation
between layer $\ell$ and the output, which at middle layers is substantial.
The refinement of \citet{gurnee2026workspace} replaces $W_U$
with $W_U J_\ell$, where
\[
  J_\ell \;=\; \E\!\left[\frac{\partial \v h^{(L)}_{t'}}{\partial \v h^{(\ell)}_{t}}\right]
\]
is the average, over positions and prompts of a natural corpus, of the
Jacobian of the final residual with respect to the residual at layer
$\ell$: the model's own average linear account of how content inserted at
layer $\ell$ arrives at the output.  We estimate $J_\ell$ once per model
from OpenWebText.  At the final layer the estimate is proportional to the
identity to nine decimals, a sanity check on the estimation.
The two readouts agree at late layers, where little mediation remains:
at layer 9 of GPT-2 the raw lens already broadly reproduces the J-lens
decode, and at layer 11 they coincide.  We report the J-lens decode in the
tables; where a head sits late enough for the two to agree, the raw lens
gives the same reading.

\paragraph{Split-half stability.}
The ranking is stable across the corpus: recomputed
end-to-end on each of two disjoint document-interleaved halves, the two
rankings have Spearman correlation $\rho = 0.999$ over the $144$ heads and share $18$
of their top twenty, the two discrepancies falling near rank twenty.

\subsection{What the Lenses Show}
\label{app:readings}

\paragraph{Natural text against repeated text.}
The repeated-text comparison of \S\ref{app:protocol} separates the top twenty by how
each group responds to repetition, and shows where the corpus ranking mis-scales.

The induction heads rise the most.  On natural text their per-occurrence deviations
sit in the bottom third of the network, since natural text rarely contains the
repetition they respond to, but under repeated text \hInd{L5H1}, \hInd{L7H2},
\hInd{L7H10} and \hInd{L6H9} show the four largest increases of any head, each
between $+0.48$ and $+0.64$.  The corpus score understates them, since their trigger
is rare in natural text.

The positional heads move the other way, the only group whose deviation \emph{falls}
under repetition: \hPos{L2H7} by the largest amount of any head, $0.395$ to $0.166$,
and the layer-1 members by $0.04$--$0.06$.  When the only structure in the text is
repetition, a head that transports absolute position becomes more predictable, not
less.

The remaining groups do not respond as units.  The letter mover L8H11 falls
under repetition and patterns with the positional cluster rather than with the movers
it is grouped with, and \hOther{L8H1}, which carries a nonzero induction score and the
SAE survey's own induction annotation, rises with the induction heads; the other band
members change little.

\paragraph{Head readings rest on occurrences, not type means.}
The readings below characterize heads from their occurrences rather than from a
type-level score, because a type's mean can be mispredicted by a small consistent
offset that produces no occurrence-level effect.  The month effect of
\hOther{L6H2} shows why.  At type level it replicates under the current kernel:
month types sit at median rank $96$ of $999$ (Mann--Whitney $p=1.4\times10^{-4}$).
But a full-network sweep finds the effect is not unique, since L3H1 and L9H10 show
stronger month elevation ($p=4.5\times10^{-5}$ and $1.0\times10^{-5}$; three of
$144$ heads survive a Bonferroni threshold).  And mining month \emph{occurrences}
directly gives a deviation difference of only $+0.007$ ($p=0.75$), while the same
test detects \hCont{L9H8}'s date sensitivity at $p=7\times10^{-8}$.

\paragraph{Characterizing heads outside the circuit literature.}  
For each head in Table~\ref{tab:ranking-cc} with no documented circuit role, we ran the following procedure:
mine the head's eight highest-deviation occurrences ($600$k
OpenWebText tokens, deviation against the context-conditional mean field),
then examine the context around the max-deviation token, the logit lens of the
moved value, and the J-lens of the moved value (the deviation vector
$\v\Delta = \v z W_O - \hat{\v z}$). Tables~ \ref{tab:l9h8}, \ref{tab:l5h2-occ},
\ref{tab:l8h3-occ} and \ref{tab:l6h8-occ} give the
retained occurrences for the four heads whose readings rest on this evidence.

\begin{table}[t]
\centering\footnotesize
\begin{tabular}{p{6.9cm} l p{4.4cm}}
\toprule
max-deviation context (query in [[\,]]) & top key (share) & J-lens decode \\
\midrule
\emph{\ldots on inauguration day, Trump's chief of[[ staff]] ordered all} & \texttt{Trump} (.86) & \emph{Trump, Ivanka, Melania} \\
\emph{\ldots had above average wait times. However, the U.K[[.]] and France} & \texttt{K} (.53) & \emph{Britain, \pounds, UKIP, London} \\
\emph{\ldots he sustained earlier in the week. Lucas Duda has[[ one]] less error} & \texttt{A} (.60) & \emph{Padres, Baseball, MLB, Astros} \\
\emph{\ldots his rookie year and the aforementioned 2013--[[14]] campaign} & BOS (.56) & \emph{NBA, Raptors, Knicks, Pacers} \\
\emph{\ldots guns over the eight-year period cost Americans more than \$6.[[6]] billion} & BOS (.51) & \emph{FBI, Obama, Federal} \\
\emph{\ldots 13 consecutive games and led the nation in[[ time]] of possession} & \texttt{points} (.66) & \emph{Lakers, NBA, Clippers, Celtics} \\
\emph{\ldots truth about Donald Trump and Russia; do so in order to[[ understand]] the methods} & BOS (.58) & \emph{Trump, TRUMP, Melania} \\
\emph{\ldots 117--146.\ Meer, J., and J[[.]] West.} & \texttt{er} (.51) & \emph{Bengal, Delhi, India, Bangalore} \\
\bottomrule
\end{tabular}
\caption{\hCont{L9H8}: the eight highest-deviation occurrences from
$600$k OpenWebText tokens, the top attended key with its share, and the top of
the J-lens decode of each occurrence's deviation vector. All 8 decode to
the document's topic.}
\label{tab:l9h8}
\end{table}

\begin{table}[t]
\centering\footnotesize
\begin{tabular}{p{7.2cm} l p{4.6cm}}
\toprule
max-deviation context (query in [[\,]]) & top key (share) & decode of the deviation \\
\midrule
\emph{\ldots .\textbackslash n\textbackslash n"I have been a big booster of the Susan G[[.]] Komen} & \texttt{ G} (0.61) & \emph{quartered, illions, ailable, duly} \\
\emph{\ldots  the event."\textbackslash n\textbackslash nWith tickets listed at \$1,000 a[[ head]], the} & \texttt{ a} (0.38) & \emph{minimum, whopping, Bundle, coffers} \\
\emph{\ldots  conflict with one another. However, our commitment to the safety, well[[-]]being,} & BOS (0.34) & \emph{contingent, Bezos, Dawkins, SPONSORED} \\
\emph{\ldots , you'll never know when, if, or how many times that[['s]] ever happened} & \texttt{ times} (0.27) & \emph{], ]), ]., ],} \\
\emph{\ldots  conducted on such a controlled, small scale as to have little, if[[ any]], negative} & \texttt{,} (0.43) & \emph{hin, ailable, afloat, incidental} \\
\emph{\ldots  be far better.\textbackslash n\textbackslash nAs with all 4-2-3[[-]]1 v} & \texttt{3} (0.43) & \emph{anthem, pont, omorphic, apiece} \\
\emph{\ldots  whether the player is available for transfer. If A agrees, only then[[ is]] B allowed} & \texttt{ then} (0.75) & \emph{nor, unsub, authorised, defence} \\
\emph{\ldots  which he was able to showcase his skills in team and one-on[[-]]one drills} & \texttt{on} (0.71) & \emph{thereof, himself, Himself, bip} \\
\bottomrule
\end{tabular}
\caption{\hCont{L5H2}.  Its attention is
on a specific nearby content token ($0.4$--$0.8$), and its deviation decodes to
\emph{context-appropriate degree and function words}: a \$$1{,}000$-a-head
ticket passage decodes to \emph{minimum, whopping, coffers, maximum}; ``little,
if any'' to \emph{whatsoever, incidental}; a transfer-rules passage
(``only then is B allowed'') to \emph{nor, authorised}; ``one-on-one drills''
to \emph{thereof, dual, apiece}. The head appears to move an abstract
quantifier/degree completion computed from the local construction.
The reading of L5H2 as movement of degree or limitation in Table~\ref{tab:ranking-cc} summarizes 
these occurrences.}
\label{tab:l5h2-occ}
\end{table}

\begin{table}[t]
\centering\footnotesize
\begin{tabular}{p{7.2cm} l p{4.6cm}}
\toprule
max-deviation context (query in [[\,]]) & top key (share) & decode of the deviation \\
\midrule
\emph{\ldots .\textbackslash n\textbackslash nThere will be good days with your boyfriend. There will[[ be]] miraculous days} & \texttt{ days} (0.87) & \emph{ones, those, highs, happiest} \\
\emph{\ldots  10\textbackslash n\textbackslash nMarch 13\textbackslash n\textbackslash nApril 9\textbackslash n\textbackslash nMay 8\textbackslash n[[\textbackslash n]]June 4} & \texttt{ 8} (0.36) & \emph{Champ, mingham, "))} \\
\emph{\ldots  7\textbackslash n\textbackslash nFebruary 10\textbackslash n\textbackslash nMarch 13\textbackslash n\textbackslash nApril 9\textbackslash n[[\textbackslash n]]May 8} & \texttt{ 9} (0.47) & \emph{mingham, actionDate, raltar, Champ} \\
\emph{\ldots  by two of the young women, identified as Jane Doe No. 1[[ and]] Jane Doe} & \texttt{ 1} (0.69) & \emph{PID, NAME, RandomRedditorWithNo, NAME} \\
\emph{\ldots  1 start with \$2.8M, ahead of the debuts[[ of]] Moana} & \texttt{ and} (0.31) & \emph{Both, Both, latter, Neither} \\
\emph{\ldots  cases involving an increase in the minimum wage with a similar control group,[[ and]] two,} & \texttt{,} (0.62) & \emph{secondly, Secondly, Secondly, Option} \\
\emph{\ldots  be deployed Tuesday, June 5. I've discussed its contents\textbackslash n\textbackslash n[[0]].9} & \texttt{10} (0.44) & \emph{uyomi, Measure} \\
\bottomrule
\end{tabular}
\caption{\hCont{L8H3}. Its max-deviation occurrences are
\emph{parallel and enumerated frames}: a repeated ``There will be $X$
days'' construction (decoding to \emph{ones, those, highs}), month--day
lists, ``Jane Doe No.~1 and Jane Doe'' (decoding to \emph{NAME, PID},
placeholder tokens), a numbered listicle, a box-office enumeration. The
attention is on the frame's anchor ($0.9$ on \texttt{days}, $0.7$ on
\texttt{1}). This suggests structure-frame tracking. 
 The reading of L8H3 as parallel and enumerated frames in Table~\ref{tab:ranking-cc} rests on these occurrences.}
\label{tab:l8h3-occ}
\end{table}

\begin{table}[t]
\centering\footnotesize
\begin{tabular}{p{7.2cm} l p{4.6cm}}
\toprule
max-deviation context (query in [[\,]]) & top key (share) & decode of the deviation \\
\midrule
\emph{\ldots  like, but if you cut it as close to the wire as you[[ can]], the} & \texttt{ as} (0.76) & \emph{pires, yip, dstg, rha} \\
\emph{\ldots  6 Friday.\textbackslash n\textbackslash nIn the video, which is nearly two and[[ a]] half hours} & \texttt{ and} (0.79) & \emph{peeled, removable, rec, reversible} \\
\emph{\ldots  by and among AIG and the big banks, is striking to say[[ the]] least.} & \texttt{ say} (0.75) & \emph{standpoint, purposes, understatement, \$.} \\
\emph{\ldots ron, the world's most expensive car. But then, who in[[ this]] economy could} & \texttt{ in} (0.63) & \emph{pires, blinked, reinvent, stole} \\
\emph{\ldots , either."\textbackslash n\textbackslash nThe conservative group met with President Trump Donald John[[ Trump]]READ:} & BOS (0.64) & \emph{ordon, uffle, ighton, arrison} \\
\emph{\ldots . That original vapor has little structure and is about as simple as it[[ could]] be.} & \texttt{ as} (0.54) & \emph{Dragonbound, Vaugh, Hubble, illusions} \\
\emph{\ldots . Feeling uncomfortable from the whole situation, I left as quickly as I[[ could]] and then} & \texttt{ as} (0.56) & \emph{dstg, Berks, Nay, MpServer} \\
\emph{\ldots \textbackslash n\textbackslash nWhat do you think? How many more times will Mighty No[[.]] 9 be} & \texttt{ No} (0.69) & \emph{sburg, ®,, Faces, Merit} \\
\bottomrule
\end{tabular}
\caption{\hCont{L6H8}. This head's highest-deviation contexts are
\emph{idiomatic comparative frames}: ``as close \ldots as you can,'' ``to say
the least,'' ``about as simple as it could be,'' ``as quickly as I could,''
with attention on the frame's pivot (\texttt{as}, \texttt{say} at
$0.5$--$0.8$). The reading of L6H8 as idiomatic comparative frames in Table~\ref{tab:ranking-cc} is based on these occurrences.}
\label{tab:l6h8-occ}
\end{table}

\paragraph{Heads at the bottom of the ranking.}
\label{app:bottom}

Table~\ref{tab:bottom} lists the fifteen \emph{lowest}-deviation
heads of each model under the deviation of \eqref{eq:ddec},
annotated with the behavioral repetition scores and,
for GPT-2, the head's \emph{mean-field impact} rank, the signal complementary to
deviation: a head with large impact carries type-level geometry that the mean field
reproduces, while a head with large deviation departs from that geometry in
context.\footnote{To measure impact we ablate the head from the rollout of
\S\ref{sec:centroid-rollout}, dropping its kernel term $P^{\ell h}$ and its value
contribution at layer $\ell$ while leaving the other heads, the MLPs and the
residual structure intact, re-run the open-loop prediction from the input
centroids, and record the drop in mean-over-layers centered cosine.  The two
columns are therefore scored against different kernels: the corpus kernel $P$ for
impact, the context-averaged context-conditional kernel for $\bar D$.}
The parenthesized IOI labels are that circuit's own head names, whose three
duplicate-token heads are 0.1, 0.10 and 3.0 \citep{wang2023ioi}, and ``FV core
head'' marks a head in the task-norm top-10 on at least one of the three tasks of
\S\ref{sec:fv}.
A head is left blank rather than guessed: none of the blank entries is named by the
IOI circuit~\citep{wang2023ioi} or the acronym
circuit~\citep{garciacarrasco2024acronyms}, and we do not fall back on the
unrefereed SAE head survey here.
Membership in the table is not an artifact of which kernel scores it: against the
corpus kernel $P$ instead, GPT-2 keeps thirteen of the fifteen, with L1H1 and L4H4
replacing \hPrev{L1H0} and \hPrev{L2H3}, and Gemma keeps all fifteen in a different order.

\begin{table}[h]
\centering\footnotesize
\begin{minipage}[t]{0.58\textwidth}
\centering
GPT-2 (median $\bar D = 0.272$)\\[2pt]
\begin{tabular}{rlccl}
\toprule
\# & head & $\bar{S}^{\ell h}$ & MF-impact rank & note \\
\midrule
1  & \hDup{L0H1}  & 0.001 & \textbf{4}   & duplicate (IOI 0.1) \\
2  & \hDup{L0H5}  & 0.003 & \textbf{5}   & duplicate (behav.\ 0.72) \\
3  & L0H3  & 0.004 & \textbf{3}   & \\
4  & L1H11 & 0.009 & \textbf{8}   & \\
5  & L0H4  & 0.017 & 38  & \\
6  & L4H7  & 0.021 & 37  & \\
7  & \hDup{L0H10} & 0.033 & \textbf{2}   & duplicate (IOI 0.10) \\
8  & \hPrev{L0H7}  & 0.052 & 14  & prev-token 0.19 \\
9  & L3H10 & 0.066 & 49  & BOS-sink 0.38 \\
10 & \hDup{L3H0}  & 0.077 & 23  & duplicate (IOI 3.0) \\
11 & L1H5  & 0.082 & 10  & \\
12 & \hPrev{L1H0}  & 0.088 & 34  & prev-token 0.18 \\
13 & L0H8  & 0.091 & \textbf{7}   & \\
14 & \hPrev{L2H3}  & 0.093 & 17  & prev-token 0.30 \\
15 & \hPrev{L2H5}  & 0.098 & 19  & prev-token 0.33 \\
\bottomrule
\end{tabular}
\end{minipage}\hfill
\begin{minipage}[t]{0.40\textwidth}
\centering
Gemma-2-2B (median $\bar{S}^{\ell h} = 0.149$)\\[2pt]
\begin{tabular}{rlcl}
\toprule
\# & head & $\bar D$ & note \\
\midrule
1  & L5H4  & 0.001 & duplicate 0.38 \\
2  & L6H0  & 0.002 & duplicate 0.19; \textbf{FV} \\
3  & L4H5  & 0.002 & \\
4  & L3H2  & 0.003 & duplicate 0.53 \\
5  & L24H4 & 0.006 & \\
6  & L2H5  & 0.007 & \\
7  & L7H4  & 0.008 & \\
8  & L0H1  & 0.009 & duplicate 0.33 \\
9  & L0H2  & 0.014 & \\
10 & L25H7 & 0.014 & \textbf{FV core head} \\
11 & L1H4  & 0.016 & duplicate 0.75 \\
12 & L2H3  & 0.017 & \\
13 & L19H4 & 0.019 & \textbf{FV core head} \\
14 & L12H6 & 0.022 & \textbf{FV core head} \\
15 & L15H6 & 0.023 & \\
\bottomrule
\end{tabular}
\end{minipage}
\caption{The fifteen lowest-deviation heads of each model.
$\bar{S}^{\ell h}$ is the $1-\cos\theta$ deviation of \eqref{eq:ddec} against the
context-averaged context-conditional kernel, as in the body.
``MF-impact rank'' is the head's rank of 144 by mean-field impact,
1 = largest, bold in the top eight.
``duplicate'', ``prev-token'' and ``BOS-sink'' are attention shares on repeated
random sequences, on the query type's earlier occurrence, the immediately
preceding position, and the BOS token respectively, listed when they exceed
$0.10$, $0.15$ and $0.35$.}
\label{tab:bottom}
\end{table}

We observe three patterns. \textbf{(i) The bottom of the ranking is where the
type-level geometry lives.} In GPT-2, four of the six most
mean-field-predictable heads are simultaneously among the top eight heads
by mean-field impact: these layer-0 heads do almost exactly what the
kernel says, and what the kernel says carries much of the centroid
prediction.
The two readings dissociate: measured against the mean field these heads are the
most predictable in the network, while measured within it they are among the most
important. \textbf{(ii) The duplicate-token machinery populates the bottom in both models.}  All
three GPT-2 IOI duplicate-token heads and the strongest behavioral
duplicate head (\hDup{L0H5}, at $0.72$) sit here, as do Gemma's two lowest-deviation heads
plus three more of its bottom fifteen.  ``Attend to your own type'' is
the diagonal of $W_{st}$, so duplicate detection is expressible in the
mean field and therefore invisible to the deviation.
\textbf{(iii) Function-vector core heads sit at the bottom in
Gemma-2-2B.} Four of Gemma's (L12H6, L25H7, L19H4, L6H0) are in its bottom
fifteen, L6H0 on the capital task.  GPT-2's function-vector heads sit low but not this low:
L11H0 ranks $19$th of $144$ and L11H11 $26$th, so we state this pattern
for Gemma only.  The heads whose task residuals
causally transport tasks in \S\ref{sec:fv} are thus among the
\emph{most} mean-field-predictable heads on generic text in Gemma.

\subsection{Layer-1 Positional Transport: Re-encoding the Query's Position}
\label{app:positional}

This appendix details the positional analysis of GPT-2's early-layer deviation referenced in \S\ref{sec:catalog-findings}: whether a head's within-type output variation is \emph{positional} rather than lexical.  This is visible because GPT-2 adds a learned positional embedding to the residual stream, so at layer~0 a head's within-type variation is exactly its OV map applied to $W_{\mathrm{pos}}[p]$. On the other hand, Gemma-2 uses RoPE, which keeps absolute position out of the values. 

All analyses here use GPT-2 small and Gemma-2-2B over OpenWebText; the within-type write deviation of a head is $\v\Delta=\v z^{\ell h}_{sX}-\overline{\v z}^{\ell h}_{s}$ with $s=T(q)$, its per-occurrence output minus the corpus (or sample) mean output $\overline{\v z}^{\ell h}_{s}$ for the query's token type. 
Note that $\v\Delta$ is a different quantity from the deviation $\Dmf$ of
\S\ref{sec:instrument}, which compares $\v z^{\ell h}_{sX}$ to the
\emph{per-context} mean-field prediction $\hat{\v z}^{\ell h}_{sX}$.  We use
$\Dmf$ to \emph{find} the cluster (Table~\ref{tab:ranking-cc}), the five
positional heads L1H2, L1H3, L1H8, L1H10, and L2H7, and $\v\Delta$ to
\emph{analyze} it, since the appendix's question is what drives the total
within-type variation of these heads' writes.
Because $\hat{\v z}^{\ell h}_{sX}$ is built from type kernels and type means, it carries no positional dependence, so the positional variance measured below is insensitive to which of the two baselines is subtracted.
\paragraph{Positional fraction.} For each head we fit each coordinate of $\v\Delta$, by ordinary least squares, as a smooth function of the normalized query position $\tilde p = p/T$: the regressors are an intercept together with the basis $\{\tilde p,\ \tilde p^2,\ \sin(2\pi k\tilde p),\ \cos(2\pi k\tilde p): k=1,\dots,4\}$ ($11$ in all).  We report \emph{posVar}, the fraction of $\v\Delta$'s variance \emph{about its mean} that the fitted position function explains; the intercept absorbs the mean and contributes nothing to posVar. This is a more conservative measure than \emph{decoding} position from $\v\Delta$ (a $20$-dimensional positional component is easily decodable while contributing little variance). We probe the high-deviation heads studied in the text, the highest-deviation head in each layer, and control heads; Gemma uses the same procedure.

\paragraph{What is moved (decoding).} Knowing that the cluster's output
varies with position (posVar) does not say \emph{what} positional
content it writes.  There are three candidates: the \emph{absolute} position of the
query; the absolute position of the attended key; or the \emph{relative}
offset between them.  Orthogonally, the written content may be merely
position-correlated, or a deterministic function of the
model's own position representation, a linear image of the embedding
row $W_{\mathrm{pos}}[p]$.
Restricting to the layer-1 cluster (\hPos{L1H2/H3/H8/H10}) and \hPos{L2H7}, four
measurements separate these.  (i)~Ridge regression decodes the query's
absolute position from $\v\Delta$ at $R^2\approx0.67$ (position-shuffled
baseline $\approx 0$).  (ii)~Regressing
$\v\Delta$ directly on the embedding rows $W_{\mathrm{pos}}[p]$ shows
that $0.61$--$0.67$ of the write's variance is a linear image of
$W_{\mathrm{pos}}$: the write is a fixed linear function of the model's
own absolute-position representation, not some other position-correlated
signal.
(iii)~Specificity: a subset of layer-1 heads carries $W_{\mathrm{pos}}$
(variance $0.56$--$0.67$) while others do not ($\le0.13$), and position
per se remains broadly decodable even from non-carriers (for example,
L2H1 at $R^2\,0.56$ with $W_{\mathrm{pos}}$ variance $\le0.04$), so
$W_{\mathrm{pos}}$-transport, not mere positional decodability, is the
discriminating property; the carriers' position-readout directions are
mutually near-orthogonal ($|\cos|\le0.11$), each head writing its tag into
its own subspace.  (iv)~Relative-code control.  A relative code predicts
that the write is a function of the query--key offset
$o = p - k$.  We measured each head's own attended-key position
on fresh text and partitioned $\v\Delta$'s variance between offset
features (one-hot small offsets plus coarse bins) and
$W_{\mathrm{pos}}[p]$.  The cluster's attention geometry is
heterogeneous, and the relative hypothesis fails differently in each
regime:
\begin{itemize}
\item \textbf{\hPos{L1H2} attends locally} (offset s.d.\ $2.0$, median $1$),
so a relative code would predict a
nearly constant write.  Instead, $0.67$ of $\v\Delta$'s variance
sweeps through a linear image of $W_{\mathrm{pos}}[p]$ as $p$ ranges over
$8$--$191$.
\item \textbf{\hPos{L1H10}'s offset varies widely} (s.d.\ $70$, spanning
$0$--$161$), so a relative code has ample variance to appear in,
but offset features explain only $0.26$ of $\v\Delta$'s variance against
$0.66$ for $W_{\mathrm{pos}}[p]$, and add nothing ($+0.00$) once
$W_{\mathrm{pos}}[p]$ is included: the offset-correlated part of the
write is just the part that co-varies with absolute position.
\item \textbf{\hPos{L1H3}, \hPos{L1H8}, and \hPos{L2H7} attend to BOS}, for
$90$--$100\%$ of query positions.  The question degenerates for them: the
offset to a fixed key at position zero is the absolute position, so no
regression could separate the two hypotheses, and what identifies the
content is finding~(ii), that $0.61$--$0.65$ of the write is
a linear image of $W_{\mathrm{pos}}$.
\end{itemize}
GPT-2 has no relative-position
mechanism, position entering only through the learned absolute table
$W_{\mathrm{pos}}$, so an absolute tag has a source in the residual
stream while a relative one would have to be computed from scratch.
The moved content is therefore an \emph{absolute
position tag}, a deterministic image of the key's absolute
position, and not a relative offset or incidental position
correlation.

\paragraph{Mechanism: the BOS gate.}  The relative-code control leaves a
puzzle.  For \hPos{L1H3}, \hPos{L1H8}, and \hPos{L2H7} the top-attended key is BOS, and the
BOS value is the same vector at every query position, carrying
no positional variance, so these heads cannot be reading
position out of the token they attend to.  Their write nonetheless varies
with $p$ through the QK side: the query at
position $p$ carries $W_{\mathrm{pos}}[p]$ into the attention
computation, so the amount of attention parked on BOS can depend
on $p$ even though the value retrieved there does not.  Measurement
confirms this.  The scalar attention-on-BOS, $a(p)$, is a nearly
deterministic smooth function of the query position: regressing $a$ on
$W_{\mathrm{pos}}[p]$ gives $R^2 = 0.996$ (\hPos{L1H3}), $0.935$ (\hPos{L1H8}), and
$0.911$ (\hPos{L2H7}).  That one scalar by itself explains $0.36$, $0.23$, and
$0.41$ of $\v\Delta$'s total variance respectively, and the top singular
direction of the positional write, the line through the fixed
BOS-value image, carries $0.66$--$0.77$ of it; the remainder comes
from the complementary tail of attention over nearby keys, whose
distribution also shifts with $p$.  The positional
write of every head in the cluster is effectively rank~$3$: the
top three singular directions of the $W_{\mathrm{pos}}$-fit carry all
of its variance, \hPos{L1H2} and L1H10 included.  Each head thus emits
the query's position as a coefficient pattern over about three fixed
write directions.  The mechanism is a \emph{gate}: the
head reads $p$ off its own QK circuit, leans on the BOS value as a fixed
reference vector, and generates a position-dependent output, with
nothing positional moved from the attended token.  This is a concrete
functional use for the attention-sink resting behavior of
early heads.

\paragraph{Magnification versus re-encoding.}  The residual stream at
position $p$ already contains $W_{\mathrm{pos}}[p]$, deposited by the embedding,
so a head whose output is a function of $p$ might simply be amplifying (or
attenuating) the code that is already there.  The two possibilities separate in
write space: if the cluster magnifies, its positional write should lie inside
the subspace spanned by the existing code and align with $W_{\mathrm{pos}}[p]$
pointwise; if it re-encodes, the write should occupy new directions.
The top ten principal components of the centered code
$W_{\mathrm{pos}}[8{:}191]$ capture $99.8\%$ of its variance, yet the heads'
positional writes place little of their own variance in that subspace: $5.0\%$
(\hPos{L1H2}), $2.3\%$ (\hPos{L1H3}), $0.2\%$ (\hPos{L1H8}), $7.1\%$ (L1H10),
and $18.8\%$ (\hPos{L2H7}).  The pointwise cosine between the write and
$W_{\mathrm{pos}}[p]$ is near zero for every head ($|\cos|$ between $0.01$ and
$0.18$).  The write is also as large as or larger than the code itself (RMS norm
ratios $0.97$--$1.90$).  The cluster is therefore writing a \emph{second}
position code, at full amplitude, rather than reinforcing the embedding's.  The
ablation supplies the causal evidence: it removes only the cluster heads'
position subspace from their outputs, leaving the embedding's
$W_{\mathrm{pos}}[p]$ untouched in the residual stream, and loss still rises by
$+0.0117$ nats, $38\times$ the controls.  The original code does not supply what
the layer-2 consumers need: downstream reads the re-encoded dimensions.

\paragraph{The geometry of the new code.}  Do the five heads write
position into the \emph{same} new subspace, or does each keep its own?
Each head's positional write is effectively two-dimensional (the top
two singular directions carry $0.97$--$0.98$ of it), so we compare the
heads pairwise by the mean $\cos^2$ of the principal angles between
their write subspaces (chance level for random subspaces of this size in
$768$ dimensions: $0.004$).  The overlaps form a spectrum:
\begin{itemize}
\item \textbf{The two non-BOS heads nearly share a code.}  \hPos{L1H2} and
L1H10 overlap at $0.61$, and in variance-weighted terms $0.85$ of
L1H10's positional write lies inside \hPos{L1H2}'s span: the heads that attend
to real (non-BOS) keys write into substantially the same channels.
\item \textbf{\hPos{L1H8} and \hPos{L2H7} are the most private.}  Their mutual
overlap is $0.02$, essentially disjoint, and each overlaps the
remaining heads only modestly ($0.02$--$0.27$ for \hPos{L1H8};
$0.14$--$0.31$ for \hPos{L2H7}).  \hPos{L2H7} is the cluster's layer-2
member, writing after the layer-1 heads.
\item \textbf{No pair is orthogonal either.}  Every overlap sits well
above the $0.004$ chance floor.  The code is a partially redundant
bus: a few heavily shared channels plus per-head private ones.
\end{itemize}
The union of these partially overlapping channels sits
inside the $20$-dimensional cluster subspace, capturing $0.88$ of the
combined output's variance, used in the ablation below.

The shape of the code within each channel is also informative.  A
BOS-gated head might encode its distance from BOS in the \emph{magnitude}
of what it writes.  The gauge is real:
for the BOS-attending heads the attention on BOS is strongly
\emph{monotone} in the query position (Spearman $\rho$ of $-0.85$
for \hPos{L1H3}, $-0.93$ for \hPos{L1H8}, and $+0.94$ for \hPos{L2H7}; the layer-1 pair
releases attention from BOS as $p$ grows while
\hPos{L2H7} accumulates it), so each head maintains a graded reading of
how far into the context it is.  But position is \emph{not} encoded
in the write's magnitude: the norm of the positional write correlates
only weakly with $p$ ($|\rho| \le 0.31$ for these heads).  Instead the
\emph{signed coefficient} along the leading axis sweeps monotonically
from positive through zero to negative as $p$ grows ($\rho = -0.93$,
$-0.96$, $+0.96$), making the norm U-shaped, smallest mid-context
where the coefficient crosses zero, while the second axis's
coefficient traces an arc.  The code is thus a smooth two-dimensional
curve, roughly (monotone trend, curvature), traversed as $p$ advances:
\emph{coordinate} coding along fixed axes, not magnitude coding.
This is the same format as the embedding's own code: the
participation ratio of the centered $W_{\mathrm{pos}}[8{:}192]$
spectrum is $2.0$ effective dimensions, so the original code is likewise
a low-dimensional curve with position given by location along it.  The
re-encoding therefore preserves the representational format, a smooth
two-dimensional position curve, while relocating it into channels the
embedding does not occupy, with per-head gain and a per-head warp of the
position axis.

\paragraph{Why re-encode?}  We flag this paragraph as interpretation
rather than measurement.  The embedding's code is a single shared
signal, frozen in its geometry and entangled with token content in the
same residual directions; it cannot be scaled, cleaned, or reshaped for
a particular consumer.  Republishing position through attention buys
three things.  (i)~\emph{Clean channels}: each carrier's write is
position and essentially nothing else (posVar $0.90$--$0.99$), in
directions of its own, so a downstream QK circuit can read position
without also picking up token content.  (ii)~\emph{Independent gain}:
each channel's amplitude is set by a head that can be modulated
separately from the embedding and from the other carriers.
(iii)~\emph{A reshaped scale}: the gate $a(p)$ is a softmax-derived
function of position, so the re-encoded signal can warp the position
axis, compressing late positions or sharpening early ones, in ways
the fixed embedding cannot.  On this reading the cluster republishes the
query's absolute position in a standardized, consumable form, and the
dereferencing experiment below identifies the subscribers.

\paragraph{Dereferencing (causal ablation).} We compute the $20$-dimensional subspace $S$ into which the cluster writes position, the top principal directions of the $W_{\mathrm{pos}}$-fit of the cluster heads' combined per-head output, capturing $0.88$ of its variance, and, mid-forward, project that component out of the cluster heads' outputs. We compare the next-token-loss increase to two controls: removing the \emph{same} $S$ from the layer-1 heads that do not write into it (specificity), and removing a random equal-dimension subspace. The carrier ablation costs $+0.0117$ nats versus $+0.0003$ (non-carrier) and $+0.0011$ (random), a $38\times$ specificity ratio. To find the consumers we record, per downstream head, the $L_2$ change in its attention pattern under the ablation; the change is concentrated in layer-2 heads and decays with depth.

\paragraph{Function (category breakdown).} Under the same ablation we group the per-token loss increase by the category of the token being predicted: tokens inside a number, word-initial tokens, within-word subword continuations, and punctuation. The increase is largest for numeric tokens ($+0.028$, $2.6\times$ the average), moderate for word-initial tokens ($+0.016$), negligible for punctuation ($+0.002$), and absent for subword continuations ($-0.006$), identifying forward sequence progression (next-word and numeric/ordinal continuation), not detokenization, as the supported behavior.

\subsection{Gemma-2-2B Head Survey}
\label{app:gemma}

We repeat the head survey 
of \S\ref{sec:catalog-findings}
on Gemma-2-2B (26 layers, 8 query heads each,
grouped-query attention, rotary position embeddings).  
The measurements are those of \S\ref{sec:instrument}, computed from a 2M-token
OpenWebText pass over the model's $208$ heads, with the same support mask
($C \ge 20$) and budget as the GPT-2 survey in the body.
No circuit catalog exists for Gemma, so the survey
tests whether the method reproduces itself but not whether its readings agree with
an independent account.
Table~\ref{tab:gemma} lists the fifteen heads with the largest deviation.

\begin{table}[h]
\centering\small
\begin{tabular}{rlcccc}
\toprule
\# & head & $\bar S^{\ell h}$ & $M_{\mathrm{cov}}$ & $M_{\mathrm{ctx}}$ & induction \\
\midrule
1  & L13H7   & 0.481 & $+0.30$ & $+0.18$ & 0.00 \\
2  & L15H5   & 0.321 & $+0.11$ & $+0.21$ & 0.00 \\
3  & L15H0   & 0.307 & $+0.18$ & $+0.12$ & \textbf{0.94} \\
4  & L23H2   & 0.304 & $+0.13$ & $+0.17$ & 0.07 \\
5  & L24H7   & 0.301 & $+0.12$ & $+0.18$ & 0.09 \\
6  & L1H0   & 0.298 & $+0.07$ & $+0.23$ & 0.00 \\
7  & L4H4   & 0.294 & $+0.18$ & $+0.11$ & 0.33 \\
8  & L17H3   & 0.293 & $+0.14$ & $+0.16$ & 0.37 \\
9  & L14H1   & 0.290 & $+0.16$ & $+0.13$ & 0.05 \\
10 & L14H3   & 0.285 & $+0.15$ & $+0.13$ & 0.01 \\
11 & L22H4   & 0.285 & $+0.10$ & $+0.19$ & 0.35 \\
12 & L14H6   & 0.284 & $+0.17$ & $+0.11$ & 0.02 \\
13 & L7H5   & 0.278 & $+0.08$ & $+0.19$ & 0.00 \\
14 & L10H3   & 0.277 & $+0.14$ & $+0.13$ & 0.00 \\
15 & L13H5   & 0.275 & $+0.10$ & $+0.17$ & 0.00 \\
\bottomrule
\end{tabular}
\caption{Gemma-2-2B: the fifteen heads with the largest corpus deviation
$\bar S^{\ell h}$ \eqref{eq:ddec}, over the $999$ OpenWebText query types with at
least $50$ occurrences.}
\label{tab:gemma}
\end{table}

The ranking again recovers induction machinery without supervision.  The
model's strongest behavioral induction head, L15H0 (induction $0.94$, the largest
of all $208$ heads), ranks third, and three more induction heads follow in the top
fifteen (L4H4, L17H3, L22H4).  Across all $208$ heads the deviation--induction
Spearman correlation is $\rho = 0.45$.

The additive split places the induction heads on the \emph{covariance}
side: L15H0 draws a covariance share of $0.60$ against a network mean of $0.52$,
and L4H4 sits at $0.63$.  This matches GPT-2, where the strongest induction head is
also the most covariance-leaning of the documented heads
(Table~\ref{tab:ranking-cc}): attending to whatever followed the query's earlier
occurrence is a pattern effect, not a value effect.  The covariance lean is a
property of these particular heads, not a network trend; across all $208$ heads
induction score and covariance share are mildly \emph{anti}-correlated
($\rho = -0.31$).
  These heads are also at rest on generic text: on the
per-occurrence survey L15H0 scores $1-\cos\theta = 0.070$, rank $203$ of $208$,
the trigger-dependent signature GPT-2's induction heads also show.

Unlike GPT-2, no early-layer positional cluster appears, since RoPE keeps
absolute position out of the values (\S\ref{app:positional}).  The top of the
per-occurrence survey is instead a dense mid-stack band: all $20$ of its top $20$
heads lie in layers $7$--$17$, median layer $11$.  The band leans covariance
rather than contextualization, mean split $M_{\mathrm{cov}} = 0.43$ against
$M_{\mathrm{ctx}} = 0.32$.%

\section{Function-Vector Extraction: Baselines, Controls, and Results}
\label{app:fv}

Here we provide supporting evidence and additional results for \S\ref{sec:fv}.  We first
validate that our
average-indirect-effect (AIE) baseline reproduces the published
\citet{todd2024fv} selection, that the null subtraction is what makes MFA
selection work, and that the head-to-head comparison does not depend on the
injection protocol. We then present the task-by-task results behind \S\ref{sec:fv} and close with the full analysis of what an injected vector contains at GPT-2 small's scale.

\subsection{Validations}
\label{app:fv-validations}
The results below compare two ways of selecting function-vector heads: the MFA
task-residual and \citeauthor{todd2024fv}'s average indirect effect (AIE).
In both cases we take the top-10 of Gemma-2-2B's $208$ heads; the MFA selection is stable
across the three splits (sharing $7$--$10$ of ten heads), and the AIE selection
is held fixed per task ($20$ clean and $10$ corrupted prompts).

\paragraph{The AIE baseline reproduces the published selection.}  We first anchor to
GPT-J-6B and use the ten universal heads of \citet{todd2024fv} verbatim (L15H5,
L9H14, L12H10, L8H1, L11H0, L13H13, L8H0, L14H9, L9H2, L24H6), which fixes
head-for-head comparability with their results.  Running our own AIE procedure
over all $448$ GPT-J heads recovers $7$, $8$, and $7$ of those ten on the antonym,
capital, and past tasks, and reproduces their top five exactly on the antonym task, even
though their set is averaged over ${\sim}40$ tasks and ours is computed individually per
task.  Thus, we confirm that our AIE arm is a faithful reimplementation.  
That fidelity comes at a cost the MFA selection avoids:
AIE sweeps every head with a patched forward pass per corrupted prompt ($448$ prompts
on GPT-J), where the task residual is read from a single forward pass.  To apply MFA to GPT-J
we estimate the type-level null at `\texttt{:}'
directly, as the corpus-mean write of each head at `\texttt{:}' positions
($241$ occurrences from an OpenWebText slice), rather than building full
mean-field statistics; this is the quantity $\hat{\v z}^{\ell h}_{\text{`:'}}$
approximates elsewhere.  GPT-2 small is treated separately (\S\ref{app:fv-gpt2}).

\paragraph{The null subtraction is what selects task heads.} MFA ranks heads
by the magnitude of the deviation \eqref{eq:taskresid}, which subtracts the
corpus baseline from the head's task-average output.  The body assumes
throughout that this subtraction is what matters, so we first check that it carries
the selection.  To do so, we define a raw-magnitude control, which ranks by the same average with nothing
subtracted.  The two rankings agree on only $5$--$6$ heads of ten (on
the original 20-pair pools), and transfer under the raw-magnitude selection
collapses on capital ($0.00$ at every scale): \emph{without the subtraction,
selection is captured by heads whose outputs are large for reasons unrelated to
the task, which are useless as function vectors.}

\paragraph{The injection protocol does not bias the comparison.}  We consider two injection protocols.
In the first (Table~\ref{tab:fv-matrix-body}), we sum the extracted vectors and add them to the residual; in the second (reported only in this appendix) we inject each extracted vector into its corresponding head output.
The two
head-to-head tables inject the selected heads differently (\S\ref{app:fv-results}),
so we check that the choice does not reorder the selections it is used to
compare.  Per-head injection is uniformly stronger than summing the heads into
one vector, by $0.127$ averaged over the $72$ paired
task$\times$split$\times$selection$\times$construction cells ($0.570$ against
$0.442$, Wilcoxon $p<10^{-4}$).  That penalty falls almost equally on the two
selections ($-0.125$ for AIE against $-0.130$ for MFA, Mann--Whitney $p=0.83$),
even though the AIE heads sit at L12--L15, much further from the source layer than the MFA heads at L6--L9.  The protocol rescales the comparison without reordering it, so the
selection verdicts hold under either table.

\subsection{Detailed Results}
\label{app:fv-results}

\begin{table}[t]
\centering\small
\begin{tabular}{lccccc}
\toprule
 & & \multicolumn{2}{c}{AIE-selected heads} & \multicolumn{2}{c}{MFA-selected heads} \\
task & AIE$\cap$MFA & raw & deviation & raw & deviation \\
\midrule
antonym     & 2/10 & 0.79 & \textbf{0.86} & 0.77 & 0.79 \\
capital     & 2--3/10 & \textbf{0.90} & \textbf{0.88} & 0.20 & 0.19 \\
past        & 3/10 & 0.80 & 0.89 & \textbf{0.97} & \textbf{0.97} \\
language    & 2/10 & 0.27 & 0.32 & 0.53 & \textbf{0.60} \\
currency    & 2--3/10 & 0.53 & \textbf{0.76} & 0.00 & 0.00 \\
translation & 2/10 & 0.03 & 0.24 & 0.68 & \textbf{0.71} \\
\bottomrule
\end{tabular}
\caption{Gemma-2-2B: zero-shot transfer for each combination of head
selection and vector construction, with the overlap of the two selections.
Bold: best per task, ties within $0.02$.  These cells use per-head injection,
each selected head's vector added at that head's own layer at scale $10$, and
are therefore not the same measurement as Table~\ref{tab:fv-matrix-body}, which
uses the summed single-vector injection of \citet{todd2024fv}.}
\label{tab:fv-matrix}
\end{table}

These tables separate two choices in the pipeline: which heads are selected,
and how their vectors are constructed.

\paragraph{Gemma-2.}  We first consider head selection.
In Gemma-2,
the two head selections overlap by only $2$--$3$ heads of
ten, but neither dominates.  Under the per-head injection (Table~\ref{tab:fv-matrix}) AIE
selection is superior on two tasks (capital, $0.90$ vs.\ $0.20$, and
currency, $0.76$ vs.\ $0.00$), MFA head selection is superior on three
tasks (past, language, and translation, with confidence intervals
disjoint on the latter two), and the two are statistically
indistinguishable on antonym.  Under the summed injection of
Table~\ref{tab:fv-matrix-body} the same picture holds with language moving to
a tie, which is the one per-task verdict the protocol changes.
AIE also has less headroom on some tasks than others.  It scores heads
by recovery over a label-corrupted baseline, so where the answer is guessable
from the query that baseline is already high: mean corrupted-prompt answer
probability is $0.35$ (antonym) and $0.28$ (currency), against $0.17$ (capital),
$0.16$ (language), $0.11$ (past), and $0.04$ (translation).  Its ranking is
correspondingly noisier on the guessable tasks.
The two selections differ because they look in different places: on every task
the AIE top-10 concentrates on the same compact mid-stack group (L14H0 is
rank~1 on all six tasks, and L12H1, L14H1, L13H4, L14H3, L13H5, L15H3 recur),
while the MFA norms select earlier-layer writers (an L6--L9 band) whose
selection varies more by task.  On the three added tasks the recurring group is
recovered in its entirety.

Turning to vector construction, the MFA deviation dominates: under per-head injection, across the
twelve selection$\times$task cells the deviation vector matches or beats
the raw average in ten, the two exceptions both on capital and both within the
$0.02$ tie margin, with decisive margins exactly where the raw average is
weakest ($+0.23$ on currency, $+0.21$ on translation, both with AIE-selected
heads; at most $0.07$ on MFA-selected heads, where the largest shortfall
is $0.01$, on capital).  Under the summed injection of
Table~\ref{tab:fv-matrix-body} this advantage grows on AIE-selected heads
($+0.200$ against $+0.106$) and reverses on MFA-selected heads, from $+0.018$
(n.s.) to $-0.036$ ($p=0.018$).

\paragraph{GPT-J.}  With \citeauthor{todd2024fv}'s published set as reference, MFA head selection
recovers $7/10$ (antonym), $5/10$ (capital), and $4/10$ (past) of their
heads, and its transfer ties theirs on all three tasks; the apparent
antonym gap seen in smaller pools does not survive at $n=120$
(Table~\ref{tab:fvj-matrix}; scale 2 is used throughout on GPT-J).  Unlike
Gemma-2, the raw-norm control and MFA-norm selections agree on GPT-J: the
function-vector carriers there are not masked by generic-content heads.

\begin{table}[t]
\centering\small
\begin{tabular}{lccccc}
\toprule
 & & \multicolumn{2}{c}{Todd-selected heads} & \multicolumn{2}{c}{MFA-selected heads} \\
task & Todd$\cap$MFA & raw & deviation & raw & deviation \\
\midrule
antonym & 7/10 & 0.86 & 0.85 & 0.84 & 0.85 \\
capital & 5/10 & 0.76 & 0.76 & 0.76 & 0.77 \\
past    & 4/10 & 1.00 & 1.00 & 0.97 & 0.97 \\
\bottomrule
\end{tabular}
\caption{GPT-J-6B, the analog of Table~\ref{tab:fv-matrix}, with Todd
et al.'s published causal set in place of the AIE selection.  Overlap is
measured on the original 20-pair pools.  No cell is bolded: within each task
the columns lie within overlapping confidence intervals.}
\label{tab:fvj-matrix}
\end{table}

\paragraph{Overall.}  No single pipeline wins outright, but their union covers all six tasks:
every task has a FV transported at $0.60$--$0.97$ by at least one
selection-vector combination.  On GPT-J (Table~\ref{tab:fvj-matrix}) the two
selections are statistically indistinguishable on every task despite choosing
largely different heads, with every within-task difference inside the $95\%$
confidence intervals, which is why no cell is marked; random ten-head controls
transfer at $0.00$--$0.01$.  In other words, where the task signal is strong the pipelines are
interchangeable, and where it is weak they fail in different places.

\subsection{Decoding Category Vectors}
\label{app:fv-gpt2}

GPT-2 is  competent at two of the three original
tasks (few-shot $0.54$ capital and $0.90$ past tense on the $60$-pair pools).
On capital, injecting the deviation vector makes the model's top reply a capital
city on all $120$ held-out queries, but (aside from one lucky guess) the city it names is incorrect.
Past tense behaves differently: the deviation vector reaches $0.22$
accuracy, on a task whose answer can be obtained from the query
itself by a superficial change, but it is also inconsistent on providing categorical answers.
To see whether this reflects what the vector contains or only how the model reads it
out, we decode the task residuals and the injected vectors directly, for GPT-2 and,
as a contrast, Gemma-2.

\paragraph{Task residuals are decodable via J-Lens.}
\label{app:fv-jlens}
The lenses of \S\ref{sec:instrument} may be used as well with task-specific
deviations.  We demonstrate for GPT-2, by decoding each head's \emph{per-prompt}
residual $\v z^{\ell h}_{\text{`:'},X} - \hat{\v z}^{\ell h}_{\text{`:'}}$ through
$J_\ell$ \citep{gurnee2026workspace} for its layer and recording the rank of that
prompt's correct answer (in its leading-space token form, the form the corrected
prompt format elicits).  The MFA residual for top task-residual heads
names the answer or its ingredients via the J-lens. On
country$\to$capital, L8H11's and \hCont{L9H8}'s decodes have median answer rank
$5$ and $6$ of $50{,}257$, L10H3's $21$, and the name mover L10H0's $10$,
with individual prompts exact: for \emph{Sweden\,:}, L10H0's residual
decodes \texttt{\ Stockholm} at rank 0 (top-3: \emph{Stockholm, Swedish,
Sweden}), and L10H3's top-3 is a capital-city family (\emph{Helsinki,
Tokyo, London}). The division of labor is visible in the decodes:
\hCont{L9H8} (the aggregator of Table~\ref{tab:l9h8}) writes the query's
\emph{category} (\emph{Swedish, Swed, Sweden}; rank 6 because the answer
is nearby in that topical family), while L10H0 and L10H3 write the
\emph{answer} family. On present$\to$past the best decoding heads are
L11H5 (median rank $67$; for \emph{show\,:} its top-3 is \emph{show,
showed, show}) and L9H1 ($377$).  Even L8H6's poorly-decoding
residual (median $380$) is category-formatted: its top tokens are past
forms of the \emph{wrong} verbs (\emph{prayed, laughed, drank}),
the category-without-binding behavior established below.
On antonym, the task at GPT-2's capability floor, no head's residual
decodes the answer well (best median rank $341$). As in the catalog,
per-prompt decoding, not averaging, is what makes the content legible.

\paragraph{Memorization versus binding.}  Decoding the
aggregated injected vector itself (each top head's component through the
J-lens at its layer) allows us to distinguish FV content from model
behavior, and sharpens the answers-versus-category contrast between GPT-2 and
Gemma-2.  We rank four token populations in each decode: the ten
extraction-pool answers (seen in prompts), the ten held-out evaluation
answers, fifteen same-category words appearing nowhere in the task, and
fifteen unrelated words.  On GPT-2 the best-decoding capital head (L10H3) ranks
all three answer populations at the top of the vocabulary: seen answers at
median rank $9$, held-out at $37$, out-of-task capitals at $34$, against
${\sim}21{,}000$ for unrelated words. The GPT-2 function vector's \emph{content}
is thus category-wide, and the behavioral parroting reflects a small but
consistent edge for seen answers at the very top ranks, which argmax amplifies
into a deterministic preference.  On the past task the gradient is steeper
(seen $44$, held-out $265$, out-of-task $1{,}446$), closer to answer storage.

In contrast, Gemma-2's function vectors show \emph{no seen-answer preference at
all}: its best capital head (L17H0) ranks seen/held-out/out-of-task at
$23/52/89$ of a $256$k vocabulary, and on the past task the held-out answers
outrank the seen ones ($1{,}550$ vs $3{,}028$).

These results suggest that GPT-2's vector remembers the answers it saw, with
the category elevated beneath them; Gemma-2's vector carries the task with no
memory of which exemplars taught it.  GPT-2's deficit is thus better described
as a failure of argument \emph{binding} at readout than as an absence of
category content.

\end{document}